\documentclass[preprint,12pt,authoryear]{elsarticle}

\usepackage{amssymb}
\usepackage{amsthm}
\usepackage{amsmath}

\usepackage{subcaption}
\usepackage{algorithmic}
\usepackage{dirtytalk}
\usepackage{rotating}

\usepackage[linesnumbered,ruled,vlined]{algorithm2e}
\SetKwInput{KwInput}{Input}        
\SetKwInput{KwOutput}{Output}       
\usepackage[detect-all]{siunitx}
\usepackage{xcolor}
\newcommand{\mname}{EJ-GAN}

\usepackage{apalike}
\journal{Expert Systems with Applications}

\begin{document}

\begin{frontmatter}



\title{Incorporating Experts' \textcolor{black}{Judgment}
 into \textcolor{black}{Machine Learning Models}}

\author[label1]{Hogun Park\corref{cor1}%
\fnref{fn1}}
\ead{hogunpark@skku.edu}

\author[label2]{Aly Megahed\fnref{fn1}}

\author[label2]{Peifeng Yin\fnref{fn1}}

\author[label2]{Yuya Ong}

\author[label2]{Pravar Mahajan\fnref{fn1}}

\author[label2]{Pei Guo\fnref{fn1}}

\cortext[cor1]{Corresponding author.}
\address[label1]{Sungkyunkwan University, Suwon, Republic of Korea}
\address[label2]{IBM Research – Almaden, San Jose, CA, USA}
\fntext[fn1]{Work carried out during employment at IBM Research – Almaden.}

\begin{abstract}
\textcolor{black}{Machine learning (ML)} models have been quite successful in predicting outcomes in many applications. However, in some cases, domain experts might have a \textcolor{black}{judgment} about the expected outcome that might conflict with the prediction of ML models. One main reason for this is that the training data might not be totally representative of the population. In this paper, we present a novel framework that aims at \textcolor{black}{leveraging experts' judgment to mitigate the conflict}. The underlying idea behind our framework is that we first determine, using a generative adversarial network, the degree of representation of an unlabeled data point in the training data. Then, based on such degree, we correct the \textcolor{black}{machine learning} model's prediction by incorporating the experts' \textcolor{black}{judgment} into it, where the higher that aforementioned degree of representation, the less the weight we put on the expert intuition that we add to our corrected output, and vice-versa. We perform multiple numerical experiments on synthetic data as well as two real-world case studies (one from the IT services industry and the other from the financial industry). All results show the effectiveness of our framework; it yields much higher closeness to the experts' \textcolor{black}{judgment} with minimal sacrifice in the prediction accuracy, when compared to multiple baseline methods. We also develop a new evaluation metric that combines prediction accuracy with the closeness to experts' \textcolor{black}{judgment}. Our framework yields statistically significant results when evaluated on that metric.

\end{abstract}




\end{frontmatter}


\section{Introduction} \label{Introduction}

\begin{figure}[!b]
\centering
  \begin{subfigure}{0.45\textwidth}
    \includegraphics[width=\linewidth]{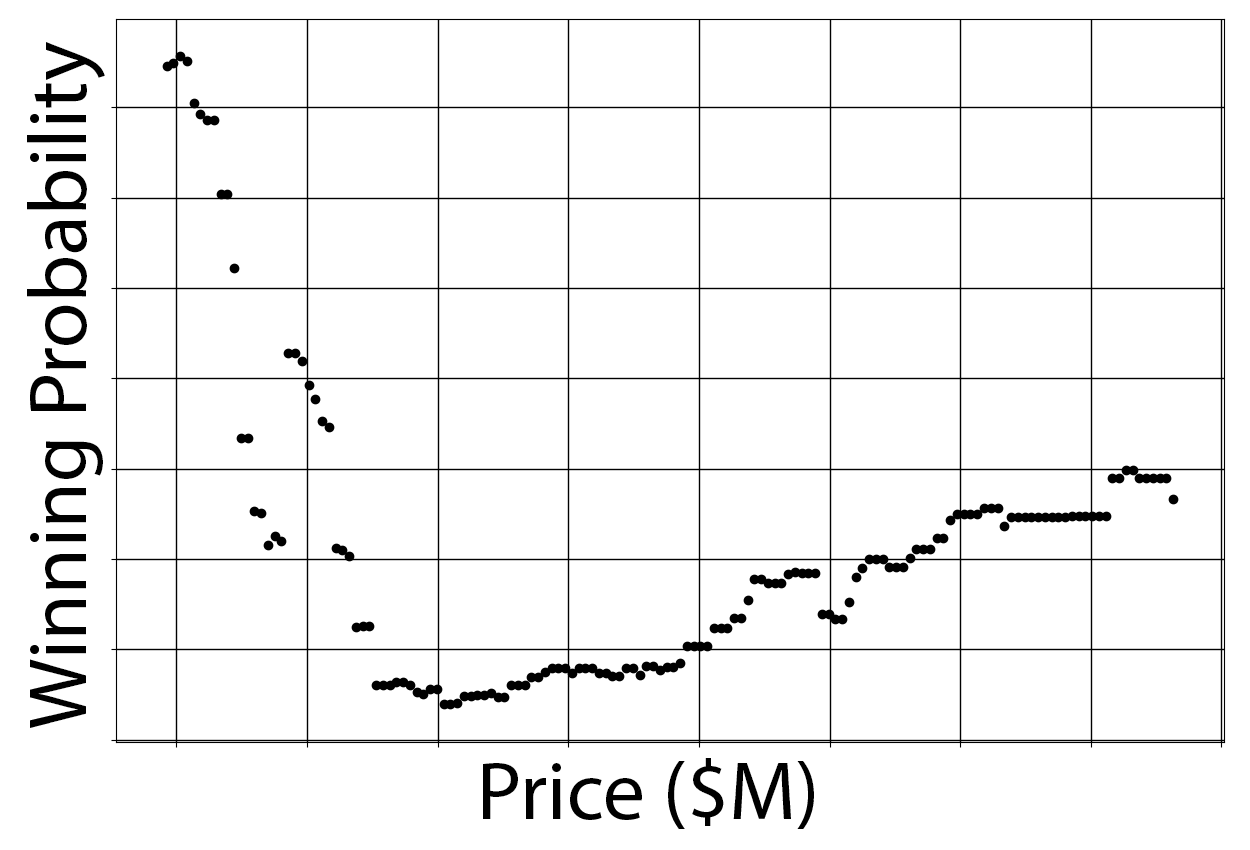}
    \caption{Deal 1}
  \end{subfigure}
  \begin{subfigure}{0.45\textwidth}
    \includegraphics[width=\linewidth]{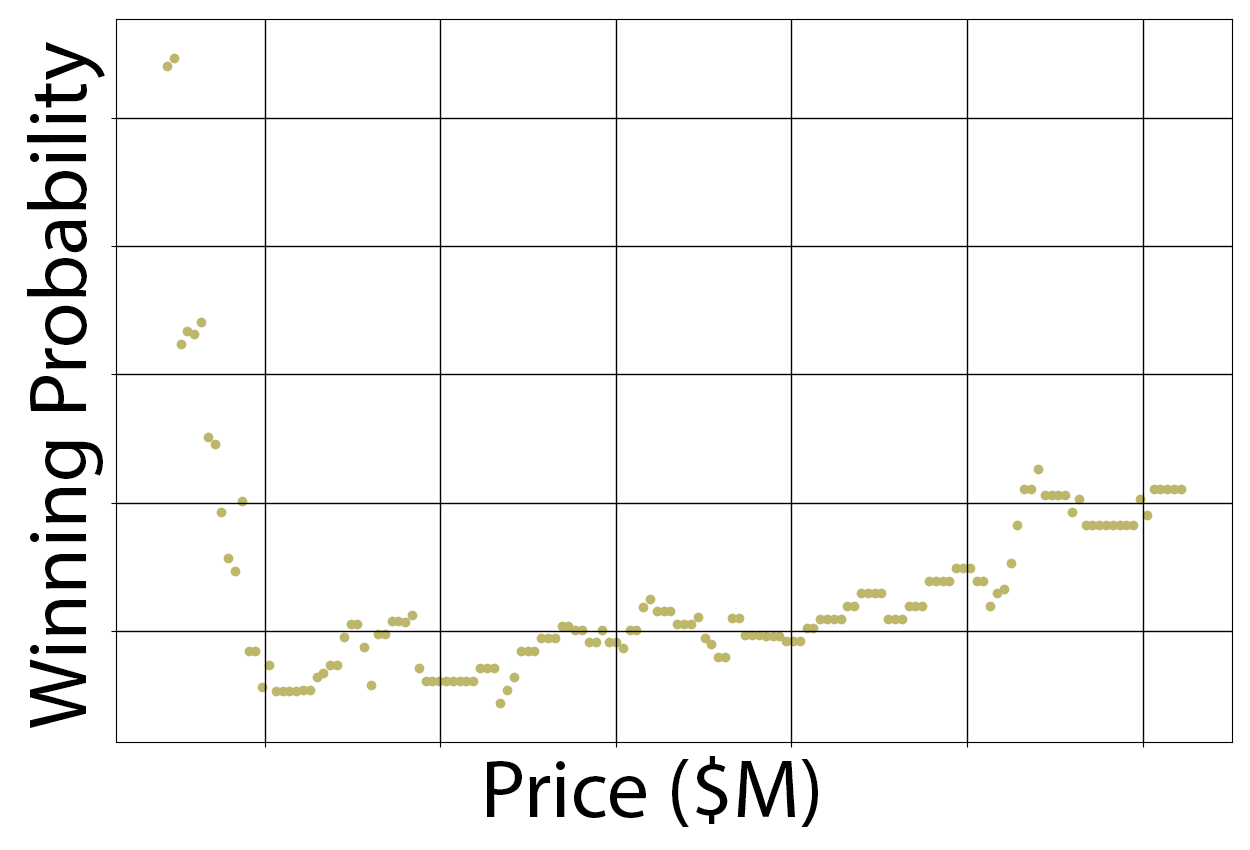}
    \caption{Deal 2}
  \end{subfigure}
  \caption{Examples of conflicts between sales experts' \textcolor{black}{judgment} and \textcolor{black}{ML} model's predictions. (Best view in color)}
  \label{fig:example}
\end{figure}

\textcolor{black}{Machine learning} (ML) models have shown a lot of success and usefulness in a lot of applications. However, the output of these models sometimes conflicts with experts' \textcolor{black}{judgment}(s) and/or does not correspond to it \citep{dietvorst2018overcoming,yu2019trust, nourani2020role,ahn2009conflict,d2019modeling,d2020conflicts,stewart2017label,jiang2018improved}. We define an expert's \textcolor{black}{judgment} as a domain/subject matter expert's insight that is neither formally defined nor statistically explained by available data (e.g., a known law of physics that cannot be learned from the collected data \citep{stewart2017label}, clinical experience about symptoms of patients \citep{jiang2018improved}, or a particular relationship between the type of soil and the resulting strength of a building based on a civil engineer's experience not captured in available training data). So, it is not something that can be captured statistically from the available training data, but it is also not a random guess from the expert; it is rather based on the expert's domain expertise/experience in the field. As a result, even sophisticated \textcolor{black}{ML} models might not capture such expert \textcolor{black}{judgment}. This paper aims at resolving this conflict between \textcolor{black}{ML} models and experts' \textcolor{black}{judgment}.	
To motivate this work, we use the following real-world example: In tendering processes, different providers submit their bids, trying to win contracts with clients. The client chooses the provider/competitor that suits them best, in terms of price and other features of the bid presented by that provider. In most of these cases, including the IT services tendering process we briefly discuss here and present in more detail in Section \ref{CaseStudy}, experienced salesmen at the providers' side would argue that the higher the bidding price for a competitor, the lower their chances are for winning a deal with a client. To predict the winnability of a deal (i.e., whether a deal will be won or lost to a competitor) at a given price point (and other features of such deal such as the client geography, prior business relationship between the provider and the client, etc), one can build \textcolor{black}{supervised ML} classification models trained on a fixed set of historical deals (See, e.g., \citep{megahed2015modeling,guo2019winnability,megahed2020analytics}). Using real-data from one of the biggest IT service providers in the world, we built such classification models and then plotted different pricing points for the same deal versus the probability of winning such deal as outputted by the classification model. Figure \ref{fig:example} presents such plot for two different deals. One can see in both plots that the relationship between the price and the probability of winning follows/agrees with the experts' (salesmen's) \textcolor{black}{judgment} in the beginning, but then a random or almost opposite relationship trend can be observed for larger prices. The reason for this is that such higher prices were not represented in the training data of historical deals, that were used to train the aforementioned ML classification model. Additionally, using traditional classification model evaluation metrics, the model had an accuracy (and precision and recall) over 90\% for out-of-sample data. However, obviously, sales executives will not trust such model when it qualitatively (looking at the figure) does not agree with the experts' \textcolor{black}{judgment} and will not make sense to them, even though its evaluation metrics' values are high. Also, one cannot resort to experts alone to do such predictions, because experts cannot give accurate predictions of specific deals, given the complexity of the multiple features that are involved. In this paper, we present a framework that yields results that are consistently much closer to the expert's \textcolor{black}{judgment} and with practically low sacrifice in the accuracy metrics. 
In many cases, we observe the sacrifice of the prediction performance when \textcolor{black}{judgment} (or constraint) is considered (e.g., \citep{bose2019compositional,rahman2019fairwalk}), but it is also important to make the model more trustable and transparent with respect to the \textcolor{black}{judgment}. In this paper, we defined a new harmonic measure to take into account both its predictive performance and the correspondence with the experts' \textcolor{black}{judgment}.

Therefore, we can define two research questions here: a) how can we build a model that resolves that conflict between experts' \textcolor{black}{judgment} and \textcolor{black}{ML} models and complements the deficiency of such \textcolor{black}{ML} models by augmenting them (and their high accuracy) with experts' input, so that one can achieve the best of both worlds, and b) since accuracy (and precision, recall, as well as other traditional metrics) does not seem to best evaluate models fall within this conflict, how can we come up with new metrics that give a better assessment of the quality of prediction models in these cases, by combining both the quantitative traditional accuracy measures with the quantitative ones (i.e., closeness to the experts' \textcolor{black}{judgment})? 

To mitigate the burden, many existing approaches have attempted to incorporate experts' \textcolor{black}{judgment}, based on supervised learning algorithms with additional regularization \citep{mann2007simple,druck2008learning}, weak supervision \citep{stewart2017label}, or model-specific constraint-based learning \citep{ben1995monotonicity,sill1998monotonic,duivesteijn2008nearest}. However, when there is not enough training data and/or the training data is biased, such \textcolor{black}{ML} models are still not reliable for predicting the unobserved world. One common solution to this problem is to collect more labeled data. The recent success of deep learning \citep{goodfellow2016deep,mikolov2013distributed,brown2020language} has also made the preparation of a large/sufficient amount of training data necessary to get performance gains. Nevertheless, collecting more data may not be practically possible in some business scenarios like our motivating example in which it is impractical to collect more labeled data within a short time as the data itself solely depends on business progress. Another example can be seen in additive manufacturing where one needs to predict whether a particular parameter setting on a machine might yield a defective product or not \citep{tapia2016prediction}. Evaluating/labeling the quality of a product can only happen via a destructive testing in which the product is cut such that any existing material defects can be seen under a microscope. For that reason, collecting enough data would be costly and therefore, impractical/time-consuming as well \citep{mahmoudi2019layerwise}. Domain experts have \textcolor{black}{judgment} about the relationship between some of the parameters and whether corresponding manufactured products are defective or not.




In this paper, we aim at answering the two aforementioned research questions by developing a new ML framework that we describe as follows: First, we estimate the reliability of using the learned \textcolor{black}{supervised ML} model for a given new unlabeled data instance. That is, we assess whether that unlabeled data point is included/ represented in the training data used to train the \textcolor{black}{ML} model. We achieve this using a generative adversarial network (GAN) model that we propose. Second, we calculate the required prediction, where if the unlabeled data point turned out to be similar to the training data, then the \textcolor{black}{ML} model is reliable and can be used to do the prediction. Otherwise, we rely on the expert \textcolor{black}{judgment}, or more generally, something in between (i.e., a weighted prediction using both the \textcolor{black}{ML} models and the expert's \textcolor{black}{judgment}). In other words, we leverage both the \textcolor{black}{ML} model and the expert's \textcolor{black}{judgment}, where the closer the new unlabeled data instance to the training data, the higher the weight we put on the former, and vice-versa.



To evaluate our framework, we initially generated four different types of synthetic datasets to test the robustness of our approach compared to two baseline cases: a) using ML models alone, and b) relying solely on experts' \textcolor{black}{judgment}. In addition, the framework was also applied to two real-world case studies: a B2B IT services scenario used above to motivate our problem and another one from the finance industry. All experiments show significant improvements of our model compared to the two baseline cases based on the performance metrics we developed and also qualitatively when plotting the results.

The rest of this paper is organized as follows: We review the relevant literature in Section \ref{LitReview}, detail our methodology in Section \ref{Methodology}, present our results on synthetic data in Section \ref{Experiments} and those on real-world data in Section \ref{CaseStudy}, and conclude the paper in Section \ref{Conclusions}.

\section{Related Work} \label{LitReview}
There are three relevant literature streams to our work. We review each of them in the below subsections.

\subsection{Constraint-based Machine Learning (CML)}
\label{sec:cml}

Relevant prior works in this stream have been mostly focused on exploiting high-level \textcolor{black}{judgment} in the form of constraints to be added to the ML models. Examples of constraints that have been used in Bayesian networks \citep{uai2005mono,niculescu2006bayesian,jiang2018improved} include monotonic assumptions, volume-wise dependency, and equalities/inequalities among parameters. Other examples can be found in classification trees \citep{ben1995monotonicity,feelders2003pruning,potharst2000decision}, neural networks \citep{archer1993application,sill1998monotonic}, gradient boosting trees \citep{israeli2019constraint}, nearest-neighbor classifiers \citep{duivesteijn2008nearest}, clustering \citep{zhi2013clustering}, and pattern decomposition \citep{ermon2015const}, where they all aimed at constructing theoretic frameworks or modifying the existing learning procedures to satisfy the \textcolor{black}{judgment} for their specific model families. However, they are only applicable to some ML models and the form/structure of that constraint is relatively simpler. Other streams of research use experts' \textcolor{black}{judgment} to reduce the burden for individual labeling. For instance, the authors in \citep{dietterich1997solving,zhou2007relation} attempted to efficiently label data by providing annotations over groups of images and learning to predict properties that hold over at least one input in a group. However, it is not possible to incorporate specific relationships between variables and labels using their methods.

The most relevant works that incorporate inputs of ``experts" (either real domain experts or latent Dirichlet allocation-like annotators) are regularization-based approaches \citep{grandvalet2005semi,mann2007simple,druck2008learning}. For example, Entropy minimization \citep{grandvalet2005semi} introduces an additional term that minimizes the entropy of the label distribution on the unlabeled data, in addition to a traditional conditional label log-likelihood objective function. In addition, expectation regularization \citep{mann2007simple} learns the label distribution by leveraging a Kullback-Leibler divergence-based regularization term. However, they only allow learning such label distribution, which is also a simpler form of experts' \textcolor{black}{judgment}. Feature feedback is also a type of experts' inputs in this context. For example, a probabilistic disjunction model \citep{poulis2017learning} is proposed to incorporate information (e.g., particular sets of features) that is relevant to labels. This paper provides computationally tractable regularization and bootstrapping-based approaches. However, the authors assume that data follows a simple generative model, so other complex functional relationships could not be learned.

Other recent works have attempted to avoid additional feature engineering and labeling by integrating \textcolor{black}{judgment} functions with the loss functions of neural networks to predict sentiments of text sentences \citep{kotzias2015group}, perform multi-label classification of images \citep{lin2016learning,zhuang2016fast}, or incorporate laws of physics \citep{stewart2017label} where the loss function is meant to penalize outputs that are not consistent with these laws and relational constraints. However, the underlying approaches are specific to a target model (e.g., Convolutional Neural Networks) or still almost entirely rely on the training data.

\subsection{Human-in-the-loop (HITL) in Machine Learning}

Human-in-the-loop (HITL) is the process of leveraging the power of both the machine and human intelligence to enhance machine learning-based artificial intelligence (AI) models. To address this goal, HITL ML systems have been proposed to support the entire ML modeling process. This stream of literature can be divided into two parts: HITL for model building and HITL for model management.

In the first category of HITL, human intervention or participation within HITL could be used for supporting model building \citep{zhang2016materialization, sparks2015automating, singh2014designing,vartak2015supporting}. Building a real-world ML model is an iterative process. Starting with one or more initial hypotheses, data scientists build models to verify their hypotheses, refine them, and repeat this process until the models meet some acceptance criteria (e.g., a certain accuracy). To support their needs, for example, \cite{vartak2015supporting} discussed requirements in the stage of model building and proposed a user interface. The interface supports incremental training/reuse and history tracking for human experts. However, their systems are limited only to preparing better training data, choosing models, or selecting better hyper-parameters, rather than correcting predictions. Meanwhile, active learning \citep{wang2017active, tong2001support, settles2011theories, park2019active, cao2020divide, trittenbach2020overview} also helps model builders develop strategies for identifying unlabeled items that are near a decision boundary in the current ML model or finding unlabeled items that are unknown to the ML model. Its learning process is also iterative. In each iteration of active learning, the model is re-trained with the new items, and the process is repeated. While it could incorporate feature-level inputs from experts, it is sensitive to the current ML model and the available training data. In addition, specific relationships between variables and labels are hard to learn.

In the second category of HITL, HITL could be approached in the view of model management \citep{modeldb,miao2017towards,miao2017model, van2017versioning}. Model management handles the storage and re-use of large numbers of ML models for future sharing and analysis. One of the previous attempts is ModelDB \citep{modeldb}, which can track ML models in their native environment, index them intelligently, and allow flexible exploration of models via SQL in an end-user interface. While the user interface provides us a way to visually explore modeling pipelines, it is not able to incorporate humans' input(s) to correct the model's predictions. Similarly, \cite{miao2017towards} also proposed a life-cycle management platform for deep learning models, which can minimize storage footprints and query workloads. They are also limited when it comes to correcting and improving predictions.

\subsection{Out-of-distribution (OOD) Detection and Transfer Learning (TL)}

The predictive uncertainty of ML models is closely related to the problem of how much potential abnormal samples are far away from in-distribution (i.e., distribution of training samples) statistically. For detecting out-of-distribution (OOD) samples, recent works have exploited the confidence from the posterior distribution \citep{hendrycks2016baseline, liang2017enhancing,feinman2017detecting}. For example, \cite{hendrycks2016baseline} utilized the maximum probability from the softmax label distribution in the neural network classifier as a baseline method, and the temperature scaling with input pre-processing \citep{liang2017enhancing} is also used for that purpose. However, it is not possible for them to correct the prediction of the ML models and make them correspond to experts' \textcolor{black}{judgment}, rather than detecting out-of-distribution samples.

\textcolor{black}{
Moreover, a DRC measure \citep{drc2,drc1} has proposed to quantify how similar training data is to unseen data. It provides a clue about when the performance of the existing supervised classifier decreases. When the DRC value to a set of unseen data is larger than 1, it indicates that the training and unseen data are not exchangeable and the existing supervised model will under-perform when applied to the unseen data. However, it is also not able to incorporate the judgment of experts and could not evaluate how much the trained model is robust with respect to a set of specific judgment variables. In addition, \cite{cabitza2020if} proposed a method to measure the quality of the labeling for understanding the output of future predictive models. Authors of \cite{cabitza2020if} attempted to address the following problems: (1) how to measure the quality of a training data set with respect to the reference population?, (2) do the raters agree with each other in their ratings?, and (3) are the raters’ annotations a true representation?. They proposed new metrics to quantify the quality of labeled data, but they are also limited in measuring how to incorporate domain experts' judgment into a predictive model.  In particular, it is also necessary to have a reference population or agreement information among raters, which makes it not applicable to our research problem.
}

Transfer learning \citep{pan2009survey,weiss2016survey,salaken2019seeded} is another related research stream. Its goal is to improve the predictive performance on a target domain (or task) by leveraging the model trained via a source domain with its data. For example, \cite{luo2017label} proposed a framework that learns representations transferable across different domains and tasks. However, the works on transfer learning have focused on re-purposing the pre-trained model for a new task or a new domain. Therefore, it is also not possible to incorporate experts' \textcolor{black}{judgment}/input and correct predictions using this methodology.

\subsection{Data Imputation (DI)}
\label{sec:di}
\textcolor{black}{In many cases, data is often missing, and the missing data may result in wrong predictions. Missing values in a dataset may significantly increase computational cost, skew the outcome, and describe the performance of the predictive model. To mitigate the burden, many works \citep{mice,lan2020multivariable,purwar2015hybrid,feng2021imputation} have proposed to estimate the missing data. For example, MICE \citep{mice} assumes that the probability of a missing variable depends on the observed data. It filled at random values to the missing data positions in the beginning and predicts them by leveraging a series of regression models. Similarly, \cite{lan2020multivariable} leverages a bayesian network for iterative imputation, and \cite{purwar2015hybrid} uses the combination of K-means clustering with Multilayer Perceptron to address the data imputation problem. Recently, \cite{feng2021imputation} shows that random forest models can enhance the data imputation problem and quantify the uncertainty of the prediction. While the approaches above show promising results to estimate the missing data, it is still difficult to model the underlying complex non-linear relationship between variables. To mitigate the problem, we propose a neural network-based generative model to exploit the judgment variable using other observations. Generative adversarial networks (GAN) \citep{gan} is a neural network-based generative model, which has shown superior performance in synthesizing images or transforming images. Inspired by the success of Generative Adversarial Networks (GAN) in image generation, we proposed a \mname{} to learn the overall distribution between a judgment variable and others with GAN, which is further used to measure the potential conflict between experts and predictive model. We also include experimental results including MICE with KNN and Random forest in Section \ref{sec:ablation} and demonstrate the benefits of our model in our real-world case studies.}

\subsection{Categorization and Discussion}

\begin{table*}[!t]
    \centering
	\caption{\textcolor{black}{Summarization of representative methods in related work}}
	\label{tab:summ:comparison}
	    \begin{minipage}[!t]{.99\textwidth }
	        \centering    
	        \scalebox{0.53}{  
				\begin{tabular}{cccccc}
				Category            & Method & Expert Input         & Target Model              & \begin{tabular}{@{}c@{}}Incorporating \\Method \end{tabular}             & \begin{tabular}{@{}c@{}}Judgment Variable \\Estimator \end{tabular} \\ \hline
				CML &  \begin{tabular}{@{}c@{}}Constraint-based bayesian network \\ \citep{jiang2018improved} \end{tabular}   & Variable-wise assumption & \begin{tabular}{@{}c@{}}A set of \\specific models \end{tabular}             &  Correlation  & $\times$           \\ \hline
				CML &   \begin{tabular}{@{}c@{}}Expectation regularization \\\citep{mann2007simple}  \end{tabular}   & Regularization function & \begin{tabular}{@{}c@{}}Most of \\ supervised models  \end{tabular}    &  Regularization & $\times$                         \\				 \hline
				CML & \begin{tabular}{@{}c@{}}Weak Supervision \\ \citep{stewart2017label}  \end{tabular}   & Supervision function & \begin{tabular}{@{}c@{}}Most of \\ supervised models  \end{tabular}  &      Joint learning  &  $\times$                           \\ \hline
				HITL                 &  \begin{tabular}{@{}c@{}} ModelHub \\\citep{miao2017towards}  \end{tabular}   & Human intervention     & \begin{tabular}{@{}c@{}}A set of \\specific models \end{tabular}     & \begin{tabular}{@{}c@{}} Domain-specific\\ query \end{tabular}  & $\times$                   \\ \hline
				OOD                 & DRC~\citep{drc2}         & Base distribution      & \begin{tabular}{@{}c@{}}Most of \\ supervised models  \end{tabular}  & $\times$    & $\times$                   \\ \hline
				TL   &    \citep{salaken2019seeded}         & $\times$   & $\times$             & \begin{tabular}{@{}c@{}} Seeded \\knowledge transfer   \end{tabular}                        & $\times$             \\ \hline
				DI     &    \citep{feng2021imputation} & $\times$    & $\times$    & $\times$                       & Random forest    \\ \hline 
				          -          & \textbf{Our Method }  &  Judgment function     & \begin{tabular}{@{}c@{}}Most of \\ supervised models  \end{tabular}  & Conflict Resolution              & \mname{}                         
				\end{tabular}
				}
			    \hrule height 0pt
			    \end{minipage}%
\end{table*}

\textcolor{black}{
Table \ref{tab:summ:comparison} compares the properties of ML models that use expert inputs, considering the nature of the inputs, target models, incorporating methods, and estimators of the judgment variables. For the comparison, the representative methods are selected from our categorizations following section \ref{sec:cml} and section \ref{sec:di}.
First, many constraint-based ML (CML) models are mostly designed for specific models (e.g., \citep{jiang2018improved}). They have shown high accuracy in the corresponding models, but it is difficult to use diverse expert inputs and they may not be robust to instances that are not known from the training data. Expectation regularization (ER) \citep{mann2007simple} takes the regularization function as expert input and could be applied to most supervised models that use gradient descent updates. Similarly, weak supervision (WS) models like \citep{stewart2017label} take supervision functions, which explain a law in physics and show performance gains in closeness. In contrast, the ER and WS models are not able to leverage how reliable the predictions of new inputs are for the learned model. Meanwhile, Human-in-the-loop (HITL) methods such as \citep{miao2017towards} also attempt to utilize constraints through human interactions.  Nevertheless, they primarily focus on collecting high-quality training data, reusing models, and managing the existing models, which still much depend on the prediction of the learned ML models. For Out-of-distribution (OOD) models like DRC~\citep{drc2}, they take the simple statistical base distribution as expert input and measure how training data is different from the new instances. However, there are no incorporating methods to correct the predictions of the existing model. Transfer Learning (TL) methods are also limited in modifying the predictions for exploiting the expert input. Many Data Imputation (DI) techniques using MICE (e.g., \citep{feng2021imputation}) can estimate missing judgment variables from the known data, but they have not been applied to correct the predictions of the current models.}

\textcolor{black}{
Different from all these previous methods, we aim at exploiting situations in which the experts' judgment disagrees with the ML models, where we resolve the conflict(s). We find these conflicts by examining how much new unlabeled inputs are relevant/reliable/presented in terms of the ML model's expectation. Then, we develop a method to subsequently correct these ML outputs.}

\section{Methodology} \label{Methodology}
We describe our problem setup and overall framework to resolve the potential conflict between experts' \textcolor{black}{judgment} and a \textcolor{black}{ML} model in Section \ref{sec:problemsetup}. In this framework, we estimate the reliability of the \textcolor{black}{ML} model w.r.t. the \textcolor{black}{judgment} variable by leveraging a proposed generative adversarial network, which we detail in Section \ref{sec:eigan}. Lastly, our proposed evaluation metrics are described in Section \ref{sec:evaluationmeasure}.	
\subsection{Problem Setup and Our Framework to Correct \textcolor{black}{ML} Model Predictions} \label{sec:problemsetup}	
Given a set of $n$ examples that have the form,  $ \{(x_1, y_1), \ldots, (x_n, y_n)\}$ such that $x_i$ is the features vector of the $i^{th}$ data point and $Y_i$ is its label, a \textcolor{black}{ML} classification model learns a function $f: X \rightarrow Y$ that maps the input space $X$ to the output space $Y$. 	
In addition, experts have their own \textcolor{black}{judgment} function $g: Z \rightarrow Y$, where $Z$ is the \textcolor{black}{judgment} feature(s) and so $Z\subset X$. In this paper, we assume $Z$ is single feature/variable for simplicity. For example, in the above IT service deal motivating example, $Z$ could be a price and $Y$ could be winning probability. Note that in many real-world applications, particularly when AI systems (or experts) communicate with customers, interpretations using a single or a small subset of variables are often preferred \citep{rudin2019stop}. It is also worth noting that the correctness of the expert \textcolor{black}{judgment} is out of the scope for this work. Further, we divide $X$ into two subsets, $X_{train}$ and $X_{test}$, for training and testing, respectively. Similarly, $Y_{train}$ and $Y_{test}$ are their corresponding label sets. For any data point $x_i\in X_{test}$, we denote the prediction of its label via the \textcolor{black}{ML} model by $\hat{y}_i$. Let the \textcolor{black}{judgment} variable inside among the variables in $x_i$ be $z_i$. The value of the \textcolor{black}{judgment} function at $z_i$, $g(z_i)$ is the prediction according to the expert's \textcolor{black}{judgment}.


Our objective is to correct the prediction of the \textcolor{black}{ML} model at $x_i$ by considering the expert \textcolor{black}{judgment}, as well as develop new performance metrics that incorporate the closeness to the expert's \textcolor{black}{judgment} with traditional accuracy metrics of ML classification models. To achieve this goal, we first take the input features of $x_i$ after taking out the \textcolor{black}{judgment} variable $z_i$ from it, and develop a generative adversarial network (GAN)-based neural model, which we call \mname{} (Experts' \textcolor{black}{judgment} GAN), to predict what the value of that \textcolor{black}{judgment} feature would have been according to the training data. We denote that prediction by $z_i^{expected}$.  \textcolor{black}{Generative adversarial networks (GAN) \citep{gan} is a game-theoretic neural learning framework for generating realistic data instances. In this paper, the two-player zero-sum minimax game using a generator and a discriminator is leveraged to learn the overall distribution between a judgment variable and others. Our proposing \mname{} is used to measure the potential conflict between experts and the predictive model. In particular, if the data is observed on a biased way and the complex decision-making process of many people is involved, the generative approach is expected to be more effective.} We discuss more details of \mname{} in Section \ref{sec:eigan}. 

Then, we calculate the absolute difference between $z_i^{expected}$ and $z_i$ and denote that by $k$. Our corrected prediction $y$ is a the sum of $g(z_i)$ and $\hat{y}_i$ after multiplying the former with $w$ and the latter with $(1-w)$, where $w$ is a weight we learn based on $k$ such that the higher the value of $k$, the further away $z_i$ is from our training data and thus, the higher the weight $w$ we put on the expert's \textcolor{black}{judgment}, and vice-versa. The function to determine the weight could vary depending on the reliability of the expert's \textcolor{black}{judgment}, but we generalize its form using the sigmoid function and formulate it as follows:
 \begin{equation}
	 \begin{aligned}
 		w &= Sigmoid(k, \alpha)= ((\frac{1}{1+e^{-\alpha \cdot k}}) -0.5) \cdot 2,
 	\end{aligned}
 \end{equation}
\noindent where $\alpha$ is a parameter that we learn its optimal value that maximizes the new performance metric, that we develop in Section \ref{sec:evaluationmeasure} below, for our prediction method over the testing/validation dataset. The optimal $\alpha$ can then be used when we apply our method to unlabeled data.





Algorithm \ref{alg:framework} illustrates our method. The algorithm takes an input instance $x_i \in X_{test}$, the prediction of a \textcolor{black}{ML} model, $\hat{y}_i$, the expert \textcolor{black}{judgment} function $g$, and pre-trained \mname{}. It then estimates $z_i^{expected}$ by applying the  \mname{} to $x_i$ after taking $z_i$ out of it (Line 1). The absolute difference between $z_i^{expected}$ and $z_i$ is then calculated (Line 2) and used to compute a weight $w$ using a sigmoid function of that $k$ and a parameter $\alpha$ that is optimized as discussed above (Line 3). Lastly, the weight is used for deciding how the \textcolor{black}{ML} model's prediction is combined with the expert's \textcolor{black}{judgment} to estimate our final prediction (Line 4). That prediction is our final returned value (Line 5). We next detail how the \mname{} step works.

 \begin{algorithm}[tb]
 \SetAlgoLined
  \KwInput{An input data point $x_i \in X_{test}$, including the \textcolor{black}{judgment} variable $z_i \in x_i$, and a \textcolor{black}{ML} model's prediction, $\hat{y}_i=f(x_i)$\, an expert's \textcolor{black}{judgment} prediction, $g(z_i)$, a pre-trained \mname{};} 
  
	$z_i^{expected} \gets \mname{}(x_i / z_i)$ \;
    $k \gets ||z_i^{expected} - z_i||$ \;
    $w \gets Sigmoid(k, \alpha)$ \;
	$\hat{y}_{final} \gets w \cdot g(z_i) + (1-w) \cdot \hat{y}_i$ \;
	Return $\hat{y}_{final}$ \;
 \caption{Our method for resolving the conflict between \textcolor{black}{ML} models and experts' \textcolor{black}{judgment}}
 \label{alg:framework}
 
\end{algorithm}



\subsection{Architecture and Procedure to Learn \mname{}} \label{sec:eigan}

%
%


In this section, we illustrate the details of our proposed \mname{} and discuss how generators become specialized in estimating how much the \textcolor{black}{ML} model is trustworthy for a testing data point. The goal of our proposed \mname{} is to generate an estimation of the \textcolor{black}{judgment} variable for a data point, $z_i^{expected}$, based on the training data and given the rest of the features of that data point (i.e., the features of that testing data point minus the \textcolor{black}{judgment} feature, $x_i/z_i$, which we call $s_i$ in the rest of this paper). Let $G$ denote a generator which estimates that \textcolor{black}{judgment} valuable $z_i$. $G$ is formulated as follows:	
\begin{equation}	
	\begin{aligned}	
		z_i^{expected} = G(s_i)= ReLU(ReLU(W_0 (s_i) + b_0)W_1 + b_1),	
	\end{aligned}	
\end{equation}	
\noindent where $W_0$, $W_1$, $b_0$, and $b_1$ are trainable weight and bias matrices. The number of ReLU layers could vary depending on the dataset. 	
Meanwhile, a discriminative model $D$ is used to determine if the expected \textcolor{black}{judgment} variable with other feature vectors $x_i/z_i$ represents the given training data or not by iteratively using the real \textcolor{black}{judgment} variable $z_i$ with $x_i/z_i$. The discriminative model is described as follows:	
\begin{equation}	
	\begin{aligned}	
		&\hat{y}_i^{\text{\mname{}}} = D(z_i||s_i) \text{, when $D$ is learning} \text{ \ or} \\	
		&= D(z_i^{expected}||s_i) \text{, when $G$ \& $D$ are learning,}	
	\end{aligned}	
\end{equation}	
where $D$ is also implemented using ReLU layers with trainable weight and bias matrices. $\hat{y}_i^{\mname{}}$ represents how much the the inputs are realistic.	
The procedure to learn the $G$ and $D$ models is described in Algorithm \ref{alg:eigan}, where parameters of $G$ and $D$ are initialized (Line 1 and 2) and include weights and biases for their ReLU models. Then, the parameters are updated via the gradient update with respect to each of the learning variables on the given mini-batched examples (Line 3-7). As a result, the final $G$ will be used to determine $z_i^{expected}$ in Algorithm \ref{alg:framework} as we described above. We now end this section by detailing our new proposed evaluation metrics.	
 \begin{algorithm}[tb]	
 \SetAlgoLined	
 \algsetup{linenosize=\small}	
  \KwInput{Generative and Discriminative models, $G$ and $D$, respectively, and mini-batch size $m$\, \textcolor{black}{judgment} features, $Z_{train}$, and other features $S_{train}$ from $X_{train}$ for training\;}	
  	Initialize $\theta_G$ (Weights and Biases in $G$)\;	
  	Initialize $\theta_D$ (Weights and Biases in $D$)\;  	
  	\While {until converge}{	
  		S = $\leftarrow \{s_1, ..., s_m\}$, a mini-batch of real samples except \textcolor{black}{judgment} variables from $S_{train}$\;	
  		Z = $\leftarrow \{z_1, ..., z_m\}$, a mini-batch of real samples from \textcolor{black}{judgment} variables in $Z_{train}$\;	
    	Update $\theta_G$ via			
        \begin{align*}	
             &\nabla_{\theta_G} \frac{1}{m} \sum_{i=1}^{m} log(1-D(s_{i}, G(s_i))	
        \end{align*}	
		Update $\theta_D$ via	
        \begin{align*}	
             &\nabla_{\theta_D} \frac{1}{m} \sum_{i=1}^{m} -logD(s_i, z_i) - log(1-D(s_i, G(s_i)))	
        \end{align*}	
    }	
    return $G$\;	
 \caption{Procedure to learn the Generative and Discriminative models in \mname{}}	
 \label{alg:eigan}	
\end{algorithm} 	
 	
\subsection{Our New Proposed Evaluation Metrics} \label{sec:evaluationmeasure}	
Using prediction accuracy metrics might not be sufficient in cases like the one we study in this paper, as we have seen in the motivating example above. To address the problem, a new evaluation metric, $closeness$, is proposed to quantify how much an input ML model corresponds to experts' \textcolor{black}{judgment}. 	
We use Jensen-Shannon Divergence (JSD) \citep{manning1999foundations} to define the $closeness$ metric that measures how much the prediction corresponds to the experts' \textcolor{black}{judgment}. 	
\begin{equation}		
		closeness = JSD(\widetilde{f}(X_{test})||\widetilde{g}(X_{test})).
\label{eq:closeness}	
\end{equation}	
\noindent To get $\widetilde{f}(X_{test})$ and $\widetilde{g}(X_{test})$, we need to prepare $q$ number buckets first, which linearly divide possible values of the \textcolor{black}{judgment} variable into $q$ buckets. By the criteria of buckets, instances of $X_{test}$ are assigned accordingly to the buckets. Then, we can get $q$ number of averaged predictions by leveraging $f$ or $g$ on each of the datasets in the buckets. As a result, we have $\widetilde{f} \in \mathcal{R}^{|X_{test}| \cdot q}$ and $\widetilde{g}  \in \mathcal{R}^{|X_{test}| \cdot q}$, and the two 1-dimensional vectors are used for measuring the JSD. 	
On the other hand, solely using the $closeness$ that evaluates a prediction onto experts' \textcolor{black}{judgment} ignoring prediction accuracy would not be reasonable and might yield very inaccurate results. To mitigate this, a harmonic measure between $accuracy$ and $closeness$ metrics is defined as follows:	
 	
 \begin{equation}	
 	combined = \frac{2 \cdot accuracy \cdot closeness}{accuracy + closeness}
 \end{equation}	

\textcolor{black}{
\noindent The harmonic measure is a Schur-concave function \citep{varberg1973convex}, which is defined as the reciprocal of the arithmetic mean of the reciprocals. Therefore, it cannot be made arbitrarily large by having one of the values with high scores; it is not sensitive to extremely large values. Moreover, the harmonic mean is appropriate if the data values are ratios of two variables with different measures, and it has also shown more accurate adjusted averages in many different areas, such as F-score~\citep{taha2015metrics} in information retrieval, thin-lens equation~\citep{hecht1998optics} in optics, and semi-latus rectum~\citep{eves1963survey} in math.
}


\begin{figure*}[!t]
\centering
  \begin{subfigure}{1.1\textwidth}
    \includegraphics[trim=900 0 0 0,clip,width=\linewidth]{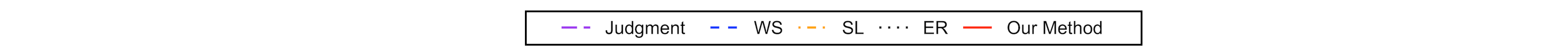}
  \end{subfigure}
  
  \begin{subfigure}{0.24\textwidth}
    \includegraphics[width=\linewidth]{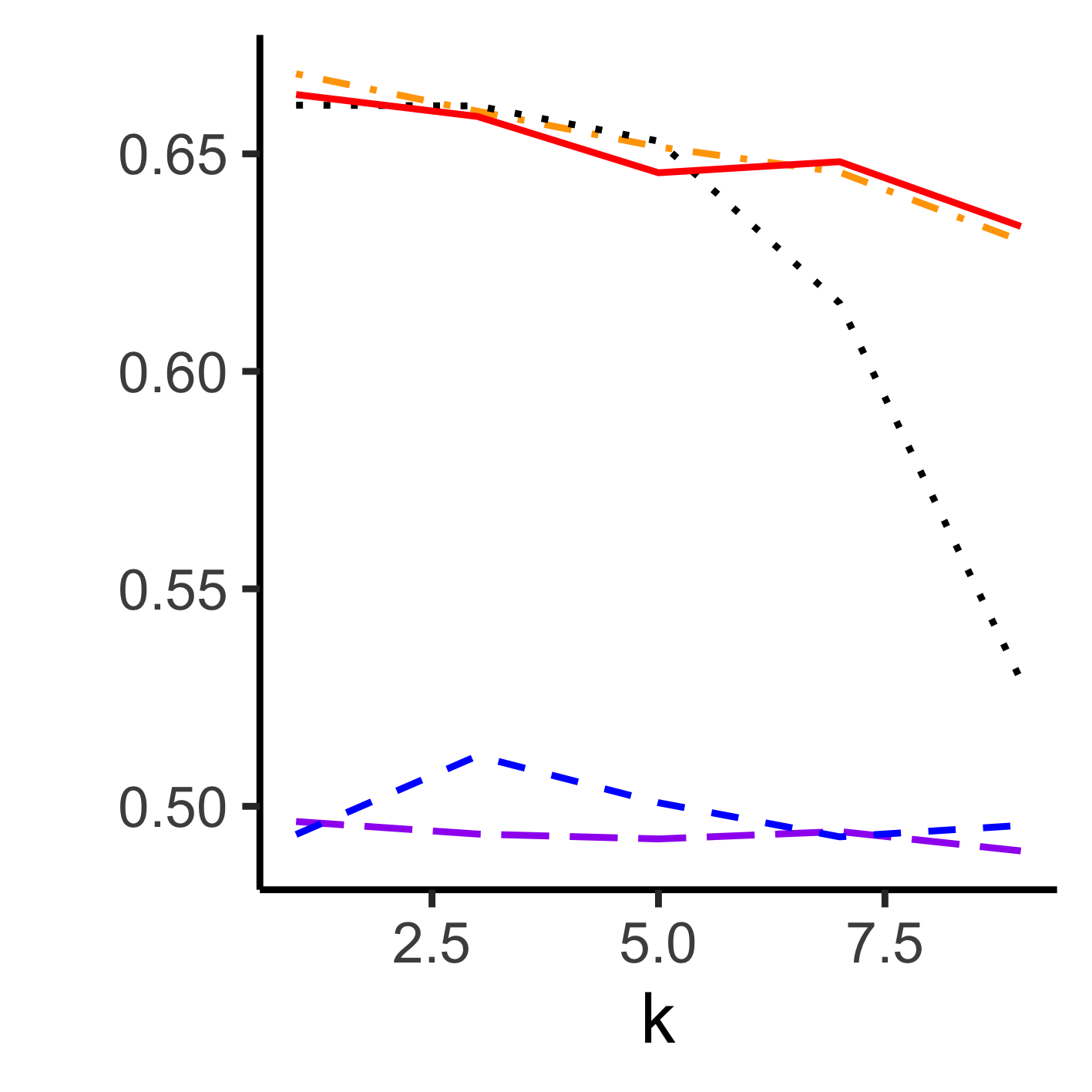}
    \caption{Sigmoid}
  \end{subfigure}
  \begin{subfigure}{0.24\textwidth}
    \includegraphics[width=\linewidth]{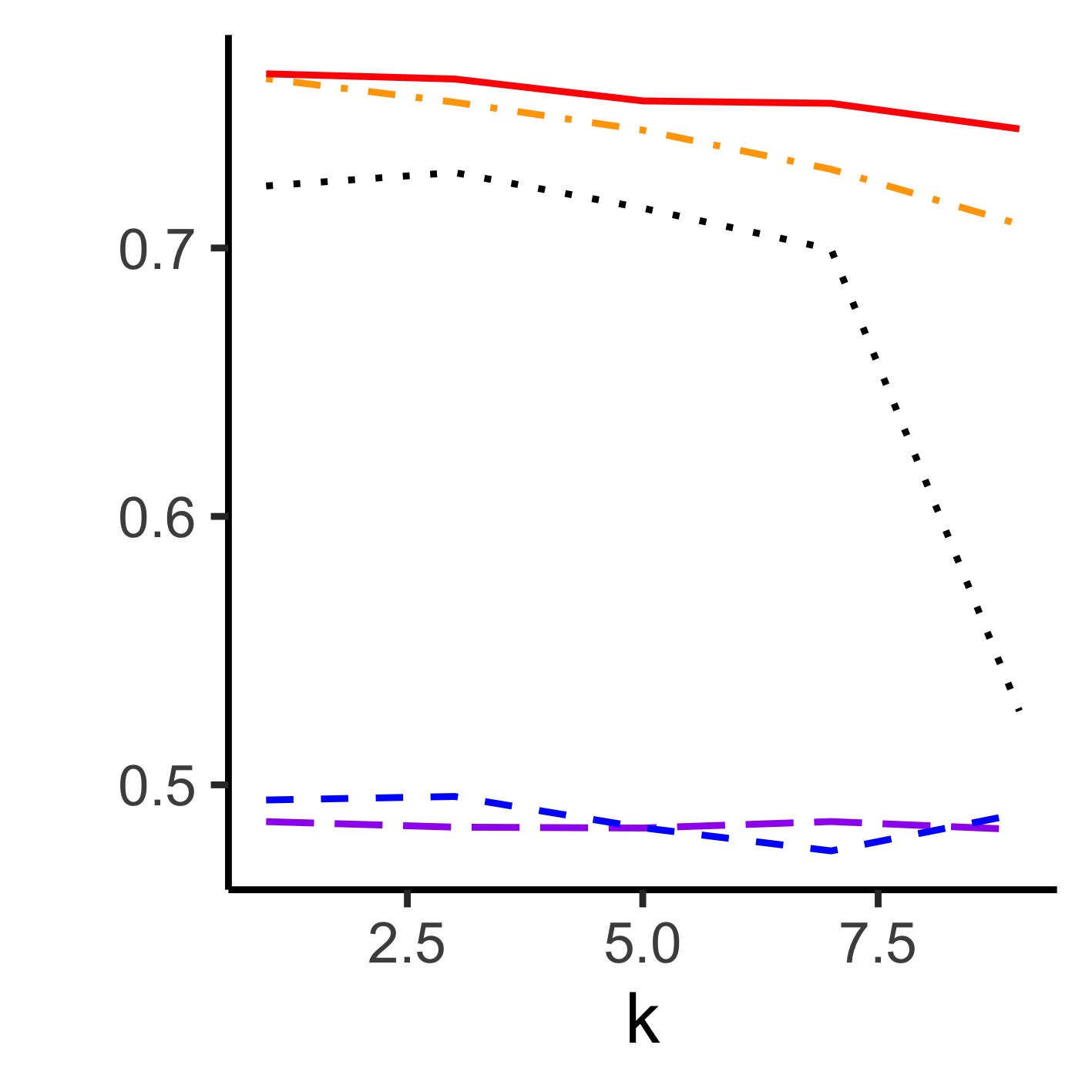}
    \caption{Exponent (Exp)}
  \end{subfigure}
  \begin{subfigure}{0.24\textwidth}
    \includegraphics[width=\linewidth]{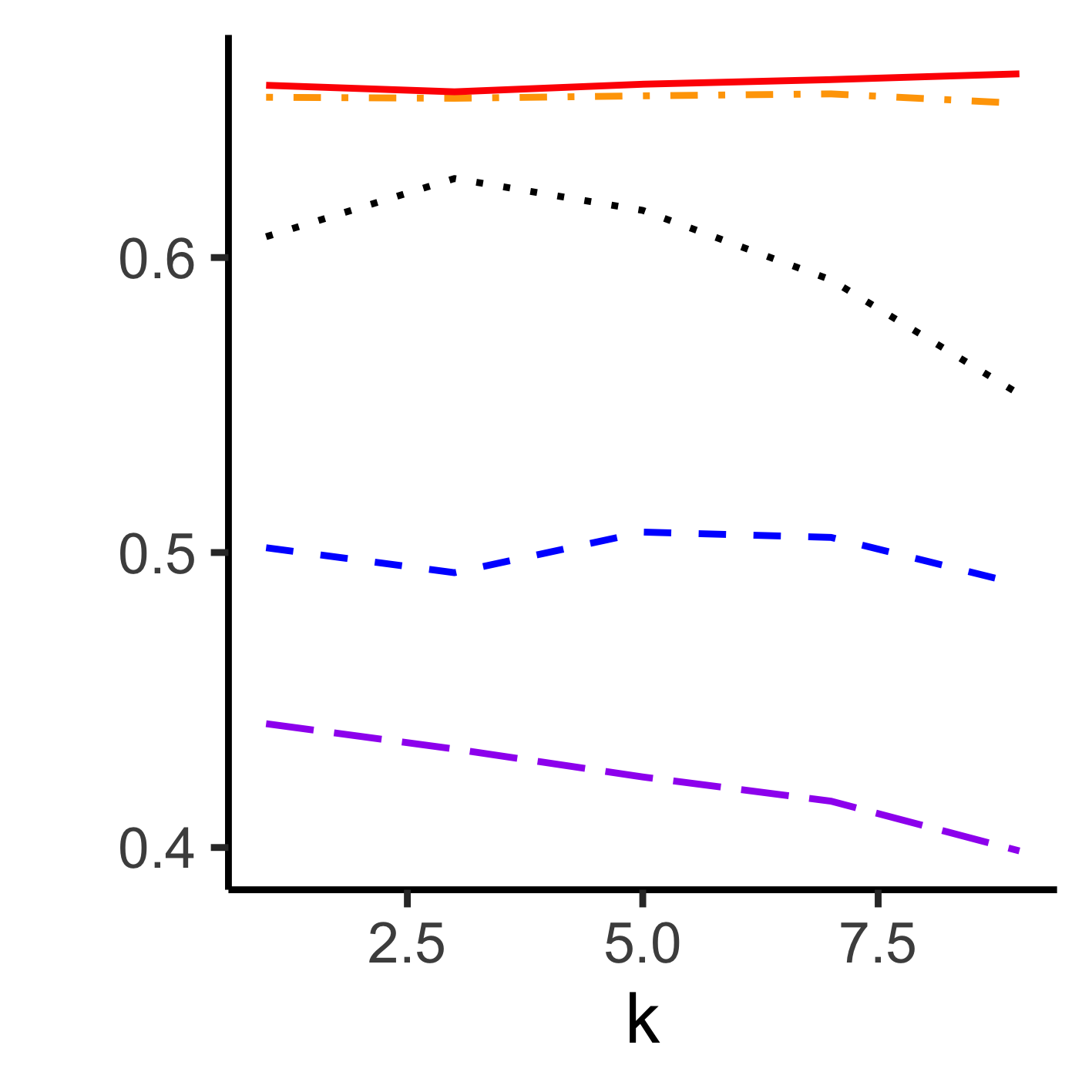}
    \caption{Exp w/ Squared}      
  \end{subfigure} 
  \begin{subfigure}{0.24\textwidth}
    \includegraphics[width=\linewidth]{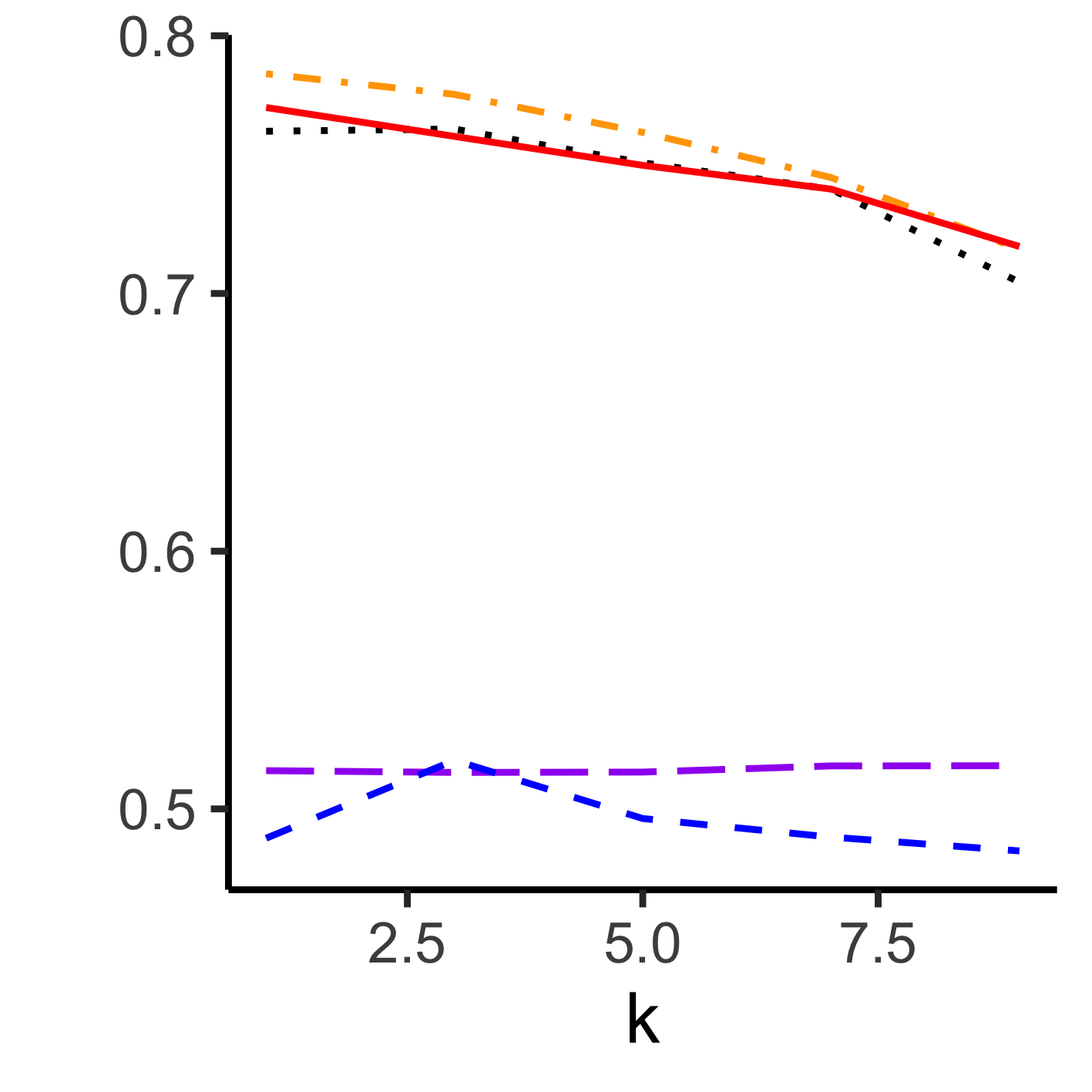}
    \caption{Exp w/ Two Weights}
  \end{subfigure}   
  \caption{Accuracy on the synthetic data as $k$ is varied. \textit{Gradient boosting} is used for the \textcolor{black}{target ML model}. (Best view in color)}
  \label{fig:syn:gb:acc}
\end{figure*}

\begin{figure*}[!t]
\centering
  \begin{subfigure}{0.24\textwidth}
    \includegraphics[width=\linewidth]{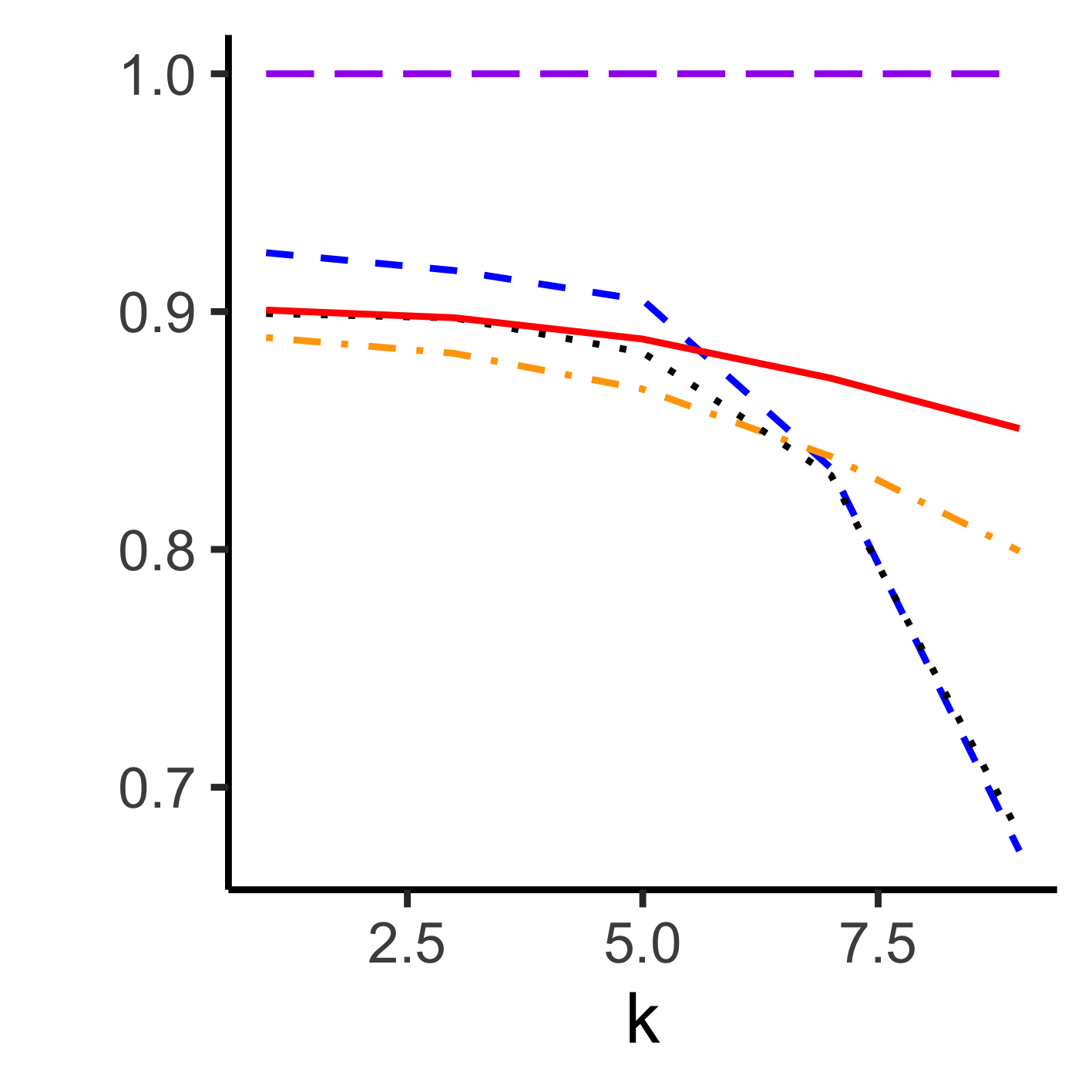}
    \caption{Sigmoid}
  \end{subfigure}
  \begin{subfigure}{0.24\textwidth}
    \includegraphics[width=\linewidth]{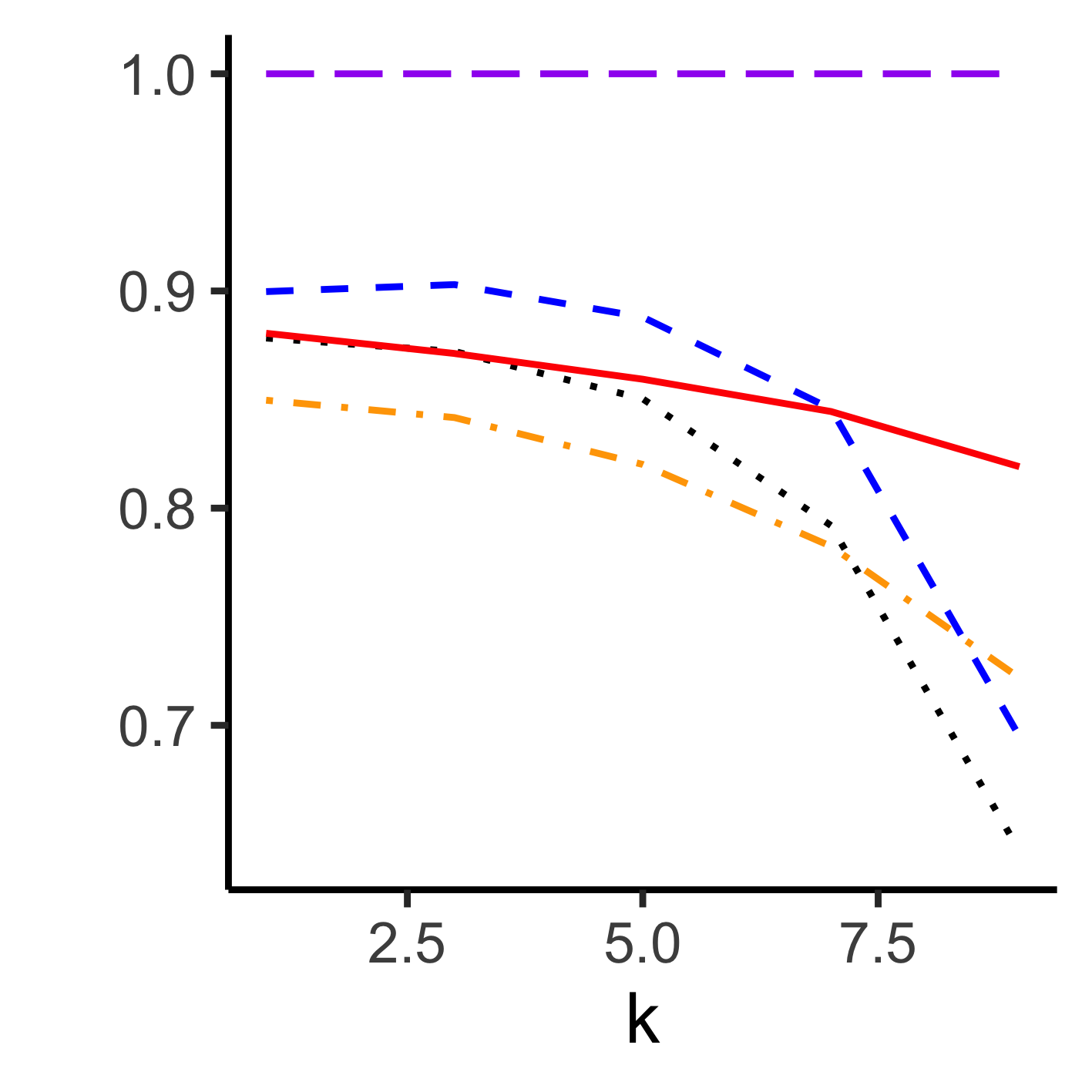}
    \caption{Exponent (Exp)}
  \end{subfigure}
  \begin{subfigure}{0.24\textwidth}
    \includegraphics[clip,width=\linewidth]{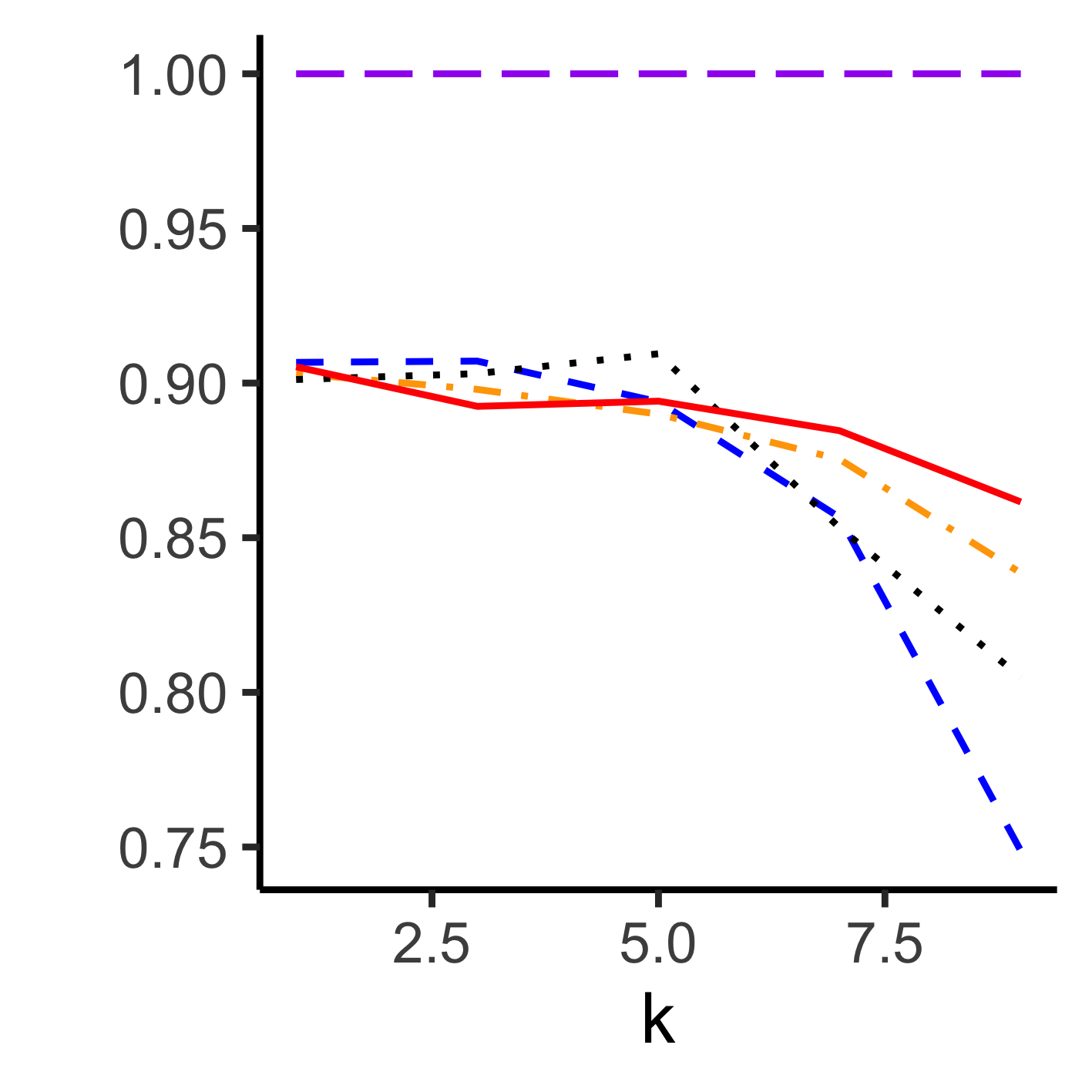}
    \caption{Exp w/ Squared}
  \end{subfigure} 
  \begin{subfigure}{0.24\textwidth}
    \includegraphics[clip,width=\linewidth]{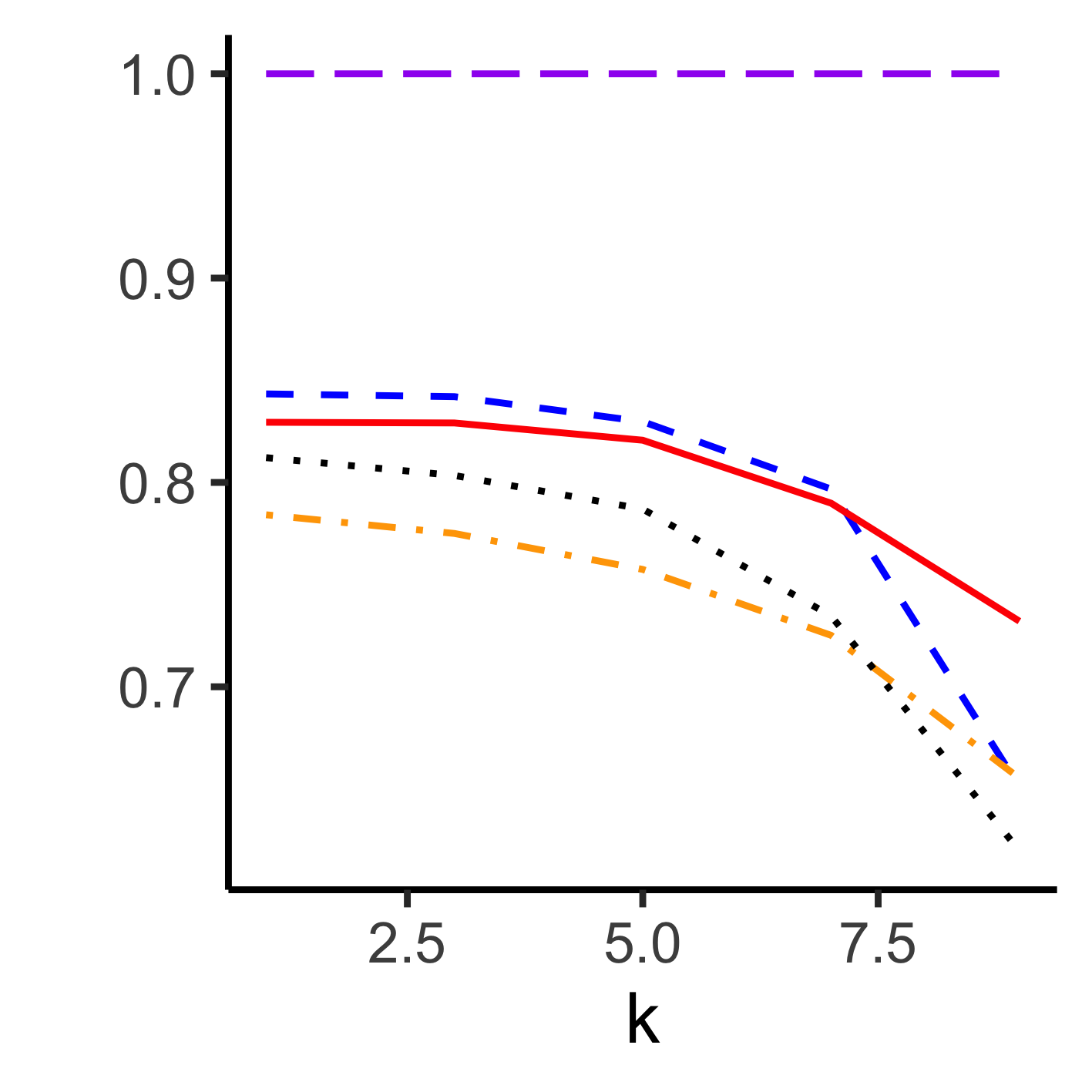}
    \caption{Exp w/ Two Weights}
  \end{subfigure}   
  \caption{\textit{Closeness} metric on the synthetic data as $k$ is varied. \textit{Gradient boosting} is used for the \textcolor{black}{target ML} function. (Best view in color)}
  \label{fig:syn:gb:clo}
\end{figure*}

\begin{figure*}[!t]
\centering
  \begin{subfigure}{0.24\textwidth}
    \includegraphics[width=\linewidth]{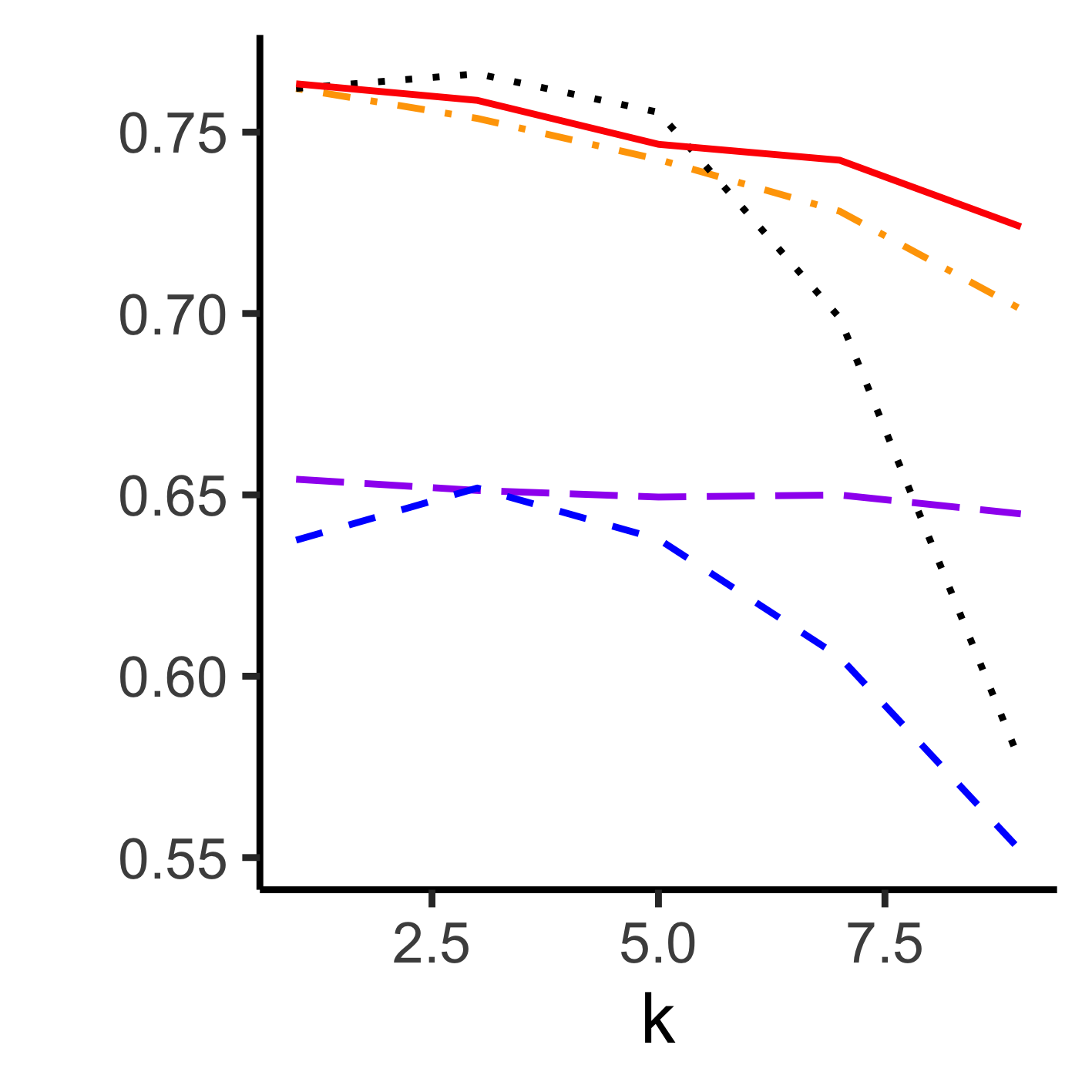}
    \caption{Sigmoid}
  \end{subfigure}
  \begin{subfigure}{0.24\textwidth}
    \includegraphics[width=\linewidth]{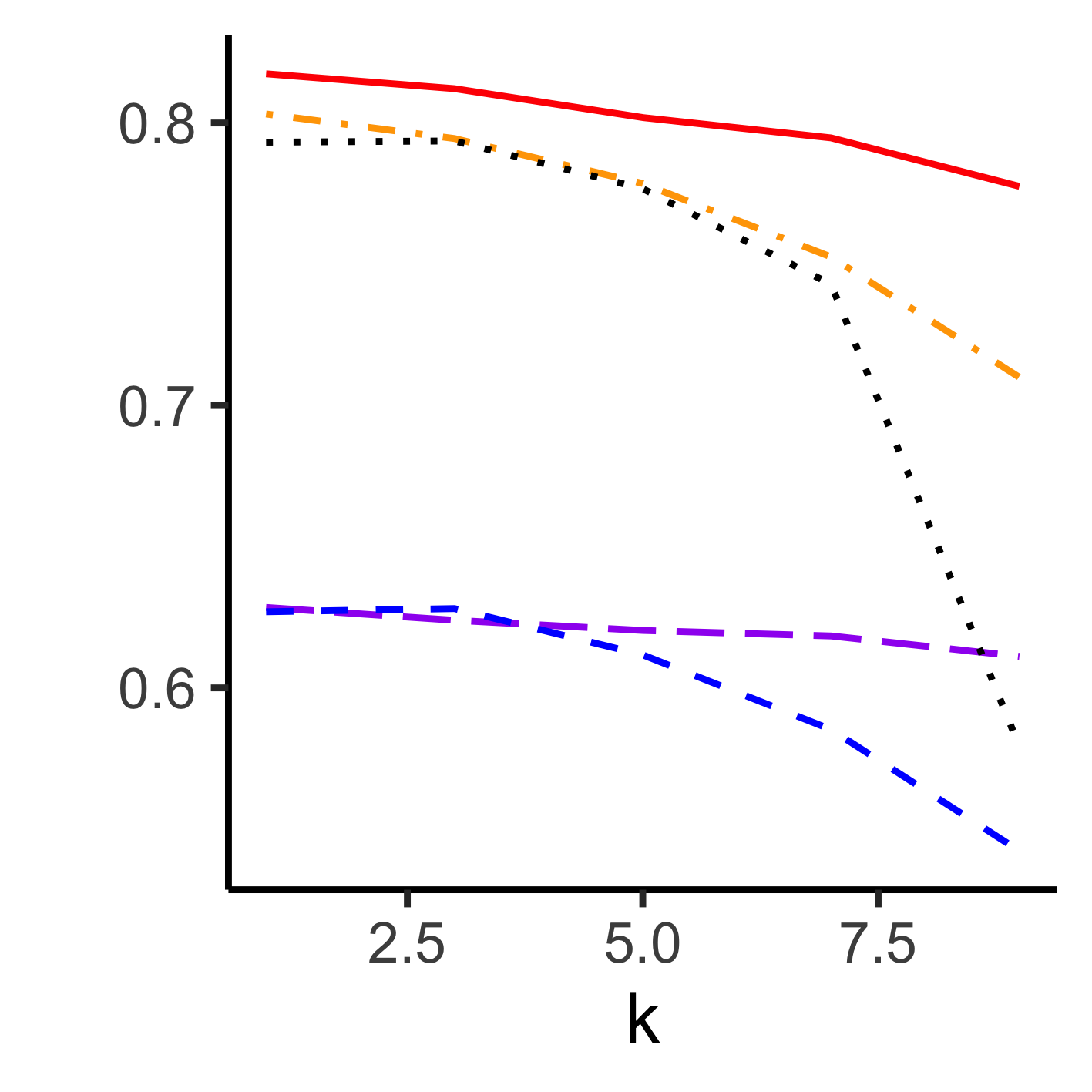}
    \caption{Exponent (Exp)}
  \end{subfigure}
  \begin{subfigure}{0.24\textwidth}
    \includegraphics[clip,width=\linewidth]{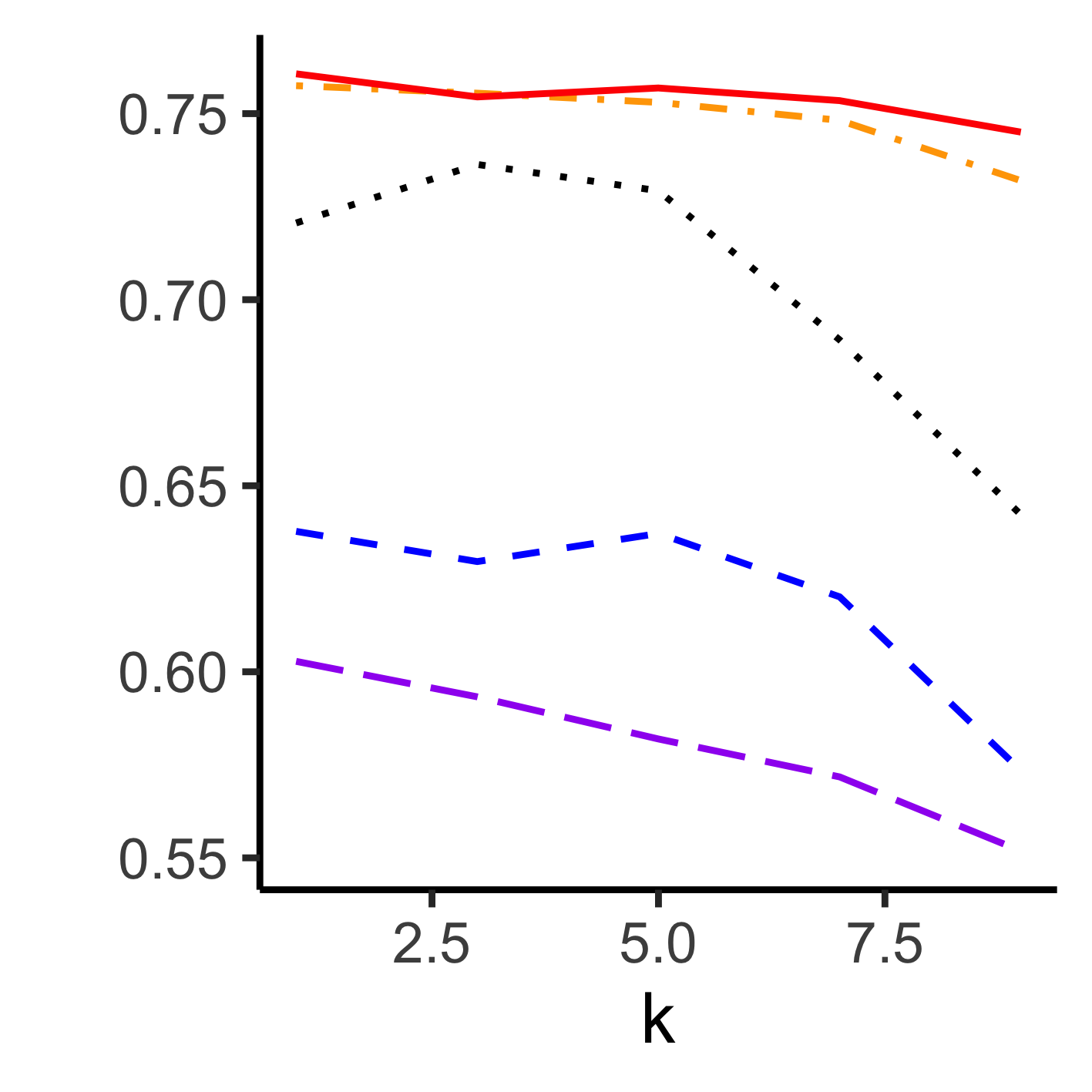}
    \caption{Exp w/ Squared}
  \end{subfigure} 
  \begin{subfigure}{0.24\textwidth}
    \includegraphics[clip,width=\linewidth]{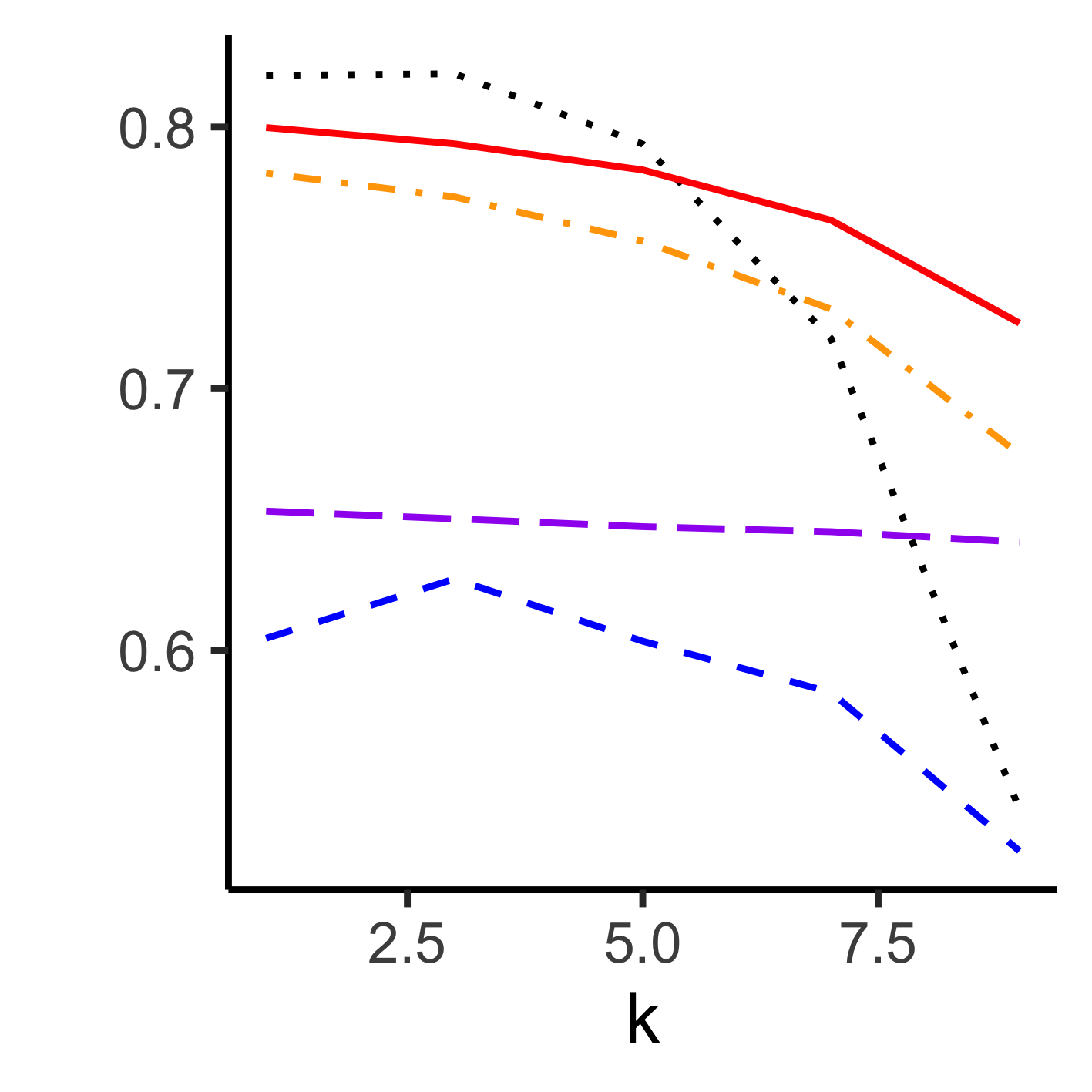}
    \caption{Exp w/ Two Weights}
  \end{subfigure}   
  \caption{\textit{Combined} metric results on the synthetic data as $k$ is varied. \textit{Gradient boosting} is used for the \textcolor{black}{target ML} function. (Best view in color)}
  \label{fig:syn:gb:com}
\end{figure*}

\section{Numerical Results} \label{Experiments}
We describe the implementation of our method and our simulated data in Section \ref{DataGeneration} and the baseline methods in Section \ref{Baselines}, and then present the results in Section \ref{ImplementationAndResults}.

\begin{figure*}[!t]
\centering
  \begin{subfigure}{1.1\textwidth}
    \includegraphics[trim=900 0 0 0,clip,width=\linewidth]{figures/legend}
  \end{subfigure}
  
  \begin{subfigure}{0.24\textwidth}
    \includegraphics[width=\linewidth]{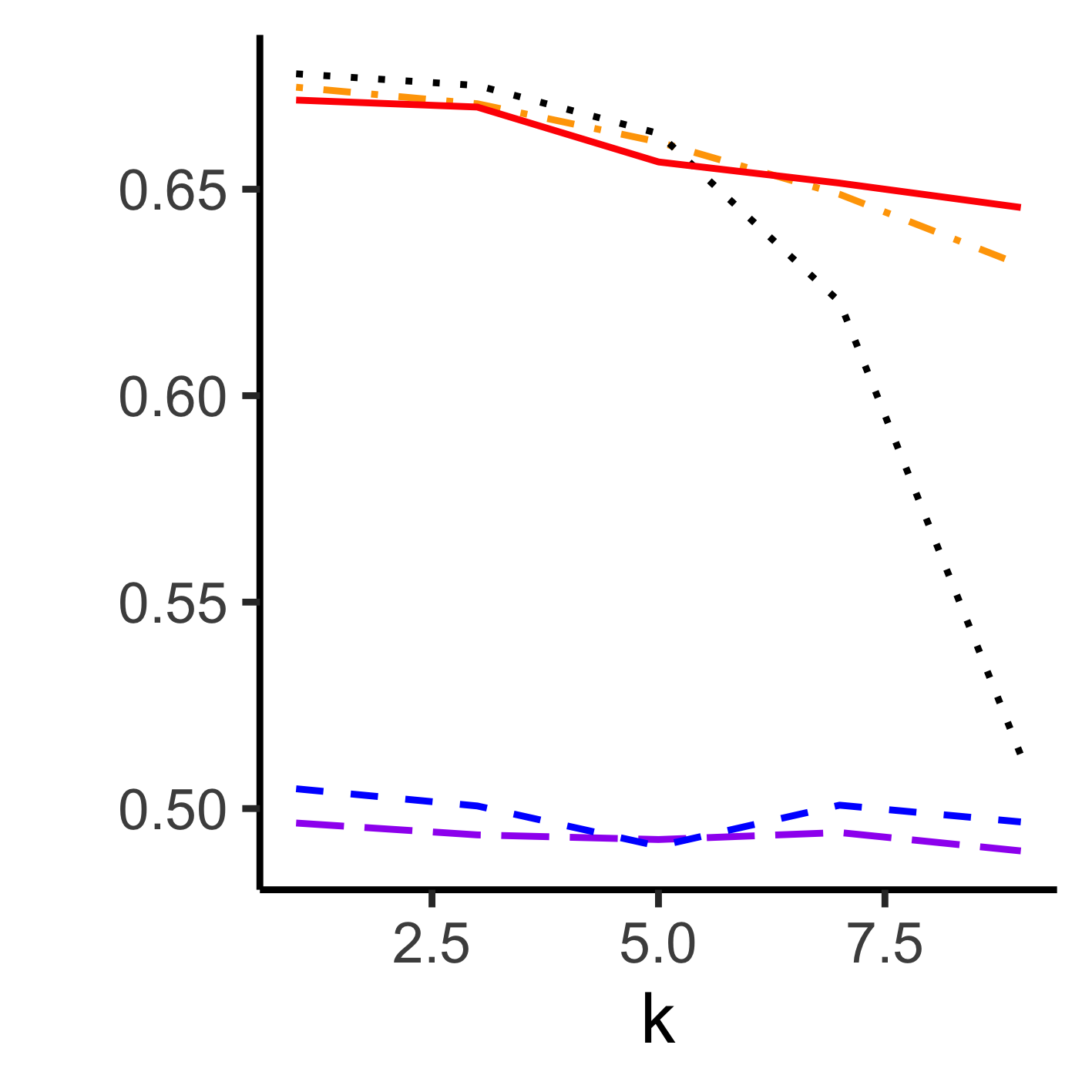}
    \caption{Sigmoid}
  \end{subfigure}
  \begin{subfigure}{0.24\textwidth}
    \includegraphics[width=\linewidth]{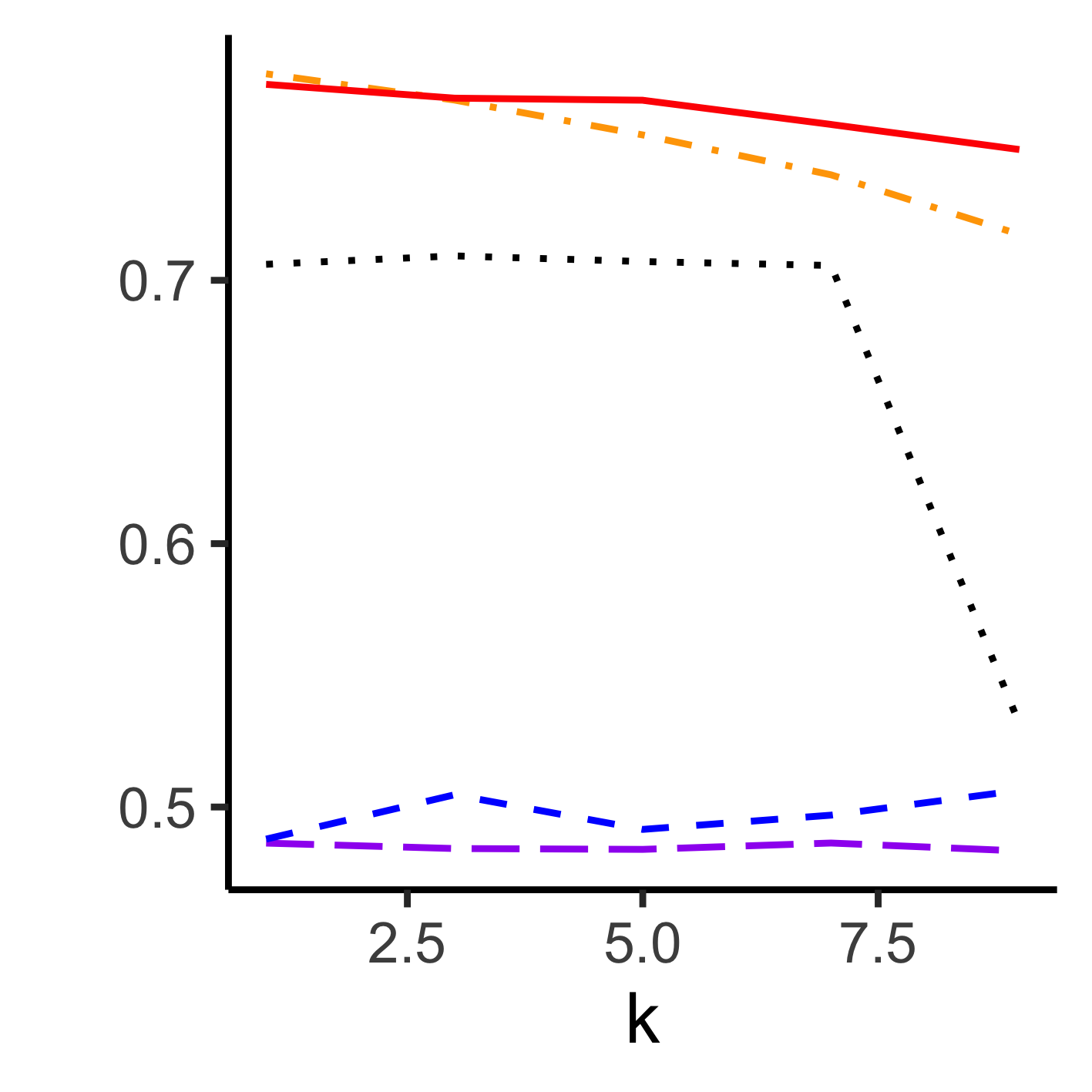}
    \caption{Exponent (Exp)}
  \end{subfigure}
  \begin{subfigure}{0.24\textwidth}
    \includegraphics[width=\linewidth]{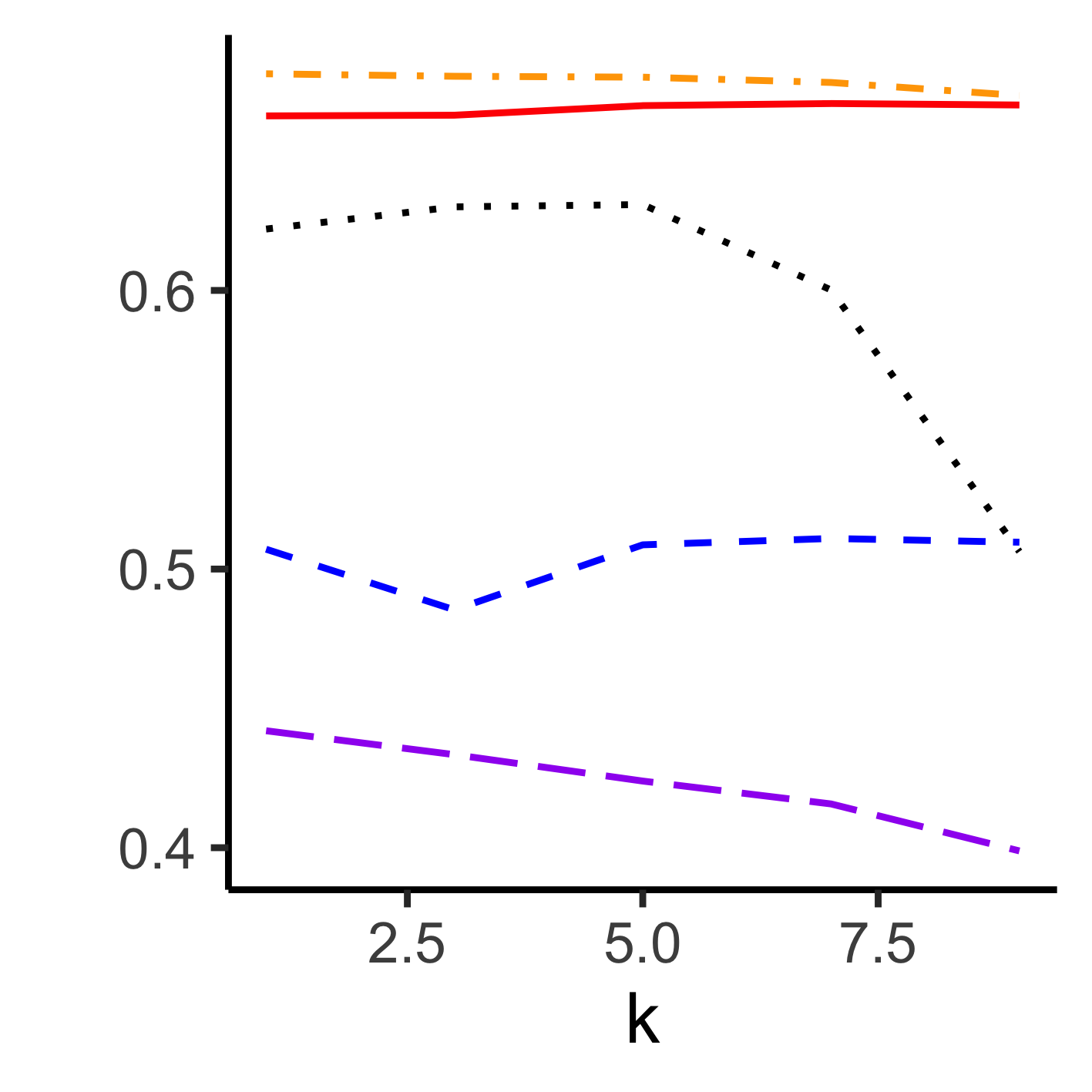}
    \caption{Exp w/ Squared}      
  \end{subfigure} 
  \begin{subfigure}{0.24\textwidth}
    \includegraphics[width=\linewidth]{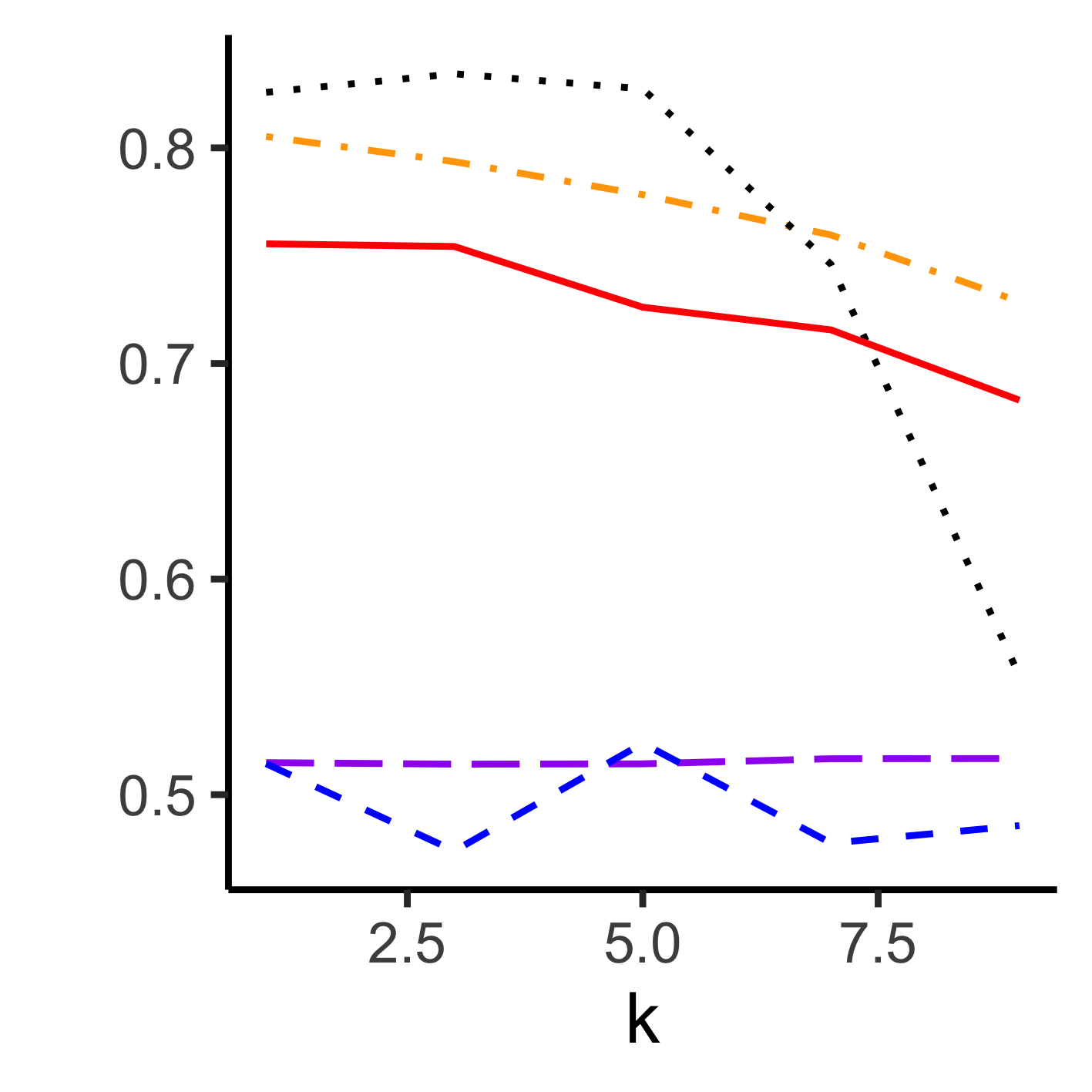}
    \caption{Exp w/ Two Weights}
  \end{subfigure}   
  \caption{Accuracy results on the synthetic data as $k$ is varied. \textit{Logistic regression} is used for the \textcolor{black}{target ML} function. (Best view in color)}
  \label{fig:syn:lr:acc}
\end{figure*}

\begin{figure*}[!t]
\centering
  \begin{subfigure}{0.24\textwidth}
    \includegraphics[width=\linewidth]{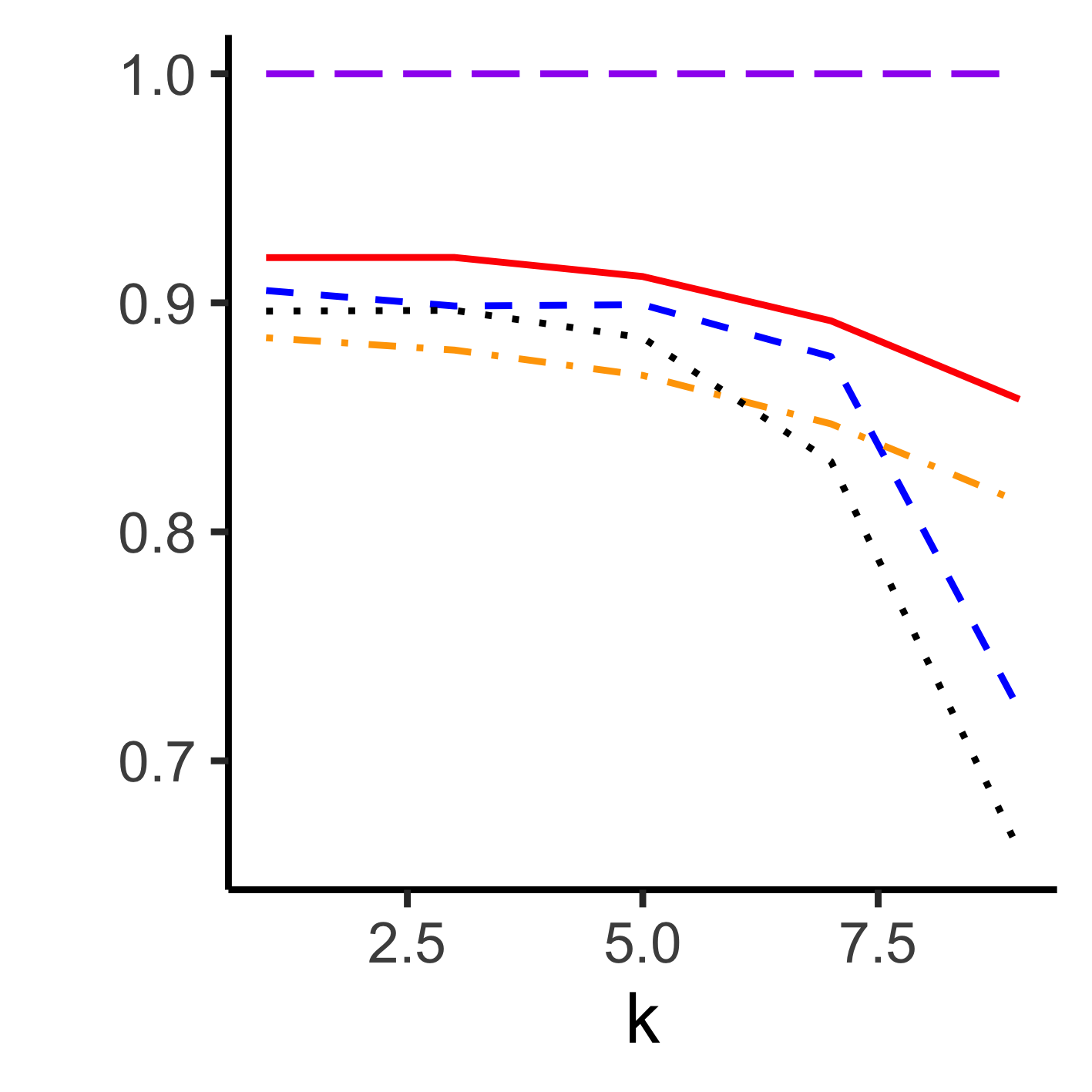}
    \caption{Sigmoid}
  \end{subfigure}
  \begin{subfigure}{0.24\textwidth}
    \includegraphics[width=\linewidth]{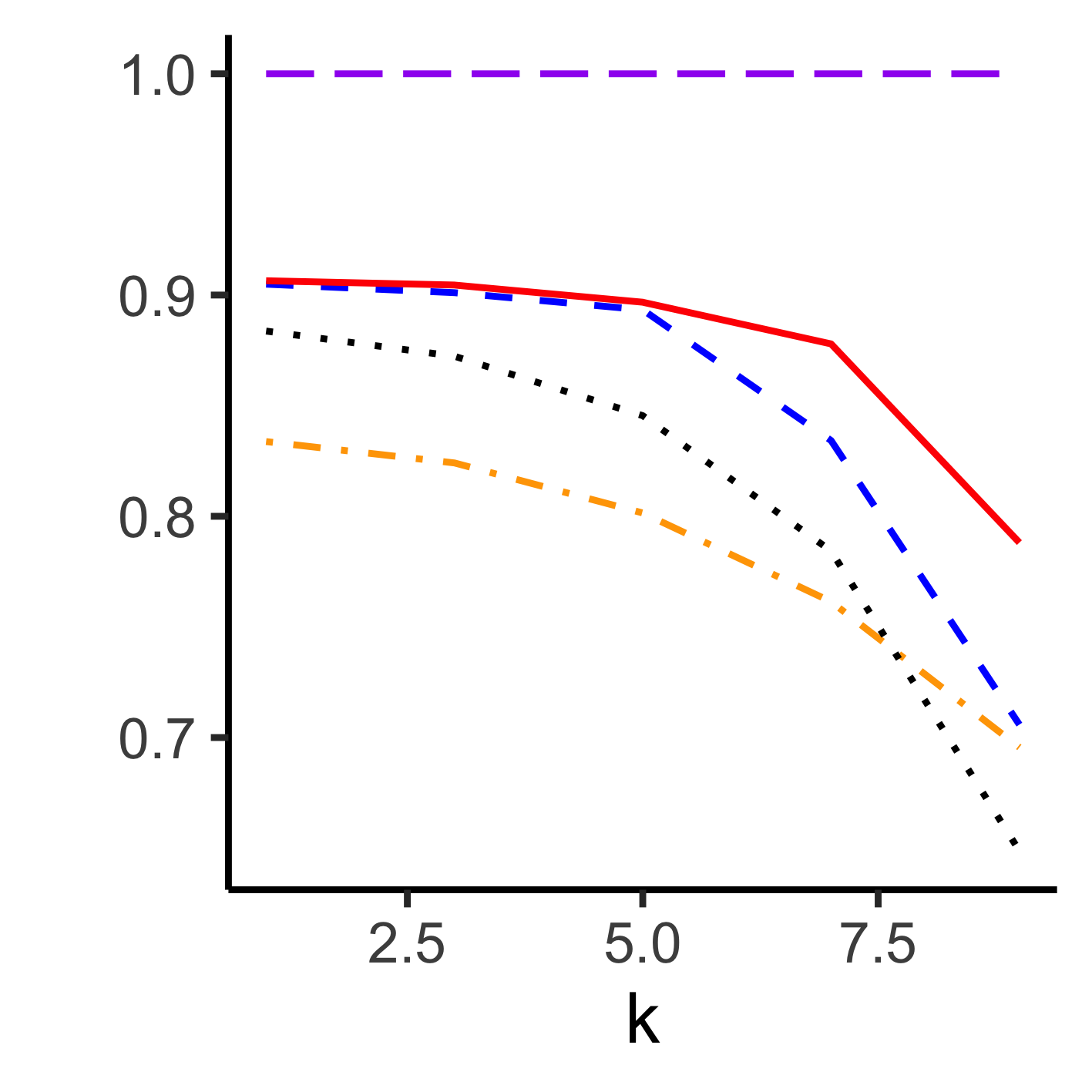}
    \caption{Exponent (Exp)}
  \end{subfigure}
  \begin{subfigure}{0.24\textwidth}
    \includegraphics[clip,width=\linewidth]{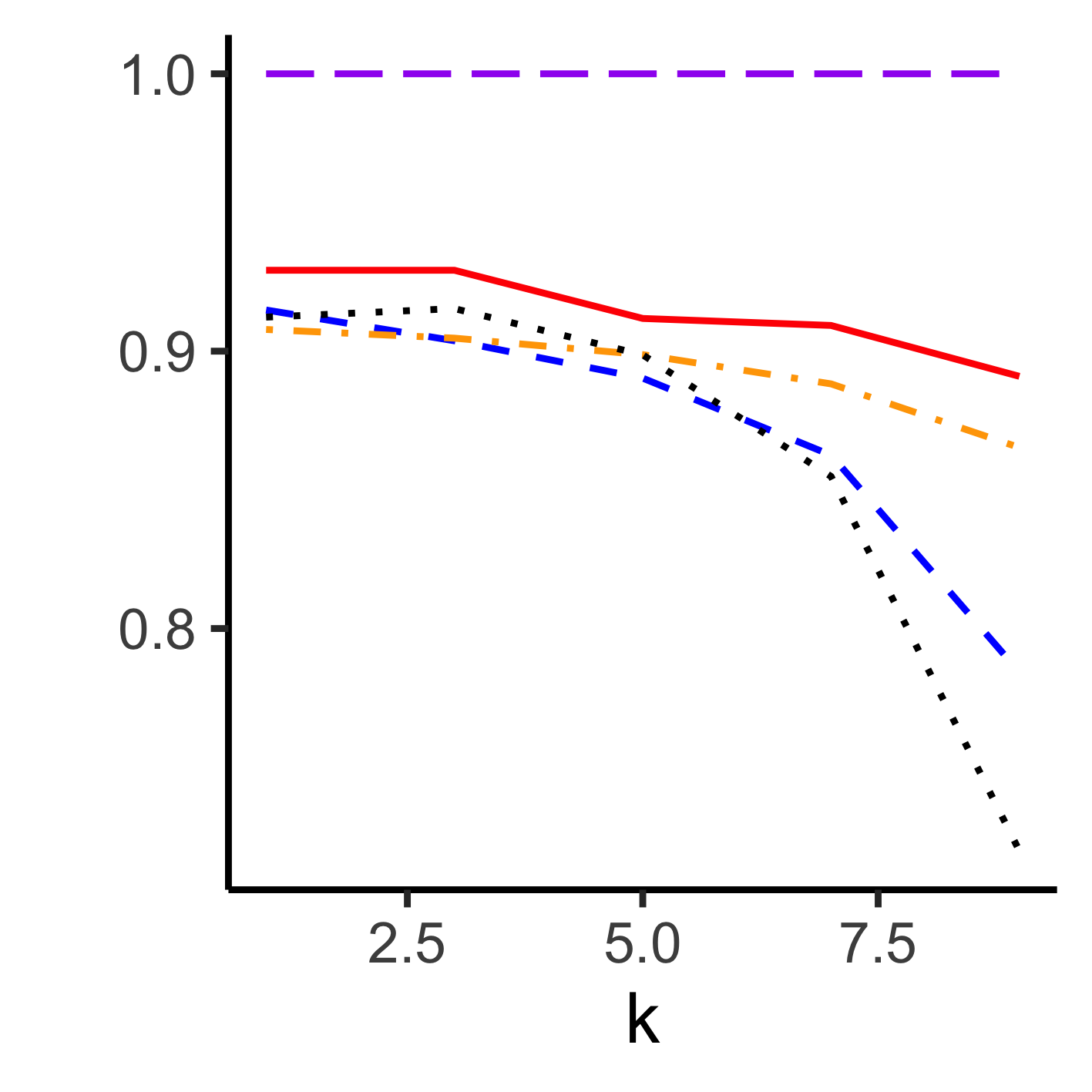}
    \caption{Exp w/ Squared}
  \end{subfigure} 
  \begin{subfigure}{0.24\textwidth}
    \includegraphics[clip,width=\linewidth]{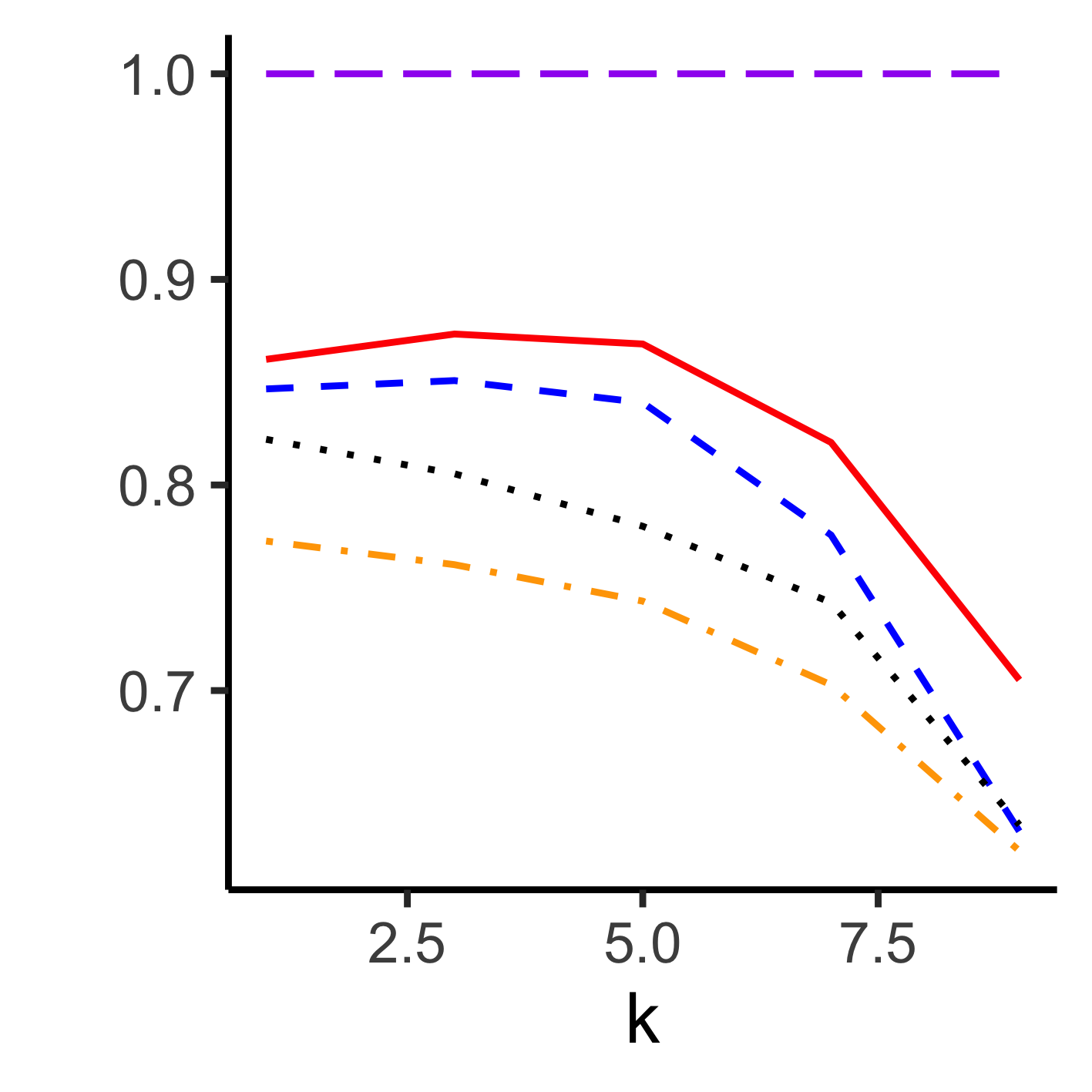}
    \caption{Exp w/ Two Weights}
  \end{subfigure}   
  \caption{Results for the \textit{Closeness} metric on the synthetic data as $k$ is varied. \textit{Logistic regression} is used for the \textcolor{black}{target ML} function. (Best view in color)}
  \label{fig:syn:lr:clo}
\end{figure*}

\begin{figure*}[!t]
\centering
  \begin{subfigure}{0.24\textwidth}
    \includegraphics[width=\linewidth]{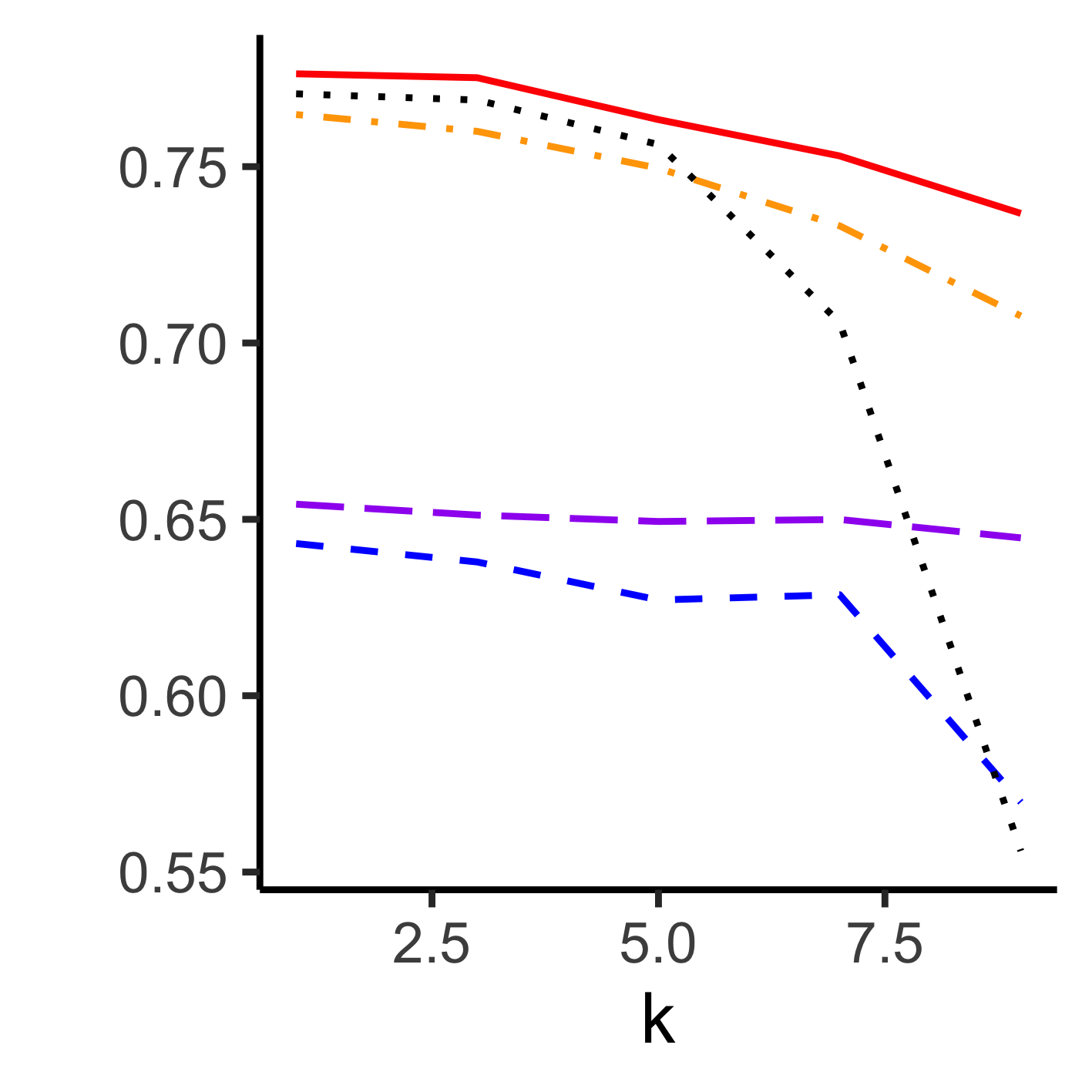}
    \caption{Sigmoid}
  \end{subfigure}
  \begin{subfigure}{0.24\textwidth}
    \includegraphics[width=\linewidth]{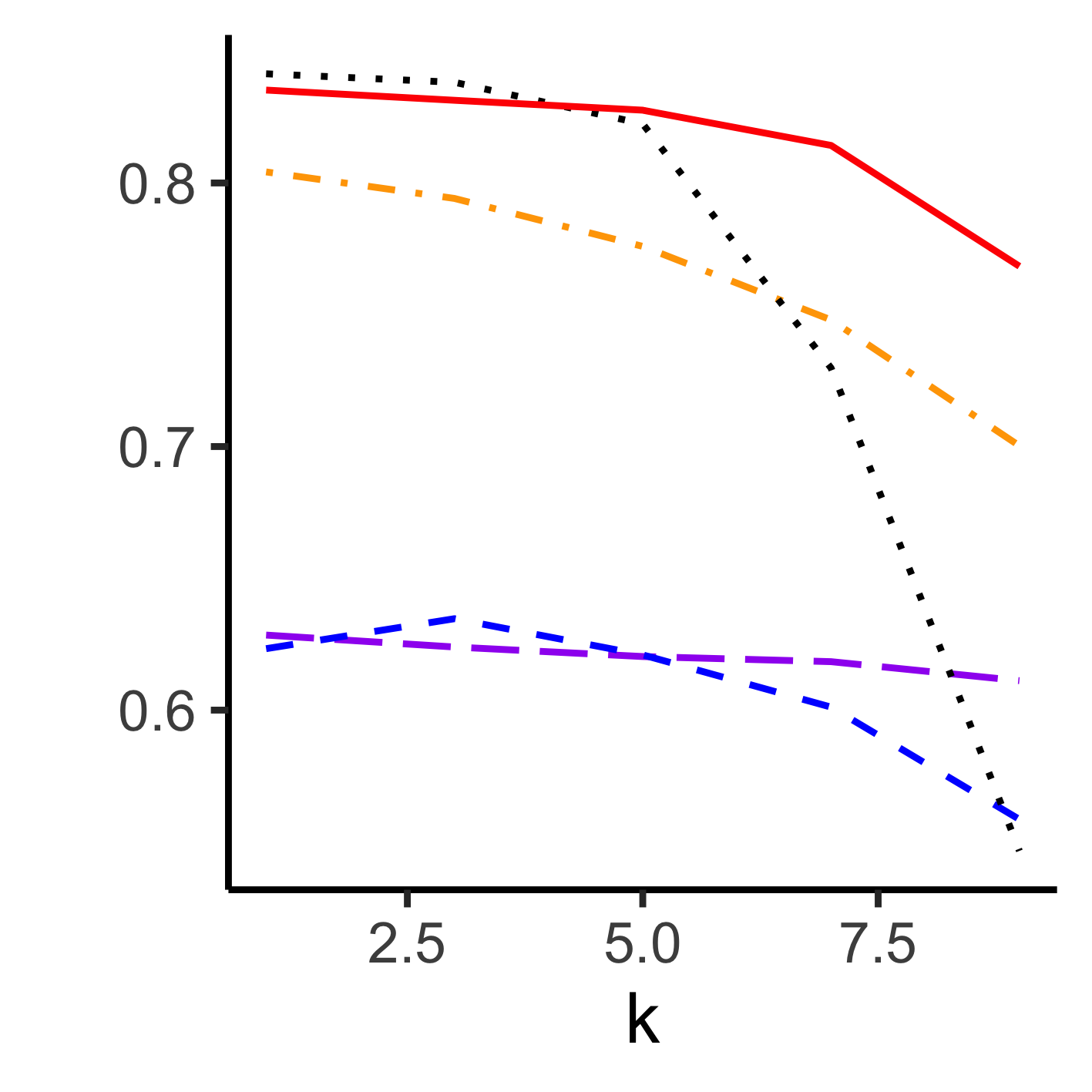}
    \caption{Exponent (Exp)}
  \end{subfigure}
  \begin{subfigure}{0.24\textwidth}
    \includegraphics[clip,width=\linewidth]{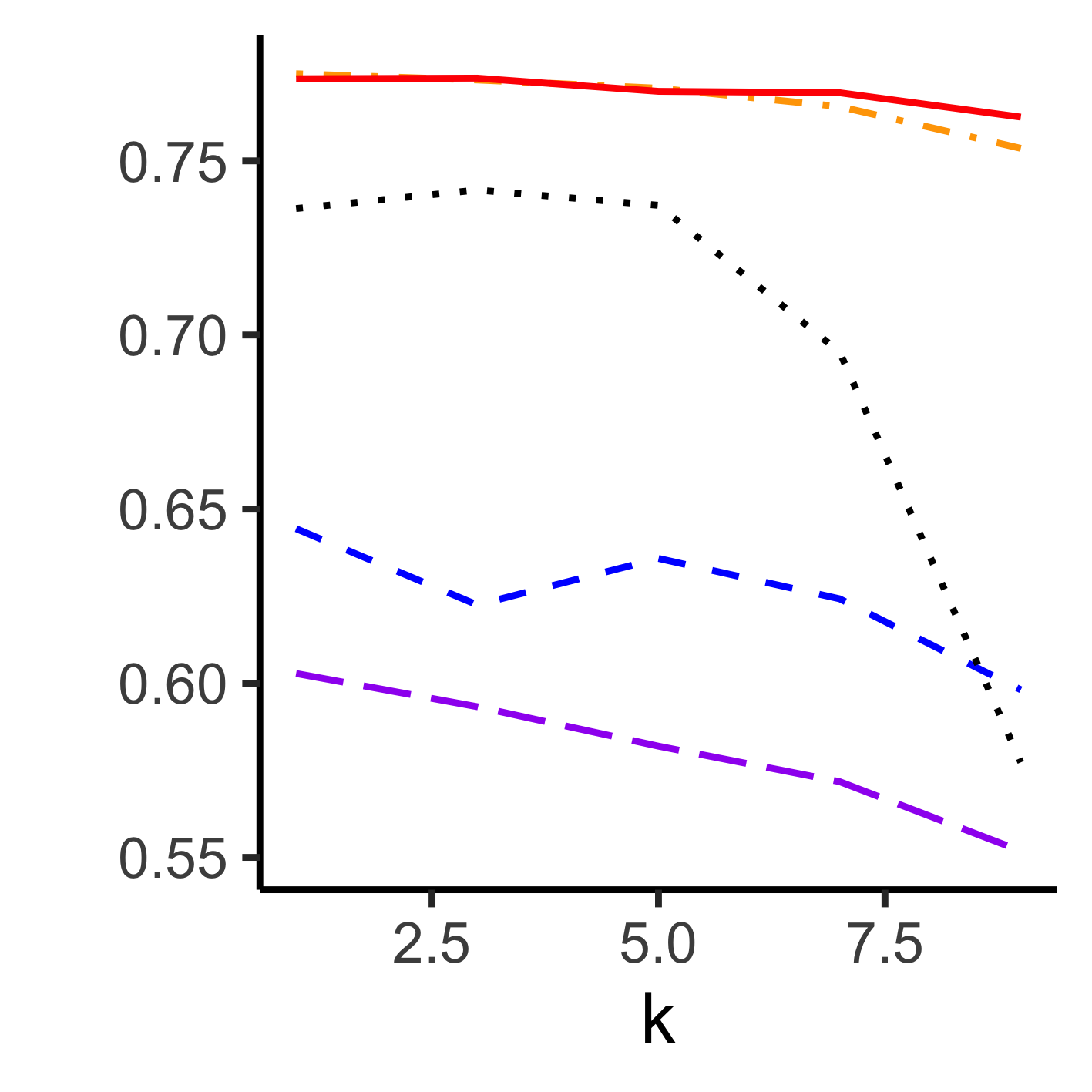}
    \caption{Exp w/ Squared}
  \end{subfigure} 
  \begin{subfigure}{0.24\textwidth}
    \includegraphics[clip,width=\linewidth]{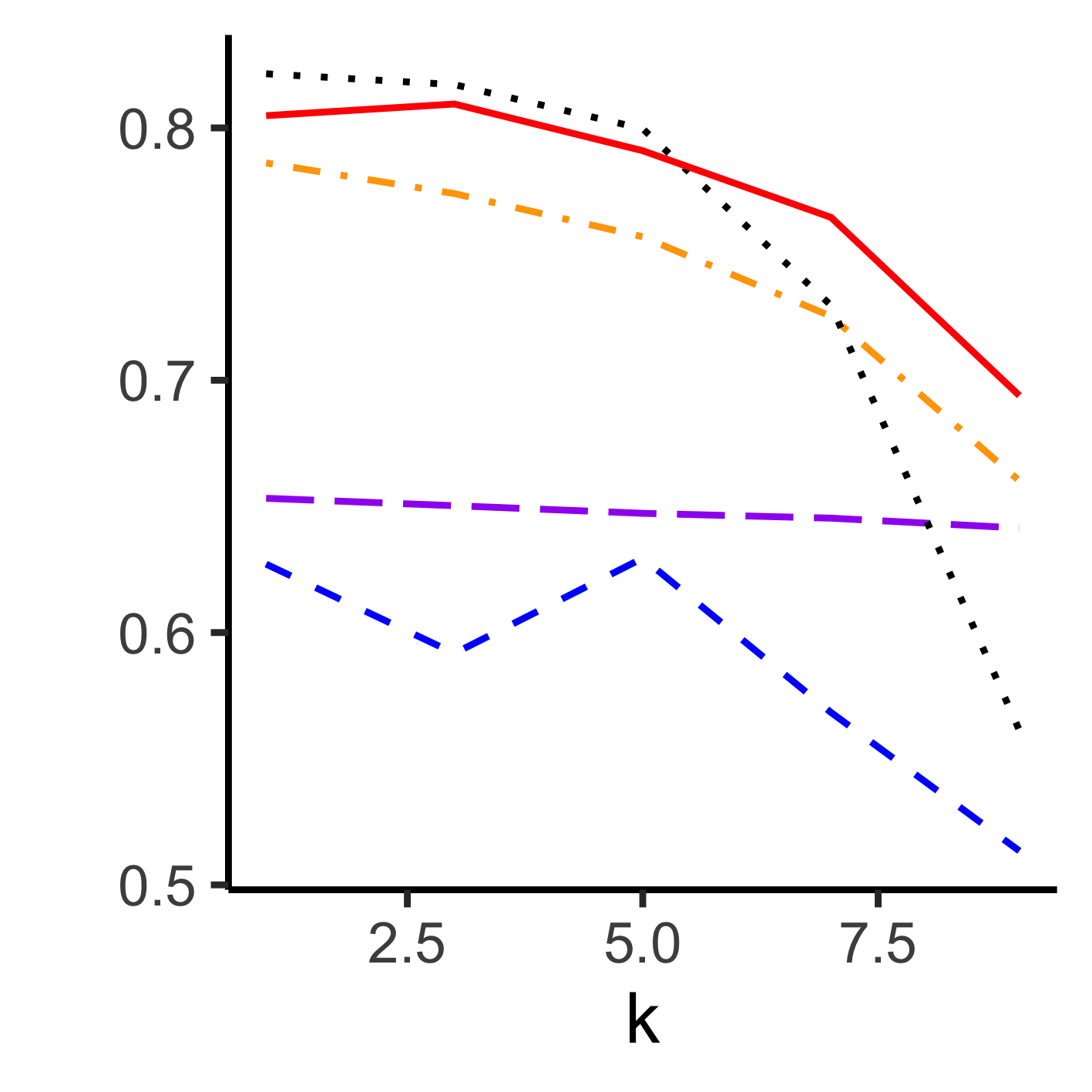}
    \caption{Exp w/ Two Weights}
  \end{subfigure}   
  \caption{Results for the \textit{Combined} metric on the synthetic data as $k$ is varied. \textit{Logistic regression} is used for the \textcolor{black}{target ML} function. (Best view in color)}
  \label{fig:syn:lr:com}
\end{figure*}

\subsection{Data Generation and Implementation Details} \label{DataGeneration}
For each dataset, we set the number of samples/data points, $n$, to $1000$ and determine the number of features, $m$, by sampling it from a uniform distribution between $1$ and $20$. We then generate the values for the corresponding input feature, $X \in \mathbb{R}^{n \times m}$, and label, $Y \in \mathbb{R}^n$. We randomly choose a single feature from $X$ and consider it to be our \textcolor{black}{judgment} variable, $Z$, where each of the features has an equal probability of being chosen. To assign the 0-1 labels, $Y$, for each data point in $X$, we use four different functions: sigmoid($W_1X$), exponent($W_1X$), exponent(${(W_1X)}^2$), and exponent($W_1X + W_2X$), where $W_1 \in \mathcal{R}^{m}$ and $W_2 \in \mathcal{R}^{m}$ are randomly sampled from normal distributions. We denote these functions later as $Sigmoid$, $Exponent \ (Exp)$, $Exp \ w/ Squared$, and $Exp \ w/ \ Two \ Weights$, respectively. For each of the 4 functions, after we apply it on $X$, we normalize the output so that it is between 0 and 1. Then, if it is less than 0.5, we set the corresponding value in $Y$ to 0, otherwise, we set it to 1. We generate the values in $X$ itself from a uniform distribution between -4 and 4 for the first, second, and fourth aforementioned functional forms, and between -1 and 1 for the second one. These values for the uniform distribution make the proportion of labels ($Y$) almost evenly distributed between 0 and 1. We generate 100 datasets for each of these four functional configurations and report the average values for each configuration below. For the \textcolor{black}{judgment} function, $g$, we assumed the same exponent (or sigmoid) function with its coefficients, which are used for generating $Y$, is leveraged. For example, in the case of $Y$=exponent($W_1X$), $g(Z)$ = exponent($W_1[t] \cdot Z$) where $t$ is the index of $Z$ in $W_1$. We also normalize its output of exponent using the same way we did in generating $Y$.
To test the effect of the presence/absence of enough values of the \textcolor{black}{judgment} variable in the training data (i.e., to vary the range of data availability as discussed in the motivation of our work in Section \ref{Introduction}), we adopted the following mechanism for randomly splitting each dataset into a training subset ($80\%$), a validation subset ($10\%$), and a testing subset ($10\%$). In detail, to sample a testing data subset, we (a) take $k\%$ data points of the total $1000$ points, which have the highest $k\%$ of the \textcolor{black}{judgment} variable values, and (b) $(10-k)\%$ of it are chosen randomly from the rest. The remaining $80 \%$ and $10 \%$ of $1000$ points is used for training and validation, respectively. Now, varying $k\%$ would increase or decrease the amount of unseen values for the \textcolor{black}{judgment} variable in the training data. This is analogous to our motivating example, where the higher prices were not represented in the training data. Among our results below are ones for varying values of $k$ at 1\%, 3\%, 5\%, 7\%, and 9\%. For all neural network models, we used the following parameter values: max epochs=1000, dropout\_rate = 0.2, learn\_rate = 0.01, and optimizer = \textit{adam}. For the sigmoid function in our framework, we set 2 for $\beta$. For the supervised loss in the models, we use cross-entropy when they are optimized. For $\alpha$, we grid-search its best value amongst the following values $[0.1, 0.5, 1, 2, 5, 8]$, where the criteria are based on minimizing the $combined$ metric on the validation datasets. For the \textit{closeness}, we set the number of buckets, $q$, to 12. We used both logistic regression and gradient boosting as the \textcolor{black}{ML} classification models in our method and as the basis for learning the supervised loss in the baseline models we explain below. All \textcolor{black}{ML} models were implemented in python using Scikit-learn \citep{pedregosa2011scikit} with its default parameters. The deep learning parts of our method are implemented on Tensorflow 1.2 \citep{abadi2016tensorflow}. We set the size of mini-batches to 8 and the size of each hidden layer to 16 for all experiments. 
\textcolor{black}{The numbers of neural network layers for our generator and discriminator are set to 2 and 4, respectively, which are searched from [1, 2, 3, 4] using the validation dataset.}



\subsection{Baseline Models} \label{Baselines}	
We compare our method to the following baseline models:	

\vspace{1mm}
\noindent\textbf{Supervised Learning (SL)}: \textcolor{black}{This represents a ML model that learns the relationship between input data and labels}, but without the correction we do in our method, i.e., without incorporating any expert's \textcolor{black}{judgment} in it. We use the learning function below to find the model's parameters, where it considers a supervised loss $\mathcal{L}$ between the prediction of \textcolor{black}{ML} model $f$ and true class label $y_i$ with the standard regularized term $R(f)$. We used L2 for $R(f)$ in our experiments.	
\begin{equation}	
f_{sl} = \underset{f \in F}{arg\ min}\sum_{i=1}^{|X_{train}|} \mathcal{L}(y_i, f(x_i)) + R(f).	
\end{equation}	

\noindent\textbf{Weak Supervision with \textcolor{black}{Judgment} (WS)}: Another baseline method we compare against is a weak supervision-based learning method \citep{stewart2017label}. This method leverages the same \textcolor{black}{judgment} function that we use in our method to penalize the model when the prediction is different from the \textcolor{black}{judgment} function. It estimates its parameters as follows:	
\begin{equation}	
f_{ws} = \underset{f \in F}{arg\ min} \sum_{i=1}^{|X_{train}|} |g(x_i) - f(x_i)| + R(f)	
\end{equation}	
\noindent $f(x_i)$ is the prediction of \textcolor{black}{ML} model and $g$ is a \textcolor{black}{judgment} function. The regularization term $R(f)$ is used in the same way as in the SL method above. We also used L2 for WS in our experiments.	

\vspace{2mm}
\noindent\textbf{Supervised Learning with Expectation Regularization (ER)}: Expectation regularization \citep{mann2007simple} is another baseline method that takes into account the expert's \textcolor{black}{judgment}. This model finds its parameters using:	
\begin{equation}	
f_{er} = \underset{f \in F}{arg\ min}\sum_{i=1}^{|X_{train}|} \mathcal{L}(y_i - f(x_i)) - \lambda H(f(x_i),g(x_i)).	
\end{equation}	
In addition to the supervised loss $\mathcal{L}$, it additionally leverages a regularization term that minimizes the entropy of the label distribution from the experts' \textcolor{black}{judgment} function, $g$, and the prediction, $f$. The exact regularization term includes Kullback–Leibler divergence between two predictions using a temperature coefficient. We follow the same label regularization term in the work of \citep{mann2007simple}. We also include the \textcolor{black}{judgment} function and call it ``\emph{\textcolor{black}{Judgment}}".

\subsection{Results and Discussion} \label{ImplementationAndResults}
Figures \ref{fig:syn:gb:acc}, \ref{fig:syn:gb:clo}, and \ref{fig:syn:gb:com} show the results of the 4 different types of synthetic datasets (using the 4 functions described above) when gradient boosting was used as the \textcolor{black}{target ML} model. Figure \ref{fig:syn:gb:acc} shows the accuracy metric for all methods, while Figure \ref{fig:syn:gb:clo} illustrates the $closeness$ metric, and Figure \ref{fig:syn:gb:com} plots the $combined$ metric. One can note and discuss multiple interesting results here: WS and ER are slightly closer to the expert \textcolor{black}{judgment} when compared to SL as observed in Figure \ref{fig:syn:gb:clo}. However, this occurs at a sacrifice on accuracy, where they yield a lower accuracy compared to SL, as shown in Figure \ref{fig:syn:gb:acc}, leading to a poor performance for the $combined$ metric (as can be seen in Figure \ref{fig:syn:gb:com}). Our method shows better closeness to the expert \textcolor{black}{judgment} when $k$ becomes larger (See Figure \ref{fig:syn:gb:clo}), and yet, its accuracy is very comparable to (mostly similar or slightly higher than) \textcolor{black}{SL} (See Figure \ref{fig:syn:gb:acc}). This leads to our method having higher $combined$ scores across most values of $k$ for almost all functional forms, as shown in Figure \ref{fig:syn:gb:com}. Moreover, our method is superior compared to all other methods in all metrics at higher values of $k$, where the range of the training distribution is not represented within the testing distribution. Because other methods simply leverages their own penalty scores only when they compute supervised loss values, it is difficult to resolve \textcolor{black}{judgment}-variable-level conflicts and correct them accordingly (in particular, when $k$ is higher). This all shows the efficacy and robustness of our method for treating the research challenge we are solving. We note that the expert \textcolor{black}{judgment} is always 1 when it comes to $closeness$ as per our definition of the $closeness$ metric. However this expert \textcolor{black}{judgment} yields a very low accuracy (and $combined$ values) since it only relies on the \textcolor{black}{judgment} variable.

Meanwhile, when we used logistic regression as the underlying \textcolor{black}{ML} model, we are also getting similar results. Figures \ref{fig:syn:lr:acc}, \ref{fig:syn:lr:clo}, and \ref{fig:syn:lr:com} show the results on the same four different types of synthetic datasets. Again, Figure \ref{fig:syn:lr:acc} shows the accuracy metric for all methods, while Figure \ref{fig:syn:lr:clo} illustrates the $closeness$ metric, and Figure \ref{fig:syn:lr:com} plots the $combined$ metric. Our method shows a better \textit{closeness} to the expert's \textcolor{black}{judgment} across most values of $k$ when compared to all other methods. In addition, its accuracy is very comparable to (either similar or just slightly lower than) \textcolor{black}{SL} (See Figure \ref{fig:syn:lr:acc}). Overall, our method returns a higher $combined$ score across almost all values of $k$ for almost all functional forms, as can be seen in Figure \ref{fig:syn:lr:com}.




\section{Real-world Case Studies} \label{CaseStudy}
We apply our method to two real-world case studies in Sections \ref{ITServices} and \ref{CreditCase}, then present a qualitative analysis in Section \ref{Qual}.

\subsection{The Salesman Knows Better} \label{ITServices}

\begin{table*}[!t]
			\caption{Results of the IT services case study, where \textit{gradient boosting} was used as the \textcolor{black}{target} ML model. Scores in () are the standard deviations. Numbers in \textbf{bold} indicate statistical significance (p-value $< 0.01$). }
			\label{tab:gb:itservice}
    \begin{minipage}[!t]{.99\textwidth }
        \centering    
        \scalebox{0.85}{  
			\begin{tabular}{cccc}
			& Accuracy              & Closeness              & Combined           \\ \hline
			\textcolor{black}{Judgment}  & 0.684 (0.014)          & \textbf{1.0 (0)} & 0.813 (0.010)     \\ \hline \hline
			WS         & 0.682 (0.014)          & 0.858 (0.127) & 0.755 (0.055)          \\
			SL         & \textbf{0.930 (0.006)} & 0.774 (0.061) & 0.843 (0.037)          \\
			ER & 0.925 (0.0062)         & 0.714 (0.147) & 0.798 (0.092)          \\ \hline     
			\textbf{Our Method with LR} & 0.920 (0.0066)  & 0.826 (0.057)          & 0.870 (0.030)\\
			\textbf{Our Method with \mname{}} & 0.862 (0.041)  & 0.920 (0.081)          & \textbf{0.889 (0.051)}
			\end{tabular}
			}
		    \hrule height 0pt
		    \end{minipage}%
\end{table*}

\begin{table*}[!t]
			\caption{Results of the IT services case study, where \textit{logistic regression} was used as the \textcolor{black}{target ML} model. Scores in () are the standard deviations. Numbers in \textbf{bold} indicate statistical significance (p-value $< 0.01$).}
			\label{tab:lr:itservice}
    \begin{minipage}[!t]{.99\textwidth }
        \centering
        \scalebox{0.85}{                     
			\begin{tabular}{cccc}
			& Accuracy              & Closeness              & Combined           \\ \hline
			\textcolor{black}{Judgment}  & 0.684 (0.014)          & \textbf{1.0 (0)} & 0.813 (0.010)     \\ \hline \hline
			WS         & 0.682 (0.014)          & 0.858 (0.127) & 0.755 (0.055)          \\
			SL         & \textbf{0.928 (0.007)} & 0.769 (0.076) & 0.839 (0.046)          \\
			ER & 0.682 (0.0014)         & 0.858 (0.126) & 0.755 (0.055)          \\ \hline     
			\textbf{Our Method with LR} & 0.917 (0.006)  & 0.830 (0.059)          & 0.870 (0.029)\\
			\textbf{Our Method with \mname{}} & 0.890 (0.016)  & 0.949 (0.011)          & \textbf{0.918 (0.012)}
			\end{tabular}
			}
 		    \hrule height 0pt
		    \end{minipage}%
\end{table*}

We here present the results of applying our method (as well as all other baseline methods) to the IT services case study that we motivated this work within Section \ref{Introduction}. 
This was done at one of the biggest IT service providers in the world. We used a dataset that is composed of 4,695 historical deals. 
After discussing with the sales experts of the company, we defined the \textcolor{black}{judgment} function as $g(z) = 1-z$, where $z$ is the normalized price in log-scale, to make the range of the value between 0 and 1. We were able to build accurate \textcolor{black}{ML} models (see accuracy results in Table \ref{tab:gb:itservice} and \ref{tab:lr:itservice}) by including five more features in training the models, namely the type of the client (some clients are cost-savers, others are technology partners, etc.), the country of the client, the industrial sector of the client, how the deal was identified (e.g., from the field, from a prior engagement with the same client, from an ad or posting), and the relationship with the client (how often, if any, the provider has done business with the client before). 

We also implemented a slightly modified approach to ours in which we used linear regression (LR) rather than \mname{} to predict $z_i^{expected}$, using all other features in $X$ as the variables for training the linear regression model. We call this method \say{Our Method with LR} in the results tables below. 	

Tables \ref{tab:gb:itservice} and \ref{tab:lr:itservice} show the results of comparing all methods on all evaluation metrics. In Table \ref{tab:gb:itservice}, we leveraged $gradient \ boosting$ as the \textcolor{black}{ML} classification/prediction model, while for the results in Table \ref{tab:lr:itservice}, we used \textit{logistic regression}. In both tables, one can see that our methods (our \mname{}-based method and our method using LR) show a statistically significantly better performance in the \textit{combined} metric when compared to all other baseline models without a significant deterioration in the accuracy. That statistical significance has a \textit{p-value} less than 0.01. We also note that our \mname{}-based method gives better results than our method using LR.

\subsection{Credit Approval Prediction} \label{CreditCase}

\begin{table*}[!t]
	\caption{Results of the financial case study, where \textit{gradient boosting} was used as the \textcolor{black}{target ML} model. Scores in () are the standard deviations. Numbers in \textbf{bold} indicate statistical significance (p-value $< 0.01$).}
	\label{tab:gb:credit}
    \begin{minipage}[!t]{.99\textwidth }
        \centering
        \scalebox{0.85}{                     
			\begin{tabular}{cccc}
			& Accuracy          & Closeness              & Combined           \\ \hline
			\textcolor{black}{Judgment}           & 0.675 (0.032)         & \textbf{1.0 (0.0)} &   0.806 (0.023)           \\ \hline \hline
			WS                  & 0.500 (0.10)          &  0.848 (0.060) & 0.621 (0.080)          \\
			SL                  & \textbf{0.761 (0.020)} & 0.810 (0.069)          & 0.783 (0.030)          \\
			ER & 0.480 (0.110)  & 0.841 (0.055)  & 0.606 (0.097)      \\ \hline
			\textbf{Our Method with LR}  & \textbf{0.761 (0.020)} & 0.847 (0.080)    & 0.799 (0.034)\\
			\textbf{Our Method with \mname{}}  & 0.736 (0.025) & 0.907 (0.024)    & \textbf{0.812 (0.020)}
			\end{tabular}
			}
		    \hrule height 0pt
		    \end{minipage}%

\end{table*}

\begin{table*}[!t]
\caption{Results of the the financial case study, where \textit{logistic regression} was used as the \textcolor{black}{target ML} model. Scores in () are the standard deviations. Numbers in \textbf{bold} indicate statistical significance (p-value $< 0.01$).}
			\label{tab:lr:credit}
    \begin{minipage}[!t]{.99\textwidth }
        \centering
        \scalebox{0.85}{                     

			\begin{tabular}{cccc}
			& Accuracy          & Closeness              & Combined           \\ \hline
			\textcolor{black}{Judgment}           & 0.675 (0.032)         & \textbf{1.0 (0.0)} &   0.806 (0.023)           \\ \hline \hline
			WS                  & 0.497 (0.10)          &  0.858 (0.026) & 0.623 (0.084)          \\
			SL                  & \textbf{0.754 (0.020)} & 0.804 (0.054)          & 0.777 (0.021)          \\
			ER & 0.503 (0.093)  & 0.884 (0.034)  & 0.636 (0.078)      \\ \hline
			\textbf{Our Method with LR}  & \textbf{0.753 (0.024)} & 0.837 (0.083)    & 0.790 (0.032)\\
			\textbf{Our Method with \mname{}}  & 0.743 (0.029) & 0.91 (0.024)    & \textbf{0.818 (0.021)}
			\end{tabular}
			}
		    \hrule height 0pt
		    \end{minipage}%
			
\end{table*}

This second case study comes from the financial industry, where credit approval (either an approved or a disapproved credit score) is to be predicted from a list of 202 attributes, including age, credit history, etc. The dataset\footnote{https://archive.ics.uci.edu/ml/datasets/statlog+(german+credit+data)} we use is composed of 1,000 records. We chose the existence of a guarantor as our \textcolor{black}{judgment} variable, $z$. Formally, we use $g(z) = 0.5 + 0.5 \cdot z$, where $z$ is a binary variable. When a person has a guarantor, she is most likely to be approved, while when she does not, we cannot determine the potential effect on the credit score and hence the 0.5 we used for such case. 

Table \ref{tab:gb:credit} reports the results for this dataset, where $gradient \ boosting$ was used as the \textcolor{black}{target ML} model. While the \textcolor{black}{judgment} variable yields high accuracy and $combined$ score (besides being 1 for $closeness$ of course), yet, our method still statistically significantly outperforms it as well as all other baselines. Meanwhile, as in Table \ref{tab:lr:credit}, when we use $logistic \ regression$ for the \textcolor{black}{target ML} model, our \mname{}-based method still outperforms all other baselines.

\begin{figure*}[!t]
\centering
  \begin{subfigure}{1.5\textwidth}
    \includegraphics[trim=1500 0 0 0,clip,width=\linewidth]{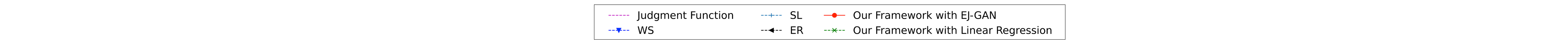}
  \end{subfigure}

\begin{subfigure}[t]{0.99\textwidth}
    \centering
  \includegraphics[trim=15 0 0 0,clip,height=2.02in]{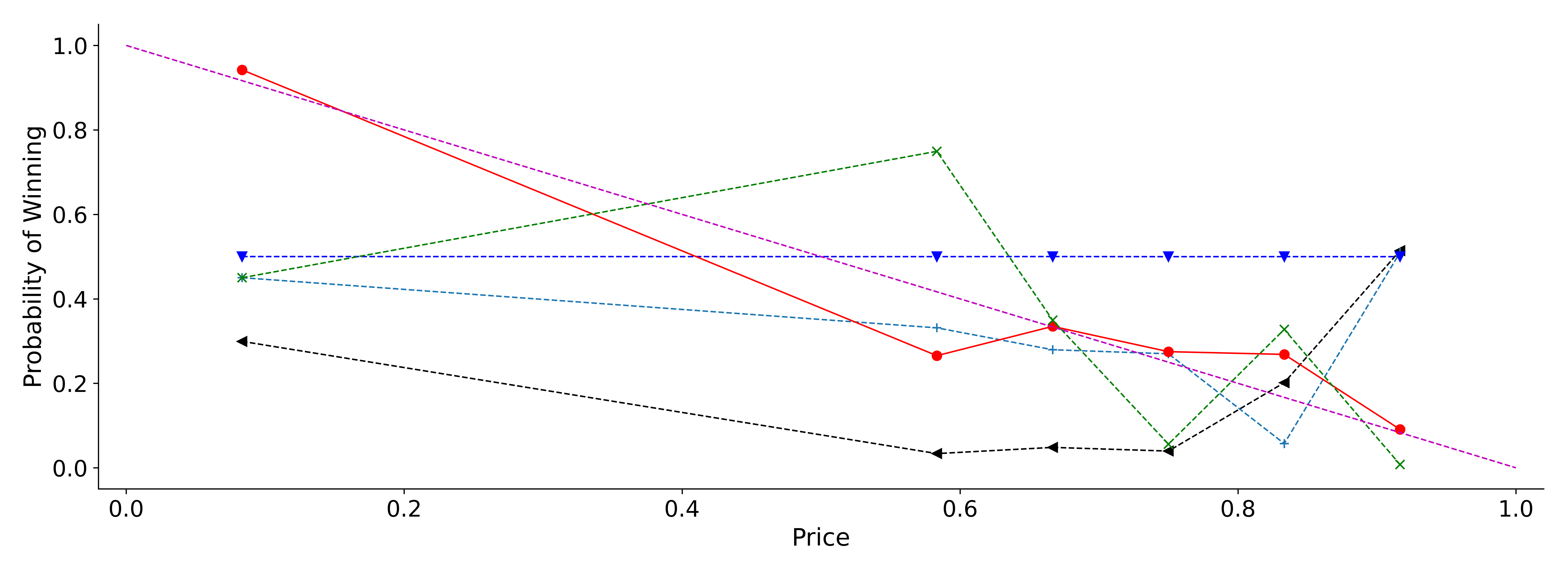}
    \caption{IT services case study}
 \end{subfigure}
\begin{subfigure}[t]{0.98\textwidth}
    \centering
  \includegraphics[trim=15 0 0 0,clip,height=2.02in]{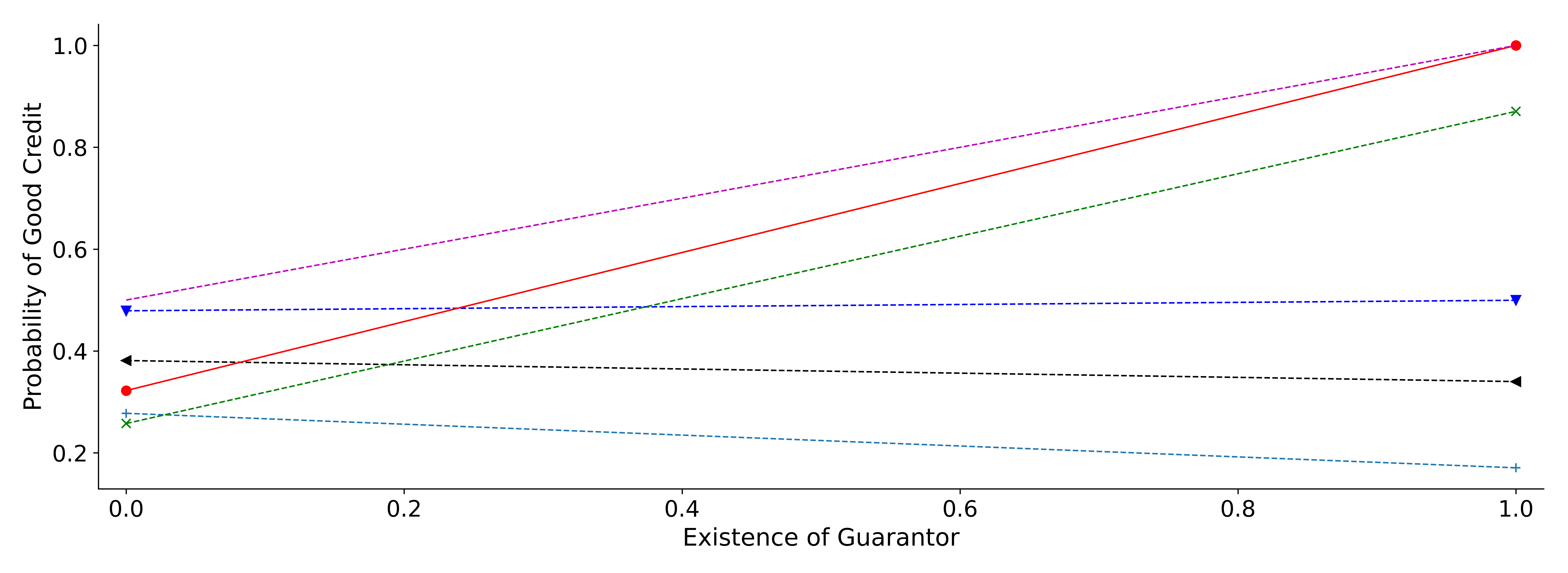}
    \caption{Financial case study}
 \end{subfigure} 
 \caption{\textcolor{black}{Prediction Probability over Different Buckets on \textcolor{black}{judgment} Variable. \textit{Gradient boosting} is used for the \textcolor{black}{target ML model}. (Best view in color)}}
  \label{fig:exp:qualitative:gb:realworld}
\end{figure*}

\begin{figure*}[!t]
\centering
  \begin{subfigure}{1.5\textwidth}
    \includegraphics[trim=1500 0 0 0,clip,width=\linewidth]{figures/legend-realworld}
  \end{subfigure}

\begin{subfigure}[t]{0.99\textwidth}
    \centering
  \includegraphics[trim=15 0 0 0,clip,height=2.02in]{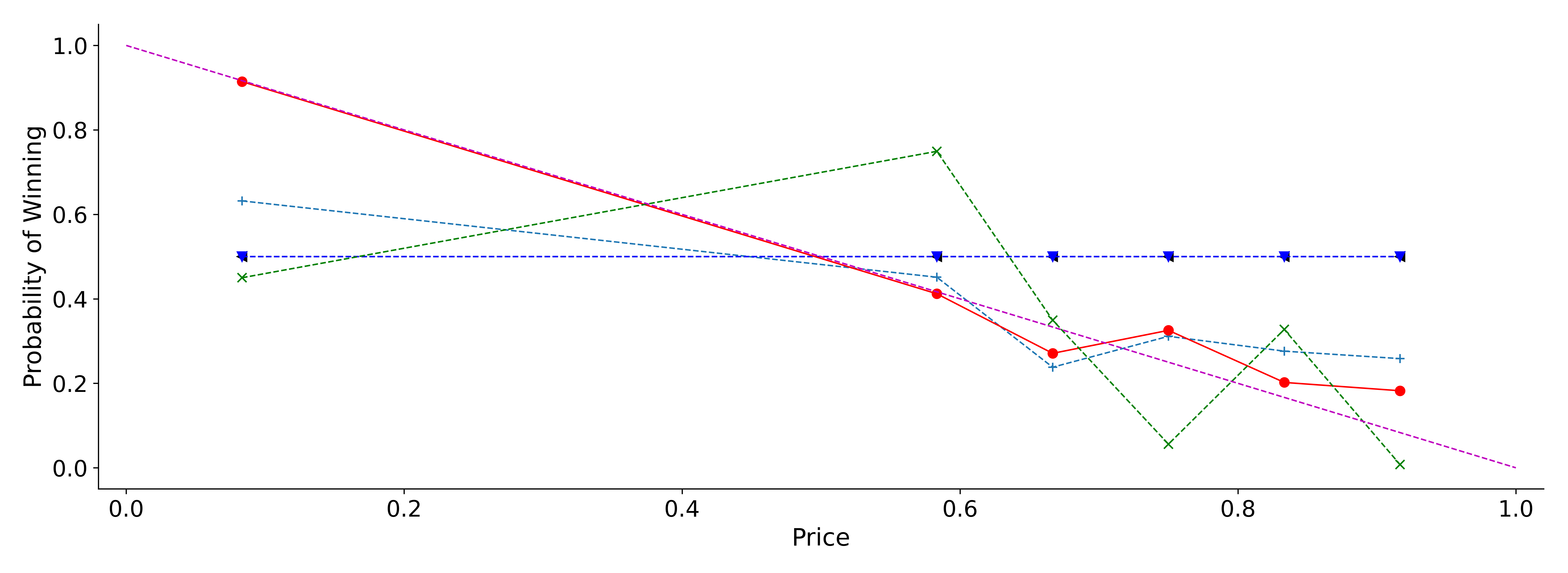}
    \caption{IT services case study}
 \end{subfigure}
\begin{subfigure}[t]{0.98\textwidth}
    \centering
  \includegraphics[trim=15 0 0 0,clip,height=2.02in]{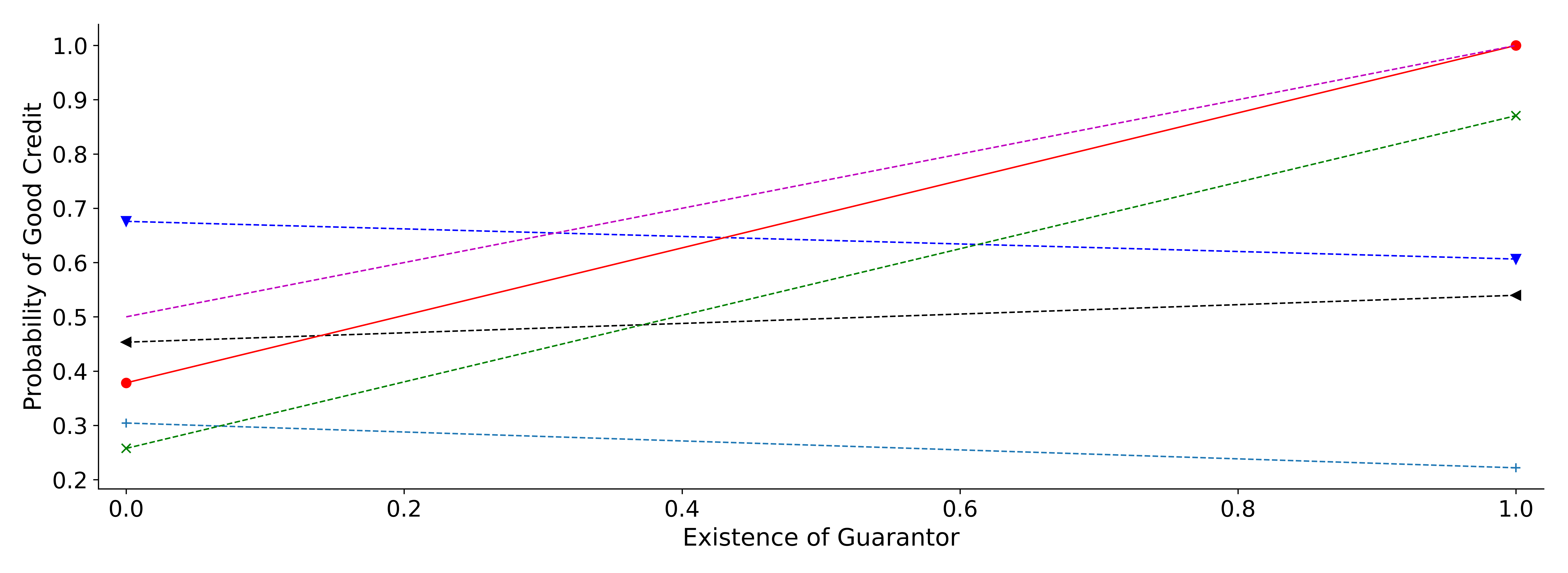}
    \caption{Financial case study}
 \end{subfigure} 
 \caption{\textcolor{black}{Prediction probability over different buckets of the \textcolor{black}{judgment} variable. \textit{Logistic regression} is used for the \textcolor{black}{target ML model}. (Best view in color)}}
 \vspace{-2mm}
  \label{fig:exp:qualitative:lr:realworld}
\end{figure*}

\subsection{Qualitative Analysis} \label{Qual}

We now do a qualitative analysis for our two real-world cases above, to further show the effectiveness of our method. In particular, we plot the \textcolor{black}{judgment} variable versus the predicted probability (after bucketing the \textcolor{black}{judgment} variable values and calculating the average result for each bucket) in Figure \ref{fig:exp:qualitative:gb:realworld} and \ref{fig:exp:qualitative:lr:realworld}. This provides a human-readable qualitative analysis of the $closeness$ metric. One can see that our method looks consistently/desirably close to the expert's \textcolor{black}{judgment} (again, while not scarifying the accuracy much as can be seen in the above tables). While its results for the $combined$ metric are statistically significantly better than all other models, our method using \mname{} statistically significantly outperforms it.

\subsection{Ablation Study of Model Components}\label{sec:ablation}

\textcolor{black}{
We conduct ablation studies on our proposed method to explore the effect of the GAN-based judgment variable estimator, \mname{}. We can develop other variants of our \mname{} by using DRC and data imputation techniques for the judgment variable estimator. Here, we implement five variants of \mname{} as:}

\begin{itemize}
	\item \textcolor{black}{\textbf{DRC}~\citep{drc2}. DRC measures can be used for quantifying the similarity between training data and unseen data and giving an indication as to when the performance of the existing supervised classifier decreases. In other words, when a DRC value is larger than 1, it implies that the training and unseen data are not exchangeable and the existing supervised model will under-perform when applied to the unseen data. In this case, we can assume that judgment functions are potentially more helpful, and to validate the assumption, the DRC-based judgment incorporation method is compared to our proposed conflict resolution method. To implement the DRC-based judgment incorporation method, the base distribution of DRC is chosen as \textit{Beta(1,1)} distribution, which is recommended by \citep{drc2}, and the separability between training data and test data is measured. DRC values are computed via the original implementation\footnote{https://github.com/eschat/DRC} and are normalized by leveraging the sigmoid function. The output values are used for determining the weight of judgment function $g$ to unseen data, (i.e., $w$ of Algorithm \ref{alg:framework}), and then it corrects the original prediction as in Line 4-5 of Algorithm \ref{alg:framework}.}
	\item \textcolor{black}{\textbf{k-NN.} This is a data imputation method for estimating the judgment variable using $k$-nearest neighbors. The values of each sample are imputed using the mean value from $k$-neighbors nearest neighbors, which are found in the training set. It was implemented using the KNNImputer\footnote{https://scikit-learn.org/stable/modules/generated/sklearn.impute.KNNImputer.html} of Scikit-learn, and the number of neighbors $k$ is chosen as 5.}
	\item \textcolor{black}{\textbf{MICE with Bayesian Ridge.} This is another data imputation method that is based on MICE \citep{mice}.  For the fitting model of the MICE, a bayesian ridge regression model \citep{tipping2001sparse} was used and implemented using the IterativeImputer\footnote{https://scikit-learn.org/stable/modules/generated/sklearn.impute.IterativeImputer.html} of Scikit-learn and its default parameters are used for training. }
	\item \textcolor{black}{\textbf{MICE with Random Forest~\citep{feng2021imputation}.}
Similar to the previous MICE with Bayesian Ridge, this method employs a random forest model \citep{geurts2006extremely} for the MICE. It fits a number of classifying decision trees on various sub-samples of the dataset and uses averaging to predict the judgment variable. }
	\item \textcolor{black}{\textbf{Linear Regression.}
We train a linear regression model using the observed records excluding the judgment variable and estimate the value of the judgment variable.}
\end{itemize}

\begin{table*}[!t]
			\caption{\textcolor{black}{Classification accuracy with different \textit{Judgment Variable Estimators} for IT Service Data. \textit{Gradient boosting} was used as the \textcolor{black}{target ML model}. Scores in () are the standard deviations, and numbers in \textbf{bold} indicate statistical significance (p-value $< 0.01$).} }
			\label{tab:ablation:gb:itservice}
    \begin{minipage}[!t]{.99\textwidth }
        \centering    
        \scalebox{0.84}{  
			\begin{tabular}{cccc}
			Judgment Variable Estimator & Accuracy              & Closeness              & Combined           \\ \hline
			DRC & 0.800 (0.010)  & \textbf{0.962 (0.031)}          & 0.874 (0.013)\\ \hline
			KNN & \textbf{0.921 (0.006)}  & 0.821 (0.053)          & 0.867 (0.030)\\
			MICE with Bayesian Ridge & 0.919 (0.006)  & 0.823 (0.053)          & 0.867 (0.030)\\
			MICE with Random Forest & \textbf{0.921 (0.006)}  & 0.818 (0.055)         & 0.865 (0.031)\\	
			Linear regression & 0.917 (0.006)  & 0.826 (0.057)          & 0.870 (0.030)\\ \hline
			\mname{} & 0.862 (0.041)  & 0.920 (0.081)          & \textbf{0.889 (0.051)}
			\end{tabular}
			}
		    \hrule height 0pt
		    \end{minipage}%
\end{table*}

\begin{table*}[!t]
			\caption{\textcolor{black}{Classification accuracy with different \textit{Judgment Variable Estimator} for IT Service Data. \textit{Logistic regression} was used as the \textcolor{black}{target ML model}. Scores in () are the standard deviations, and numbers in \textbf{bold} indicate statistical significance (p-value $< 0.01$).} }
			\label{tab:ablation:lr:itservice}
    \begin{minipage}[!t]{.99\textwidth }
        \centering    
        \scalebox{0.84}{  
			\begin{tabular}{cccc}
			Judgment Variable Estimator & Accuracy              & Closeness              & Combined           \\ \hline
			DRC & 0.797 (0.011)  & \textbf{0.960 (0.038)}          & 0.870 (0.016)\\ \hline
			KNN & \textbf{0.921 (0.006)}  & 0.830 (0.060)          & 0.872 (0.032)\\
			MICE with Bayesian Ridge & \textbf{0.919 (0.006)}  & 0.832 (0.058)          & 0.872 (0.031)\\
			MICE with Random Forest & \textbf{0.920 (0.006)}  & 0.830 (0.060)          & 0.871 (0.032)\\
			Linear regression & \textbf{0.917 (0.007)}  & 0.830 (0.059)          & 0.870 (0.029)\\ \hline
			\mname{} &  0.890 (0.016)  & 0.949 (0.011)          & \textbf{0.918 (0.012)}
			\end{tabular}
			}
		    \hrule height 0pt
		    \end{minipage}%
\end{table*}

\textcolor{black}{
Table \ref{tab:ablation:gb:itservice} and \ref{tab:ablation:lr:itservice} show the results when different variants are used for the judgment variable estimator of our proposed framework on the IT service dataset. First, in the case of DRC, the similarity between the training dataset and testing instance is leveraged to determine how to combine the judgment function with predictions. In this case, since the unobserved testing instance often differs from the expected distribution of the training dataset, when applied to our framework, the final prediction is more dependent on the judgment prediction function; thus, when using the DRC, it shows high closeness scores.
Moreover, since the test instance cannot be evaluated in terms of the reliability of the predictions of the trained ML model, it shows low accuracy values and \textit{combined} scores. In the case of other alternatives using the imputation methods above (KNN, MICE with Bayesian Ridge, MICE with random Forest, and Linear Regression), they show low closeness and \textit{combined} scores for the IT service deal prediction. This implies that they are limited in predicting the uncertainty or potential conflict with respect to the judgment variable due to their lower expressive power. However, for both gradient boosting and logistic regression, our \mname{} shows higher \textit{combined} scores than other options. This implies that our \mname{} can better learn complex decision-making processes in the IT deal with respect to the target judgment variable to assist the prediction of the existing ML model.}

\textcolor{black}{
Table \ref{tab:ablation:gb:credit}, and \ref{tab:ablation:lr:credit} show the results for the financial data. As in the previous results, DRC shows high closeness while having low performance in accuracy and \textit{combined} metrics. Data imputation methods show the highest accuracy scores but give low closeness and \textit{combined} scores. On the other hand, our \mname{} shows the highest \textit{combined} scores, so we can expect that the results of our framework are more trustable to experts.
}

\begin{table*}[!t]
			\caption{\textcolor{black}{Classification accuracy with different \textit{judgment Variable Estimator} for Financial Data. \textit{Gradient boosting} was used as the \textcolor{black}{target ML model}. Scores in () are the standard deviations, and numbers in \textbf{bold} indicate statistical significance (p-value $< 0.01$).}}
			\label{tab:ablation:gb:credit}
    \begin{minipage}[!t]{.99\textwidth }
        \centering    
        \scalebox{0.84}{  
			\begin{tabular}{cccc}
			Judgment Variable Estimator & Accuracy              & Closeness              & Combined           \\ \hline
			DRC & 0.724 (0.029)  & \textbf{0.918 (0.019)}          & 0.809 (0.020)\\ \hline
			KNN & 0.730 (0.020)  & 0.898 (0.010)          & 0.805 (0.014)\\
			MICE with Bayesian Ridge & \textbf{0.763 (0.020)}  & 0.857 (0.049)          & 0.806 (0.022)\\
			MICE with Random Forest & \textbf{0.762 (0.020)}  & 0.857 (0.049)          & 0.806 (0.022)\\	
			Linear regression & \textbf{0.761 (0.020)}  & 0.847 (0.080)          & 0.799 (0.034) \\ \hline
			\mname{} & 0.736 (0.025)  & 0.907 (0.024)          & \textbf{0.812 (0.020)}
			\end{tabular}
			}
		    \hrule height 0pt
		    \end{minipage}%
\end{table*}

\begin{table*}[!t]
			\caption{\textcolor{black}{Classification accuracy with different \textit{judgment Variable Estimator} for Financial Data. \textit{Logistic regression} was used as the \textcolor{black}{target ML model}. Scores in () are the standard deviations, and numbers in \textbf{bold} indicate statistical significance (p-value $< 0.01$).} }
			\label{tab:ablation:lr:credit}
    \begin{minipage}[!t]{.99\textwidth }
        \centering    
        \scalebox{0.84}{  
			\begin{tabular}{cccc}
			Judgment Variable Estimator & Accuracy              & Closeness              & Combined           \\ \hline
			DRC & 0.726 (0.018)  & 0.905 (0.016)          & 0.806 (0.014)\\ \hline
			KNN & 0.732 (0.025)  & 0.898 (0.012)          & 0.807 (0.018)\\
			MICE with Bayesian Ridge & \textbf{0.755 (0.020)}  & 0.851 (0.038)          & 0.799 (0.014)\\
			MICE with Random Forest & \textbf{0.754 (0.020)}  & 0.804 (0.054)          & 0.777 (0.021)\\	
			Linear regression & \textbf{0.753 (0.024)}  & 0.837 (0.083)          & 0.790 (0.032)\\ \hline
			\mname{} & 0.743 (0.029)  & \textbf{0.910 (0.024)}          & \textbf{0.818 (0.021)}
			\end{tabular}
			}
		    \hrule height 0pt
		    \end{minipage}
\end{table*}

\subsection{Parameter Sensitivity}\label{sec:param}

\textcolor{black}{In this section, we evaluate how the parameters of our proposed method can influence the performance in \textit{combined} measure. In particular, we evaluate the effect of the number of buckets $q$ when computing the closeness (Eq. \ref{eq:closeness}). The empirical results of Table \ref{tab:params:gb:itservice} and \ref{tab:params:lr:itservice} demonstrate how the performance of our method varies when the number of buckets $q$ varies. Our methods with LR, DRC, and \mname{} perform better than the existing WS, SL, and ER, regardless of the size of $q$ in both gradient boosting and logistic regression models. In particular, as $q$ increases, our method with \mname{} shows better performance than other baselines. We note that the financial dataset is not evaluated for the parameter sensitivity because the possible judgment variables are either 0 or 1.
}

\begin{table*}[!t]
			\caption{\textcolor{black}{Parameter Sensitivity with respect to the size of bucket $q$ in the \textit{Combined} measure for the IT services case study, where \textit{gradient boosting} was used as the \textcolor{black}{target ML model}. Numbers in \textbf{bold} indicate statistical significance (p-value $< 0.01$).}}
			\label{tab:params:gb:itservice}
    \begin{minipage}[!t]{.99\textwidth }
        \centering    
        \scalebox{0.85}{  
			\begin{tabular}{cccccccc}
			& 5             & 7              & 9  & 11 & 13 & 15 & 17         \\ \hline
			WS & 0.740	& 0.745	& 0.755	& 0.756	& 0.757	& 0.756	& 0.757 \\
			SL & 0.858	& 0.853	& 0.855	& 0.845	& 0.799	& 0.805	& 0.853 \\
			ER & 0.740	& 0.745	& 0.755	& 0.756	& 0.757	& 0.756	& 0.757 \\ \hline
			\textbf{Our Method with LR} & 0.888	& 0.883	& 0.890	& 0.876	& 0.865	& 0.865	& 0.880 \\ 			
			\textbf{Our Method with DRC} & \textbf{0.911}	 & 0.870	 & 0.828 	& 0.811	& 0.810	& 0.807	& 0.850 \\
			\textbf{Our Method with \mname{}}  & 0.902	& \textbf{0.911}	& \textbf{0.910}	& \textbf{0.917}	& \textbf{0.905}	& \textbf{0.901}	& \textbf{0.898}
			\end{tabular}
			}
		    \hrule height 0pt
		    \end{minipage}%
\end{table*}

\begin{table*}[!t]
			\caption{\textcolor{black}{Parameter Sensitivity with respect to the size of bucket $q$ in the \textit{Combined} measure for the IT services case study, where \textit{logistic regression} was used as the \textcolor{black}{target ML model}. Numbers in \textbf{bold} indicate statistical significance (p-value $< 0.01$).}}
			\label{tab:params:lr:itservice}
    \begin{minipage}[!t]{.99\textwidth }
        \centering    
        \scalebox{0.85}{  
			\begin{tabular}{cccccccc}
			& 5             & 7              & 9  & 11 & 13 & 15 & 17         \\ \hline
			WS & 0.740	& 0.745	& 0.755	& 0.756	& 0.757	& 0.756	& 0.757 \\
			SL & 0.853	& 0.852	& 0.854	& 0.838	& 0.789	& 0.796	& 0.851 \\
			ER & 0.740	& 0.745	& 0.755	& 0.756	& 0.757	& 0.756	& 0.757 \\ \hline
			\textbf{Our Method with LR} & 0.892	& 0.886	& 0.891	& 0.882	& 0.868	& 0.869	& \textbf{0.882} \\ 			
			\textbf{Our Method with DRC}  & \textbf{0.908}	& 0.869	& 0.827	& 0.810	& 0.808	& 0.818	& 0.845 \\ 
			\textbf{Our Method with \mname{}}  & \textbf{0.908}	& \textbf{0.904}	& \textbf{0.912}	& \textbf{0.913}	& \textbf{0.918}	& \textbf{0.893}	& \textbf{0.882}
			\end{tabular}
			}
		    \hrule height 0pt
		    \end{minipage}%
\end{table*}

\newpage



\section{Discussion}


\textcolor{black}{
In this paper, we present a novel framework for leveraging expert judgment to improve machine learning model predictions when the training data may not be entirely representative of the population. In this discussion section, we further explore the issue of aligning model predictions with expert judgment and propose potential future research directions. We have organized the discussion into three subsections: (1) assessing the quality of expert judgment, (2) incorporating multiple experts' judgment, and (3) learning with high-dimensional data.
}

\subsection{Assessing the quality of expert judgment}
\textcolor{black}{
A crucial aspect of our framework is the extent to which we rely on expert judgment. If the expert's judgment is biased or systematically incorrect \citep{kruglanski1983bias,kahneman2011thinking}, overreliance on it may lead to decreased model performance. To improve the performance of our framework, future research can focus on developing methods to assess the quality of expert judgment objectively. One potential approach is to utilize elicitation techniques \citep{o2006uncertain}, which involve systematically gathering expert judgment and quantifying their associated uncertainties. By evaluating the expert's performance in terms of calibration and discrimination \citep{cooke1991experts}, we can obtain a more comprehensive understanding of the quality of their judgment, which can inform the weighting of their judgment within our framework.}

\textcolor{black}{
Another approach for assessing expert judgment quality is to leverage machine learning techniques to identify patterns in the expert's decision-making process. For example, by training a separate model on the expert's past judgment and the associated features, we can obtain insights into the factors influencing their decisions \citep{james2013introduction,dong2021machine}. Recently, data valuation methods~\citep{yoon2020data} have been proposed to handle noisy data for a target prediction task by leveraging reinforcement learning techniques, and we can selectively use the judgment information by designing proper reward functions. The aforementioned methods can be used to identify potential biases and develop methods to correct them, ultimately improving the performance of our framework.
}

\textcolor{black}{
Lastly, it is important to consider the role of feedback in refining expert judgment over time. By providing experts with feedback on their performance, we can facilitate learning and improvement, which can enhance the quality of their input to our framework. Approaches such as the Delphi method \citep{dalkey1963experimental}, nominal group technique \citep{landeta2011hybrid}, and structured expert judgment process \citep{cooke2008tu} can be employed to iteratively refine expert judgment and ultimately improve the alignment between model predictions and expert judgment.}

\textcolor{black}{
In short, addressing the challenges associated with assessing and incorporating expert judgment quality is an essential aspect of refining our framework. By exploring these potential research directions, we can enhance the adaptability of our framework to varying levels of expert reliability and ultimately improve the performance of our approach in leveraging expert judgment for machine learning model predictions.
}

\subsection{Incorporating multiple experts' judgment}

\textcolor{black}{Another potential limitation of our framework is that it currently relies on a single expert's judgment. Individual experts may have biases, which can affect the quality of the resulting model predictions. To mitigate this issue, future research can explore incorporating multiple experts' opinions into our framework. The main difficulty of incorporating judgment from multiple experts is the existence of potential dependency between them, which is unlikely to improve prediction accuracy \citep{morris1977combining,werner2017expert}. Machine learning methods like Bayesian model averaging \citep{hoeting1999bayesian}, maximum entropy combination \citep{agmon1979algorithm}, and ensemble learning~\citep{zhou2012ensemble} can be employed to combine the judgment of multiple experts, resulting in a more robust and reliable output. The proposed framework could be extended to accommodate multiple experts' judgment and use ensemble learning techniques to derive a corrected prediction that benefits from the collective expertise.}

\subsection{Learning with high-dimensional data}

\textcolor{black}{For future work, our framework can be extended by considering high-dimensional data where the number of judgment variables is large or even greater than the sample size. Many previous works (e.g., \citep{van1999multiple,hardt2012auxiliary,nagarajan2022missing,cahan2022factor}) have examined multiple strategies in situations where a large number of variables exist. They tried to investigate how many variables could be properly considered in the target-supervised ML model. For example, \citep{van1999multiple} suggests that fewer than 15-25 variables are recommended for data imputation scenarios. Moreover, \citep{hardt2012auxiliary} attempted to answer a similar research question. They used existing data imputation methods and observed that the number of variables used for attribution to total cases should not exceed 1/3. Similarly, when learning the ML model in the environment, we need to investigate how we handle judgment variables in a high-dimensional environment.}

\section{Conclusions} \label{Conclusions}

In this work, we incorporate experts' \textcolor{black}{judgment} into machine learning models. We propose a novel framework that allows us to estimate how reliable a data-driven model would be for a given unlabeled data instance. Then, we use such estimation to correct the model's prediction. Moreover, we propose a new evaluation metric that combines prediction accuracy with closeness to the experts' \textcolor{black}{judgment}. Our results on both simulated data and two real-world case studies show the effectiveness of our approach.
The applicability of the proposed method is broad and general. As a framework that can resolve the conflict between ML models and experts' judgment, the proposed method generally provides a great tool to boost the performance of existing methods and make them more trustable.
The ability to achieve both accurate results and the \textit{combined} measure can help various applications, which follow strong statistical or environmental assumptions and have small training datasets (e.g., biology~\citep{sampaio2022exploring, lv2022new,liu2022yeast} and astronomy~\citep{sen2022astronomical,bluck2022quenching}). The proposed method could also be applied to many different ML models, and it is easy to integrate.




\bibliographystyle{model5-names}\biboptions{authoryear}
\bibliography{References}

\begin{thebibliography}{101}
\expandafter\ifx\csname natexlab\endcsname\relax\def\natexlab#1{#1}\fi
\providecommand{\url}[1]{\texttt{#1}}
\providecommand{\href}[2]{#2}
\providecommand{\path}[1]{#1}
\providecommand{\DOIprefix}{doi:}
\providecommand{\ArXivprefix}{arXiv:}
\providecommand{\URLprefix}{URL: }
\providecommand{\Pubmedprefix}{pmid:}
\providecommand{\doi}[1]{\href{http://dx.doi.org/#1}{\path{#1}}}
\providecommand{\Pubmed}[1]{\href{pmid:#1}{\path{#1}}}
\providecommand{\bibinfo}[2]{#2}
\ifx\xfnm\relax \def\xfnm[#1]{\unskip,\space#1}\fi
\bibitem[{Abadi et~al.(2016)Abadi, Barham, Chen, Chen, Davis, Dean, Devin,
  Ghemawat, Irving, Isard et~al.}]{abadi2016tensorflow}
\bibinfo{author}{Abadi, M.}, \bibinfo{author}{Barham, P.},
  \bibinfo{author}{Chen, J.}, \bibinfo{author}{Chen, Z.},
  \bibinfo{author}{Davis, A.}, \bibinfo{author}{Dean, J.},
  \bibinfo{author}{Devin, M.}, \bibinfo{author}{Ghemawat, S.},
  \bibinfo{author}{Irving, G.}, \bibinfo{author}{Isard, M.} et~al.
  (\bibinfo{year}{2016}).
\newblock \bibinfo{title}{Tensorflow: A system for large-scale machine
  learning}.
\newblock In {\it \bibinfo{booktitle}{USENIX Symposium on Operating Systems
  Design and Implementation}\/} (pp. \bibinfo{pages}{265--283}).
\bibitem[{Agmon et~al.(1979)Agmon, Alhassid \& Levine}]{agmon1979algorithm}
\bibinfo{author}{Agmon, N.}, \bibinfo{author}{Alhassid, Y.}, \&
  \bibinfo{author}{Levine, R.~D.} (\bibinfo{year}{1979}).
\newblock \bibinfo{title}{An algorithm for finding the distribution of maximal
  entropy}.
\newblock {\it \bibinfo{journal}{Journal of computational physics}\/},  {\it
  \bibinfo{volume}{30}\/}, \bibinfo{pages}{250--258}.
\bibitem[{Ahn \& Choi(2009)}]{ahn2009conflict}
\bibinfo{author}{Ahn, B.~S.}, \& \bibinfo{author}{Choi, S.~H.}
  (\bibinfo{year}{2009}).
\newblock \bibinfo{title}{Conflict resolution in a knowledge-based system using
  multiple attribute decision-making}.
\newblock {\it \bibinfo{journal}{Expert Systems with Applications}\/},  {\it
  \bibinfo{volume}{36}\/}, \bibinfo{pages}{11552--11558}.
\bibitem[{Altendorf et~al.(2005)Altendorf, Restificar \&
  Dietterich}]{uai2005mono}
\bibinfo{author}{Altendorf, E.~E.}, \bibinfo{author}{Restificar, A.~C.}, \&
  \bibinfo{author}{Dietterich, T.~G.} (\bibinfo{year}{2005}).
\newblock \bibinfo{title}{Learning from sparse data by exploiting monotonicity
  constraints}.
\newblock In {\it \bibinfo{booktitle}{Conference on Uncertainty in Artificial
  Intelligence}\/} (p. \bibinfo{pages}{18–26}).
\bibitem[{Archer \& Wang(1993)}]{archer1993application}
\bibinfo{author}{Archer, N.~P.}, \& \bibinfo{author}{Wang, S.}
  (\bibinfo{year}{1993}).
\newblock \bibinfo{title}{Application of the back propagation neural network
  algorithm with monotonicity constraints for two-group classification
  problems}.
\newblock {\it \bibinfo{journal}{Decision Sciences}\/},  {\it
  \bibinfo{volume}{24}\/}, \bibinfo{pages}{60--75}.
\bibitem[{Azur et~al.(2011)Azur, Stuart, Frangakis \& Leaf}]{mice}
\bibinfo{author}{Azur, M.~J.}, \bibinfo{author}{Stuart, E.~A.},
  \bibinfo{author}{Frangakis, C.}, \& \bibinfo{author}{Leaf, P.~J.}
  (\bibinfo{year}{2011}).
\newblock \bibinfo{title}{Multiple imputation by chained equations: what is it
  and how does it work?}
\newblock {\it \bibinfo{journal}{International journal of methods in
  psychiatric research}\/},  {\it \bibinfo{volume}{20}\/},
  \bibinfo{pages}{40--49}.
\bibitem[{Ben-David(1995)}]{ben1995monotonicity}
\bibinfo{author}{Ben-David, A.} (\bibinfo{year}{1995}).
\newblock \bibinfo{title}{Monotonicity maintenance in information-theoretic
  machine learning algorithms}.
\newblock {\it \bibinfo{journal}{Machine Learning}\/},  {\it
  \bibinfo{volume}{19}\/}, \bibinfo{pages}{29--43}.
\bibitem[{Bluck et~al.(2022)Bluck, Maiolino, Brownson, Conselice, Ellison,
  Piotrowska \& Thorp}]{bluck2022quenching}
\bibinfo{author}{Bluck, A.~F.}, \bibinfo{author}{Maiolino, R.},
  \bibinfo{author}{Brownson, S.}, \bibinfo{author}{Conselice, C.~J.},
  \bibinfo{author}{Ellison, S.~L.}, \bibinfo{author}{Piotrowska, J.~M.}, \&
  \bibinfo{author}{Thorp, M.~D.} (\bibinfo{year}{2022}).
\newblock \bibinfo{title}{The quenching of galaxies, bulges, and disks since
  cosmic noon-a machine learning approach for identifying causality in
  astronomical data}.
\newblock {\it \bibinfo{journal}{Astronomy \& Astrophysics}\/},  {\it
  \bibinfo{volume}{659}\/}, \bibinfo{pages}{A160}.
\bibitem[{Bose \& Hamilton(2019)}]{bose2019compositional}
\bibinfo{author}{Bose, A.}, \& \bibinfo{author}{Hamilton, W.}
  (\bibinfo{year}{2019}).
\newblock \bibinfo{title}{Compositional fairness constraints for graph
  embeddings}.
\newblock In {\it \bibinfo{booktitle}{International Conference on Machine
  Learning}\/} (pp. \bibinfo{pages}{715--724}).
\bibitem[{Bousquet(2008)}]{drc2}
\bibinfo{author}{Bousquet, N.} (\bibinfo{year}{2008}).
\newblock \bibinfo{title}{Diagnostics of prior-data agreement in applied
  bayesian analysis}.
\newblock {\it \bibinfo{journal}{Journal of Applied Statistics}\/},  {\it
  \bibinfo{volume}{35}\/}, \bibinfo{pages}{1011--1029}.
\bibitem[{Brown et~al.(2020)Brown, Mann, Ryder, Subbiah, Kaplan, Dhariwal,
  Neelakantan, Shyam, Sastry, Askell, Agarwal, Herbert-Voss, Krueger, Henighan,
  Child, Ramesh, Ziegler, Wu, Winter, Hesse, Chen, Sigler, Litwin, Gray, Chess,
  Clark, Berner, McCandlish, Radford, Sutskever \& Amodei}]{brown2020language}
\bibinfo{author}{Brown, T.}, \bibinfo{author}{Mann, B.},
  \bibinfo{author}{Ryder, N.}, \bibinfo{author}{Subbiah, M.},
  \bibinfo{author}{Kaplan, J.~D.}, \bibinfo{author}{Dhariwal, P.},
  \bibinfo{author}{Neelakantan, A.}, \bibinfo{author}{Shyam, P.},
  \bibinfo{author}{Sastry, G.}, \bibinfo{author}{Askell, A.},
  \bibinfo{author}{Agarwal, S.}, \bibinfo{author}{Herbert-Voss, A.},
  \bibinfo{author}{Krueger, G.}, \bibinfo{author}{Henighan, T.},
  \bibinfo{author}{Child, R.}, \bibinfo{author}{Ramesh, A.},
  \bibinfo{author}{Ziegler, D.}, \bibinfo{author}{Wu, J.},
  \bibinfo{author}{Winter, C.}, \bibinfo{author}{Hesse, C.},
  \bibinfo{author}{Chen, M.}, \bibinfo{author}{Sigler, E.},
  \bibinfo{author}{Litwin, M.}, \bibinfo{author}{Gray, S.},
  \bibinfo{author}{Chess, B.}, \bibinfo{author}{Clark, J.},
  \bibinfo{author}{Berner, C.}, \bibinfo{author}{McCandlish, S.},
  \bibinfo{author}{Radford, A.}, \bibinfo{author}{Sutskever, I.}, \&
  \bibinfo{author}{Amodei, D.} (\bibinfo{year}{2020}).
\newblock \bibinfo{title}{Language models are few-shot learners}.
\newblock In {\it \bibinfo{booktitle}{Advances in Neural Information Processing
  Systems}\/} (pp. \bibinfo{pages}{1877--1901}).
\bibitem[{Cabitza et~al.(2020)Cabitza, Campagner \& Sconfienza}]{cabitza2020if}
\bibinfo{author}{Cabitza, F.}, \bibinfo{author}{Campagner, A.}, \&
  \bibinfo{author}{Sconfienza, L.~M.} (\bibinfo{year}{2020}).
\newblock \bibinfo{title}{As if sand were stone. new concepts and metrics to
  probe the ground on which to build trustable ai}.
\newblock {\it \bibinfo{journal}{BMC Medical Informatics and Decision
  Making}\/},  {\it \bibinfo{volume}{20}\/}, \bibinfo{pages}{1--21}.
\bibitem[{Cahan et~al.(2023)Cahan, Bai \& Ng}]{cahan2022factor}
\bibinfo{author}{Cahan, E.}, \bibinfo{author}{Bai, J.}, \& \bibinfo{author}{Ng,
  S.} (\bibinfo{year}{2023}).
\newblock \bibinfo{title}{Factor-based imputation of missing values and
  covariances in panel data of large dimensions}.
\newblock {\it \bibinfo{journal}{Journal of Econometrics}\/},  {\it
  \bibinfo{volume}{233}\/}, \bibinfo{pages}{113--131}.
\bibitem[{Cao(2020)}]{cao2020divide}
\bibinfo{author}{Cao, X.} (\bibinfo{year}{2020}).
\newblock \bibinfo{title}{A divide-and-conquer approach to geometric sampling
  for active learning}.
\newblock {\it \bibinfo{journal}{Expert Systems with Applications}\/},  {\it
  \bibinfo{volume}{140}\/}, \bibinfo{pages}{112907}.
\bibitem[{Cooke(1991)}]{cooke1991experts}
\bibinfo{author}{Cooke, R.} (\bibinfo{year}{1991}).
\newblock {\it \bibinfo{title}{Experts in uncertainty: opinion and subjective
  probability in science}\/}.
\newblock \bibinfo{publisher}{Oxford University Press, USA}.
\bibitem[{Cooke \& Goossens(2008)}]{cooke2008tu}
\bibinfo{author}{Cooke, R.~M.}, \& \bibinfo{author}{Goossens, L.~L.}
  (\bibinfo{year}{2008}).
\newblock \bibinfo{title}{Tu delft expert judgment data base}.
\newblock {\it \bibinfo{journal}{Reliability Engineering \& System Safety}\/},
  {\it \bibinfo{volume}{93}\/}, \bibinfo{pages}{657--674}.
\bibitem[{Dalkey \& Helmer(1963)}]{dalkey1963experimental}
\bibinfo{author}{Dalkey, N.}, \& \bibinfo{author}{Helmer, O.}
  (\bibinfo{year}{1963}).
\newblock \bibinfo{title}{An experimental application of the delphi method to
  the use of experts}.
\newblock {\it \bibinfo{journal}{Management science}\/},  {\it
  \bibinfo{volume}{9}\/}, \bibinfo{pages}{458--467}.
\bibitem[{Dietterich et~al.(1997)Dietterich, Lathrop \&
  Lozano-P{\'e}rez}]{dietterich1997solving}
\bibinfo{author}{Dietterich, T.~G.}, \bibinfo{author}{Lathrop, R.~H.}, \&
  \bibinfo{author}{Lozano-P{\'e}rez, T.} (\bibinfo{year}{1997}).
\newblock \bibinfo{title}{Solving the multiple instance problem with
  axis-parallel rectangles}.
\newblock {\it \bibinfo{journal}{Artificial Intelligence}\/},  {\it
  \bibinfo{volume}{89}\/}, \bibinfo{pages}{31--71}.
\bibitem[{Dietvorst et~al.(2018)Dietvorst, Simmons \&
  Massey}]{dietvorst2018overcoming}
\bibinfo{author}{Dietvorst, B.~J.}, \bibinfo{author}{Simmons, J.~P.}, \&
  \bibinfo{author}{Massey, C.} (\bibinfo{year}{2018}).
\newblock \bibinfo{title}{Overcoming algorithm aversion: People will use
  imperfect algorithms if they can (even slightly) modify them}.
\newblock {\it \bibinfo{journal}{Management Science}\/},  {\it
  \bibinfo{volume}{64}\/}, \bibinfo{pages}{1155--1170}.
\bibitem[{Dong et~al.(2021)Dong, Saar-Tsechansky \& Geva}]{dong2021machine}
\bibinfo{author}{Dong, W.}, \bibinfo{author}{Saar-Tsechansky, M.}, \&
  \bibinfo{author}{Geva, T.} (\bibinfo{year}{2021}).
\newblock \bibinfo{title}{A machine learning framework towards transparency in
  experts' decision quality}.
\newblock \URLprefix \url{https://arxiv.org/abs/2110.11425}.
\bibitem[{Druck et~al.(2008)Druck, Mann \& McCallum}]{druck2008learning}
\bibinfo{author}{Druck, G.}, \bibinfo{author}{Mann, G.}, \&
  \bibinfo{author}{McCallum, A.} (\bibinfo{year}{2008}).
\newblock \bibinfo{title}{Learning from labeled features using generalized
  expectation criteria}.
\newblock In {\it \bibinfo{booktitle}{ACM SIGIR Conference on Research and
  Development in Information Retrieval}\/} (pp. \bibinfo{pages}{595--602}).
\bibitem[{Duivesteijn \& Feelders(2008)}]{duivesteijn2008nearest}
\bibinfo{author}{Duivesteijn, W.}, \& \bibinfo{author}{Feelders, A.}
  (\bibinfo{year}{2008}).
\newblock \bibinfo{title}{Nearest neighbour classification with monotonicity
  constraints}.
\newblock In {\it \bibinfo{booktitle}{Joint European Conference on Machine
  Learning and Knowledge Discovery in Databases}\/} (pp.
  \bibinfo{pages}{301--316}).
\newblock \bibinfo{organization}{Springer}.
\bibitem[{D’Acquisto(2020)}]{d2020conflicts}
\bibinfo{author}{D’Acquisto, G.} (\bibinfo{year}{2020}).
\newblock \bibinfo{title}{On conflicts between ethical and logical principles
  in artificial intelligence}.
\newblock {\it \bibinfo{journal}{AI \& SOCIETY}\/},  (pp.
  \bibinfo{pages}{1--6}).
\bibitem[{D’Orazio et~al.(2019)D’Orazio, Honaker, Prasady \&
  Shoemate}]{d2019modeling}
\bibinfo{author}{D’Orazio, V.}, \bibinfo{author}{Honaker, J.},
  \bibinfo{author}{Prasady, R.}, \& \bibinfo{author}{Shoemate, M.}
  (\bibinfo{year}{2019}).
\newblock \bibinfo{title}{Modeling and forecasting armed conflict: Automl with
  human-guided machine learning}.
\newblock In {\it \bibinfo{booktitle}{IEEE International Conference on Big
  Data}\/} (pp. \bibinfo{pages}{4714--4723}).
\bibitem[{Ermon et~al.(2015)Ermon, Bras, Suram, Gregoire, Gomes, Selman \& van
  Dover}]{ermon2015const}
\bibinfo{author}{Ermon, S.}, \bibinfo{author}{Bras, R.~L.},
  \bibinfo{author}{Suram, S.~K.}, \bibinfo{author}{Gregoire, J.~M.},
  \bibinfo{author}{Gomes, C.~P.}, \bibinfo{author}{Selman, B.}, \&
  \bibinfo{author}{van Dover, R.~B.} (\bibinfo{year}{2015}).
\newblock \bibinfo{title}{Pattern decomposition with complex combinatorial
  constraints: Application to materials discovery}.
\newblock In {\it \bibinfo{booktitle}{AAAI Conference on Artificial
  Intelligence}\/} (p. \bibinfo{pages}{636–643}).
\bibitem[{Eves(1963)}]{eves1963survey}
\bibinfo{author}{Eves, H.} (\bibinfo{year}{1963}).
\newblock {\it \bibinfo{title}{A Survey of Geometry, Volume I.}\/}.
\newblock \bibinfo{publisher}{Allyn and Bacon}.
\bibitem[{Feelders \& Pardoel(2003)}]{feelders2003pruning}
\bibinfo{author}{Feelders, A.}, \& \bibinfo{author}{Pardoel, M.}
  (\bibinfo{year}{2003}).
\newblock \bibinfo{title}{Pruning for monotone classification trees}.
\newblock In {\it \bibinfo{booktitle}{International Symposium on Intelligent
  Data Analysis}\/} (pp. \bibinfo{pages}{1--12}).
\newblock \bibinfo{organization}{Springer}.
\bibitem[{Feinman et~al.(2017)Feinman, Curtin, Shintre \&
  Gardner}]{feinman2017detecting}
\bibinfo{author}{Feinman, R.}, \bibinfo{author}{Curtin, R.~R.},
  \bibinfo{author}{Shintre, S.}, \& \bibinfo{author}{Gardner, A.~B.}
  (\bibinfo{year}{2017}).
\newblock \bibinfo{title}{Detecting adversarial samples from artifacts}.
\newblock \URLprefix \url{https://arxiv.org/abs/1703.00410}.
\bibitem[{Feng et~al.(2021)Feng, Grana \& Balling}]{feng2021imputation}
\bibinfo{author}{Feng, R.}, \bibinfo{author}{Grana, D.}, \&
  \bibinfo{author}{Balling, N.} (\bibinfo{year}{2021}).
\newblock \bibinfo{title}{Imputation of missing well log data by random forest
  and its uncertainty analysis}.
\newblock {\it \bibinfo{journal}{Computers \& Geosciences}\/},  {\it
  \bibinfo{volume}{152}\/}, \bibinfo{pages}{104763}.
\bibitem[{Geurts et~al.(2006)Geurts, Ernst \& Wehenkel}]{geurts2006extremely}
\bibinfo{author}{Geurts, P.}, \bibinfo{author}{Ernst, D.}, \&
  \bibinfo{author}{Wehenkel, L.} (\bibinfo{year}{2006}).
\newblock \bibinfo{title}{Extremely randomized trees}.
\newblock {\it \bibinfo{journal}{Machine learning}\/},  {\it
  \bibinfo{volume}{63}\/}, \bibinfo{pages}{3--42}.
\bibitem[{Goodfellow et~al.(2016)Goodfellow, Bengio \&
  Courville}]{goodfellow2016deep}
\bibinfo{author}{Goodfellow, I.}, \bibinfo{author}{Bengio, Y.}, \&
  \bibinfo{author}{Courville, A.} (\bibinfo{year}{2016}).
\newblock {\it \bibinfo{title}{Deep learning}\/}.
\newblock \bibinfo{publisher}{MIT press}.
\bibitem[{Goodfellow et~al.(2014)Goodfellow, Pouget-Abadie, Mirza, Xu,
  Warde-Farley, Ozair, Courville \& Bengio}]{gan}
\bibinfo{author}{Goodfellow, I.}, \bibinfo{author}{Pouget-Abadie, J.},
  \bibinfo{author}{Mirza, M.}, \bibinfo{author}{Xu, B.},
  \bibinfo{author}{Warde-Farley, D.}, \bibinfo{author}{Ozair, S.},
  \bibinfo{author}{Courville, A.}, \& \bibinfo{author}{Bengio, Y.}
  (\bibinfo{year}{2014}).
\newblock \bibinfo{title}{Generative adversarial nets}.
\newblock In {\it \bibinfo{booktitle}{AAAI Conference on Artificial
  Intelligence}\/}.
\bibitem[{Grandvalet \& Bengio(2005)}]{grandvalet2005semi}
\bibinfo{author}{Grandvalet, Y.}, \& \bibinfo{author}{Bengio, Y.}
  (\bibinfo{year}{2005}).
\newblock \bibinfo{title}{Semi-supervised learning by entropy minimization}.
\newblock In {\it \bibinfo{booktitle}{Advances in Neural Information Processing
  Systems}\/} (pp. \bibinfo{pages}{529--536}).
\bibitem[{Guo et~al.(2019)Guo, Megahed, Asthana \&
  Messinger}]{guo2019winnability}
\bibinfo{author}{Guo, P.}, \bibinfo{author}{Megahed, A.},
  \bibinfo{author}{Asthana, S.}, \& \bibinfo{author}{Messinger, P.}
  (\bibinfo{year}{2019}).
\newblock \bibinfo{title}{Winnability prediction for {IT} services bids}.
\newblock In {\it \bibinfo{booktitle}{IEEE International Conference on Services
  Computing}\/} (pp. \bibinfo{pages}{237--239}).
\bibitem[{Hardt et~al.(2012)Hardt, Herke \& Leonhart}]{hardt2012auxiliary}
\bibinfo{author}{Hardt, J.}, \bibinfo{author}{Herke, M.}, \&
  \bibinfo{author}{Leonhart, R.} (\bibinfo{year}{2012}).
\newblock \bibinfo{title}{Auxiliary variables in multiple imputation in
  regression with missing x: a warning against including too many in small
  sample research}.
\newblock {\it \bibinfo{journal}{BMC medical research methodology}\/},  {\it
  \bibinfo{volume}{12}\/}, \bibinfo{pages}{1--13}.
\bibitem[{Hecht(1998)}]{hecht1998optics}
\bibinfo{author}{Hecht, E.} (\bibinfo{year}{1998}).
\newblock \bibinfo{title}{Optics}.
\newblock {\it \bibinfo{journal}{Addison Wesley Longman Inc}\/},  {\it
  \bibinfo{volume}{1}\/}.
\bibitem[{Hendrycks \& Gimpel(2017)}]{hendrycks2016baseline}
\bibinfo{author}{Hendrycks, D.}, \& \bibinfo{author}{Gimpel, K.}
  (\bibinfo{year}{2017}).
\newblock \bibinfo{title}{A baseline for detecting misclassified and
  out-of-distribution examples in neural networks}.
\newblock In {\it \bibinfo{booktitle}{International Conference on Learning
  Representations}\/}.
\bibitem[{Hoeting et~al.(1999)Hoeting, Madigan, Raftery \&
  Volinsky}]{hoeting1999bayesian}
\bibinfo{author}{Hoeting, J.~A.}, \bibinfo{author}{Madigan, D.},
  \bibinfo{author}{Raftery, A.~E.}, \& \bibinfo{author}{Volinsky, C.~T.}
  (\bibinfo{year}{1999}).
\newblock \bibinfo{title}{Bayesian model averaging: a tutorial (with comments
  by m. clyde, david draper and ei george, and a rejoinder by the authors}.
\newblock {\it \bibinfo{journal}{Statistical science}\/},  {\it
  \bibinfo{volume}{14}\/}, \bibinfo{pages}{382--417}.
\bibitem[{Israeli et~al.(2019)Israeli, Rokach \&
  Shabtai}]{israeli2019constraint}
\bibinfo{author}{Israeli, A.}, \bibinfo{author}{Rokach, L.}, \&
  \bibinfo{author}{Shabtai, A.} (\bibinfo{year}{2019}).
\newblock \bibinfo{title}{Constraint learning based gradient boosting trees}.
\newblock {\it \bibinfo{journal}{Expert Systems with Applications}\/},  {\it
  \bibinfo{volume}{128}\/}, \bibinfo{pages}{287--300}.
\bibitem[{James et~al.(2013)James, Witten, Hastie \&
  Tibshirani}]{james2013introduction}
\bibinfo{author}{James, G.}, \bibinfo{author}{Witten, D.},
  \bibinfo{author}{Hastie, T.}, \& \bibinfo{author}{Tibshirani, R.}
  (\bibinfo{year}{2013}).
\newblock {\it \bibinfo{title}{An introduction to statistical learning}\/}.
\newblock \bibinfo{publisher}{Springer}.
\bibitem[{Jiang et~al.(2018)Jiang, Liang, Gao, Guo, Zhong, Yang \&
  Liu}]{jiang2018improved}
\bibinfo{author}{Jiang, Y.}, \bibinfo{author}{Liang, Z.}, \bibinfo{author}{Gao,
  H.}, \bibinfo{author}{Guo, Y.}, \bibinfo{author}{Zhong, Z.},
  \bibinfo{author}{Yang, C.}, \& \bibinfo{author}{Liu, J.}
  (\bibinfo{year}{2018}).
\newblock \bibinfo{title}{An improved constraint-based bayesian network
  learning method using gaussian kernel probability density estimator}.
\newblock {\it \bibinfo{journal}{Expert Systems with Applications}\/},  {\it
  \bibinfo{volume}{113}\/}, \bibinfo{pages}{544--554}.
\bibitem[{Kahneman(2011)}]{kahneman2011thinking}
\bibinfo{author}{Kahneman, D.} (\bibinfo{year}{2011}).
\newblock {\it \bibinfo{title}{Thinking, fast and slow}\/}.
\newblock \bibinfo{publisher}{Farrar, Straus and Giroux}.
\bibitem[{Kotzias et~al.(2015)Kotzias, Denil, De~Freitas \&
  Smyth}]{kotzias2015group}
\bibinfo{author}{Kotzias, D.}, \bibinfo{author}{Denil, M.},
  \bibinfo{author}{De~Freitas, N.}, \& \bibinfo{author}{Smyth, P.}
  (\bibinfo{year}{2015}).
\newblock \bibinfo{title}{From group to individual labels using deep features}.
\newblock In {\it \bibinfo{booktitle}{SIGKDD International Conference on
  Knowledge Discovery and Data Mining}\/} (pp. \bibinfo{pages}{597--606}).
\bibitem[{Kruglanski \& Ajzen(1983)}]{kruglanski1983bias}
\bibinfo{author}{Kruglanski, A.~W.}, \& \bibinfo{author}{Ajzen, I.}
  (\bibinfo{year}{1983}).
\newblock \bibinfo{title}{Bias and error in human judgment}.
\newblock {\it \bibinfo{journal}{European Journal of Social Psychology}\/},
  {\it \bibinfo{volume}{13}\/}, \bibinfo{pages}{1--44}.
\bibitem[{Lan et~al.(2020)Lan, Xu, Ma \& Li}]{lan2020multivariable}
\bibinfo{author}{Lan, Q.}, \bibinfo{author}{Xu, X.}, \bibinfo{author}{Ma, H.},
  \& \bibinfo{author}{Li, G.} (\bibinfo{year}{2020}).
\newblock \bibinfo{title}{Multivariable data imputation for the analysis of
  incomplete credit data}.
\newblock {\it \bibinfo{journal}{Expert Systems with Applications}\/},  {\it
  \bibinfo{volume}{141}\/}, \bibinfo{pages}{112926}.
\bibitem[{Landeta et~al.(2011)Landeta, Barrutia \&
  Lertxundi}]{landeta2011hybrid}
\bibinfo{author}{Landeta, J.}, \bibinfo{author}{Barrutia, J.}, \&
  \bibinfo{author}{Lertxundi, A.} (\bibinfo{year}{2011}).
\newblock \bibinfo{title}{Hybrid delphi: A methodology to facilitate
  contribution from experts in professional contexts}.
\newblock {\it \bibinfo{journal}{Technological Forecasting and Social
  Change}\/},  {\it \bibinfo{volume}{78}\/}, \bibinfo{pages}{1629--1641}.
\bibitem[{Liang et~al.(2018)Liang, Li \& Srikant}]{liang2017enhancing}
\bibinfo{author}{Liang, S.}, \bibinfo{author}{Li, Y.}, \&
  \bibinfo{author}{Srikant, R.} (\bibinfo{year}{2018}).
\newblock \bibinfo{title}{Enhancing the reliability of out-of-distribution
  image detection in neural networks}.
\newblock In {\it \bibinfo{booktitle}{International Conference on Learning
  Representations}\/}.
\bibitem[{Lin et~al.(2016)Lin, Lu, Chen \& Zhou}]{lin2016learning}
\bibinfo{author}{Lin, K.}, \bibinfo{author}{Lu, J.}, \bibinfo{author}{Chen,
  C.-S.}, \& \bibinfo{author}{Zhou, J.} (\bibinfo{year}{2016}).
\newblock \bibinfo{title}{Learning compact binary descriptors with unsupervised
  deep neural networks}.
\newblock In {\it \bibinfo{booktitle}{IEEE Conference on Computer Vision and
  Pattern Recognition}\/} (pp. \bibinfo{pages}{1183--1192}).
\bibitem[{Liu et~al.(2022)Liu, Wang \& Nielsen}]{liu2022yeast}
\bibinfo{author}{Liu, Z.}, \bibinfo{author}{Wang, J.}, \&
  \bibinfo{author}{Nielsen, J.} (\bibinfo{year}{2022}).
\newblock \bibinfo{title}{Yeast synthetic biology advances biofuel production}.
\newblock {\it \bibinfo{journal}{Current Opinion in Microbiology}\/},  {\it
  \bibinfo{volume}{65}\/}, \bibinfo{pages}{33--39}.
\bibitem[{Luo et~al.(2017)Luo, Zou, Hoffman \& Fei-Fei}]{luo2017label}
\bibinfo{author}{Luo, Z.}, \bibinfo{author}{Zou, Y.}, \bibinfo{author}{Hoffman,
  J.}, \& \bibinfo{author}{Fei-Fei, L.~F.} (\bibinfo{year}{2017}).
\newblock \bibinfo{title}{Label efficient learning of transferable
  representations acrosss domains and tasks}.
\newblock In {\it \bibinfo{booktitle}{Advances in Neural Information Processing
  Systems}\/} (pp. \bibinfo{pages}{165--177}).
\bibitem[{Lv et~al.(2022)Lv, Hueso-Gil, Bi, Wu, Liu, Liu \&
  Ledesma-Amaro}]{lv2022new}
\bibinfo{author}{Lv, X.}, \bibinfo{author}{Hueso-Gil, A.}, \bibinfo{author}{Bi,
  X.}, \bibinfo{author}{Wu, Y.}, \bibinfo{author}{Liu, Y.},
  \bibinfo{author}{Liu, L.}, \& \bibinfo{author}{Ledesma-Amaro, R.}
  (\bibinfo{year}{2022}).
\newblock \bibinfo{title}{New synthetic biology tools for metabolic control}.
\newblock {\it \bibinfo{journal}{Current Opinion in Biotechnology}\/},  {\it
  \bibinfo{volume}{76}\/}, \bibinfo{pages}{102724}.
\bibitem[{Mahmoudi et~al.(2019)Mahmoudi, Ezzat \&
  Elwany}]{mahmoudi2019layerwise}
\bibinfo{author}{Mahmoudi, M.}, \bibinfo{author}{Ezzat, A.~A.}, \&
  \bibinfo{author}{Elwany, A.} (\bibinfo{year}{2019}).
\newblock \bibinfo{title}{Layerwise anomaly detection in laser powder-bed
  fusion metal additive manufacturing}.
\newblock {\it \bibinfo{journal}{Journal of Manufacturing Science and
  Engineering}\/},  {\it \bibinfo{volume}{141}\/}.
\bibitem[{Mann \& McCallum(2007)}]{mann2007simple}
\bibinfo{author}{Mann, G.~S.}, \& \bibinfo{author}{McCallum, A.}
  (\bibinfo{year}{2007}).
\newblock \bibinfo{title}{Simple, robust, scalable semi-supervised learning via
  expectation regularization}.
\newblock In {\it \bibinfo{booktitle}{International Conference on Machine
  Learning}\/} (pp. \bibinfo{pages}{593--600}).
\bibitem[{Manning \& Schutze(1999)}]{manning1999foundations}
\bibinfo{author}{Manning, C.}, \& \bibinfo{author}{Schutze, H.}
  (\bibinfo{year}{1999}).
\newblock {\it \bibinfo{title}{Foundations of statistical natural language
  processing}\/}.
\newblock \bibinfo{publisher}{MIT Press}.
\bibitem[{Megahed et~al.(2020)Megahed, Nakamura, Smith, Asthana, Rose,
  Daczkowska \& Gopisetty}]{megahed2020analytics}
\bibinfo{author}{Megahed, A.}, \bibinfo{author}{Nakamura, T.},
  \bibinfo{author}{Smith, M.}, \bibinfo{author}{Asthana, S.},
  \bibinfo{author}{Rose, M.}, \bibinfo{author}{Daczkowska, M.}, \&
  \bibinfo{author}{Gopisetty, S.} (\bibinfo{year}{2020}).
\newblock \bibinfo{title}{Analytics and operations research increases win rates
  for {IBM’s} information technology service deals}.
\newblock {\it \bibinfo{journal}{INFORMS Journal on Applied Analytics}\/},
  {\it \bibinfo{volume}{50}\/}, \bibinfo{pages}{50--63}.
\bibitem[{Megahed et~al.(2015)Megahed, Ren \& Firth}]{megahed2015modeling}
\bibinfo{author}{Megahed, A.}, \bibinfo{author}{Ren, G.-J.}, \&
  \bibinfo{author}{Firth, M.} (\bibinfo{year}{2015}).
\newblock \bibinfo{title}{Modeling business insights into predictive analytics
  for the outcome of it service contracts}.
\newblock In {\it \bibinfo{booktitle}{IEEE International Conference on Services
  Computing}\/} (pp. \bibinfo{pages}{515--521}).
\bibitem[{Miao et~al.(2017{\natexlab{a}})Miao, Li, Davis \&
  Deshpande}]{miao2017model}
\bibinfo{author}{Miao, H.}, \bibinfo{author}{Li, A.}, \bibinfo{author}{Davis,
  L.~S.}, \& \bibinfo{author}{Deshpande, A.}
  (\bibinfo{year}{2017}{\natexlab{a}}).
\newblock \bibinfo{title}{On model discovery for hosted data science projects}.
\newblock In {\it \bibinfo{booktitle}{Workshop on Data Management for
  End-to-End Machine Learning}\/} (pp. \bibinfo{pages}{1--4}).
\bibitem[{Miao et~al.(2017{\natexlab{b}})Miao, Li, Davis \&
  Deshpande}]{miao2017towards}
\bibinfo{author}{Miao, H.}, \bibinfo{author}{Li, A.}, \bibinfo{author}{Davis,
  L.~S.}, \& \bibinfo{author}{Deshpande, A.}
  (\bibinfo{year}{2017}{\natexlab{b}}).
\newblock \bibinfo{title}{Towards unified data and lifecycle management for
  deep learning}.
\newblock In {\it \bibinfo{booktitle}{IEEE International Conference on Data
  Engineering}\/} (pp. \bibinfo{pages}{571--582}).
\bibitem[{Mikolov et~al.(2013)Mikolov, Sutskever, Chen, Corrado \&
  Dean}]{mikolov2013distributed}
\bibinfo{author}{Mikolov, T.}, \bibinfo{author}{Sutskever, I.},
  \bibinfo{author}{Chen, K.}, \bibinfo{author}{Corrado, G.~S.}, \&
  \bibinfo{author}{Dean, J.} (\bibinfo{year}{2013}).
\newblock \bibinfo{title}{Distributed representations of words and phrases and
  their compositionality}.
\newblock In {\it \bibinfo{booktitle}{Advances in Neural Information Processing
  Systems}\/} (pp. \bibinfo{pages}{3111--3119}).
\bibitem[{Morris(1977)}]{morris1977combining}
\bibinfo{author}{Morris, P.~A.} (\bibinfo{year}{1977}).
\newblock \bibinfo{title}{Combining expert judgments: A bayesian approach}.
\newblock {\it \bibinfo{journal}{Management Science}\/},  {\it
  \bibinfo{volume}{23}\/}, \bibinfo{pages}{679--693}.
\bibitem[{Nagarajan \& Babu(2022)}]{nagarajan2022missing}
\bibinfo{author}{Nagarajan, G.}, \& \bibinfo{author}{Babu, L.~D.}
  (\bibinfo{year}{2022}).
\newblock \bibinfo{title}{Missing data imputation on biomedical data using
  deeply learned clustering and l2 regularized regression based on symmetric
  uncertainty}.
\newblock {\it \bibinfo{journal}{Artificial Intelligence in Medicine}\/},  {\it
  \bibinfo{volume}{123}\/}, \bibinfo{pages}{102214}.
\bibitem[{Niculescu et~al.(2006)Niculescu, Mitchell \&
  Rao}]{niculescu2006bayesian}
\bibinfo{author}{Niculescu, R.~S.}, \bibinfo{author}{Mitchell, T.~M.}, \&
  \bibinfo{author}{Rao, R.~B.} (\bibinfo{year}{2006}).
\newblock \bibinfo{title}{Bayesian network learning with parameter
  constraints}.
\newblock {\it \bibinfo{journal}{Journal of Machine Learning Research}\/},
  {\it \bibinfo{volume}{7}\/}, \bibinfo{pages}{1357--1383}.
\bibitem[{Nourani et~al.(2020)Nourani, King \& Ragan}]{nourani2020role}
\bibinfo{author}{Nourani, M.}, \bibinfo{author}{King, J.}, \&
  \bibinfo{author}{Ragan, E.} (\bibinfo{year}{2020}).
\newblock \bibinfo{title}{The role of domain expertise in user trust and the
  impact of first impressions with intelligent systems, 8(1)}.
\newblock In {\it \bibinfo{booktitle}{AAAI Conference on Human Computation and
  Crowdsourcing, 8(1)}\/} (pp. \bibinfo{pages}{112--121}).
\bibitem[{O'Hagan et~al.(2006)O'Hagan, Buck, Daneshkhah, Eiser, Garthwaite,
  Jenkinson, Oakley \& Rakow}]{o2006uncertain}
\bibinfo{author}{O'Hagan, A.}, \bibinfo{author}{Buck, C.~E.},
  \bibinfo{author}{Daneshkhah, A.}, \bibinfo{author}{Eiser, J.~R.},
  \bibinfo{author}{Garthwaite, P.~H.}, \bibinfo{author}{Jenkinson, D.~J.},
  \bibinfo{author}{Oakley, J.~E.}, \& \bibinfo{author}{Rakow, T.}
  (\bibinfo{year}{2006}).
\newblock {\it \bibinfo{title}{Uncertain judgements: eliciting experts'
  probabilities}\/}.
\newblock \bibinfo{publisher}{John Wiley \& Sons}.
\bibitem[{Pan \& Yang(2009)}]{pan2009survey}
\bibinfo{author}{Pan, S.~J.}, \& \bibinfo{author}{Yang, Q.}
  (\bibinfo{year}{2009}).
\newblock \bibinfo{title}{A survey on transfer learning}.
\newblock {\it \bibinfo{journal}{IEEE Transactions on Knowledge and Data
  Engineering}\/},  {\it \bibinfo{volume}{22}\/}, \bibinfo{pages}{1345--1359}.
\bibitem[{Park \& Kim(2019)}]{park2019active}
\bibinfo{author}{Park, S.~H.}, \& \bibinfo{author}{Kim, S.~B.}
  (\bibinfo{year}{2019}).
\newblock \bibinfo{title}{Active semi-supervised learning with multiple
  complementary information}.
\newblock {\it \bibinfo{journal}{Expert Systems with Applications}\/},  {\it
  \bibinfo{volume}{126}\/}, \bibinfo{pages}{30--40}.
\bibitem[{Pedregosa et~al.(2011)Pedregosa, Varoquaux, Gramfort, Michel,
  Thirion, Grisel, Blondel, Prettenhofer, Weiss, Dubourg
  et~al.}]{pedregosa2011scikit}
\bibinfo{author}{Pedregosa, F.}, \bibinfo{author}{Varoquaux, G.},
  \bibinfo{author}{Gramfort, A.}, \bibinfo{author}{Michel, V.},
  \bibinfo{author}{Thirion, B.}, \bibinfo{author}{Grisel, O.},
  \bibinfo{author}{Blondel, M.}, \bibinfo{author}{Prettenhofer, P.},
  \bibinfo{author}{Weiss, R.}, \bibinfo{author}{Dubourg, V.} et~al.
  (\bibinfo{year}{2011}).
\newblock \bibinfo{title}{Scikit-learn: Machine learning in python}.
\newblock {\it \bibinfo{journal}{Journal of Machine Learning Research}\/},
  {\it \bibinfo{volume}{12}\/}, \bibinfo{pages}{2825--2830}.
\bibitem[{Potharst \& Bioch(2000)}]{potharst2000decision}
\bibinfo{author}{Potharst, R.}, \& \bibinfo{author}{Bioch, J.~C.}
  (\bibinfo{year}{2000}).
\newblock \bibinfo{title}{Decision trees for ordinal classification}.
\newblock {\it \bibinfo{journal}{Intelligent Data Analysis}\/},  {\it
  \bibinfo{volume}{4}\/}, \bibinfo{pages}{97--111}.
\bibitem[{Poulis \& Dasgupta(2017)}]{poulis2017learning}
\bibinfo{author}{Poulis, S.}, \& \bibinfo{author}{Dasgupta, S.}
  (\bibinfo{year}{2017}).
\newblock \bibinfo{title}{Learning with feature feedback: from theory to
  practice}.
\newblock In {\it \bibinfo{booktitle}{Artificial Intelligence and
  Statistics}\/} (pp. \bibinfo{pages}{1104--1113}).
\bibitem[{Purwar \& Singh(2015)}]{purwar2015hybrid}
\bibinfo{author}{Purwar, A.}, \& \bibinfo{author}{Singh, S.~K.}
  (\bibinfo{year}{2015}).
\newblock \bibinfo{title}{Hybrid prediction model with missing value imputation
  for medical data}.
\newblock {\it \bibinfo{journal}{Expert Systems with Applications}\/},  {\it
  \bibinfo{volume}{42}\/}, \bibinfo{pages}{5621--5631}.
\bibitem[{Rahman et~al.(2019)Rahman, Surma, Backes \&
  Zhang}]{rahman2019fairwalk}
\bibinfo{author}{Rahman, T.~A.}, \bibinfo{author}{Surma, B.},
  \bibinfo{author}{Backes, M.}, \& \bibinfo{author}{Zhang, Y.}
  (\bibinfo{year}{2019}).
\newblock \bibinfo{title}{Fairwalk: Towards fair graph embedding.}
\newblock In {\it \bibinfo{booktitle}{International Joint Conference on
  Artificial Intelligence}\/} (pp. \bibinfo{pages}{3289--3295}).
\bibitem[{Rudin(2019)}]{rudin2019stop}
\bibinfo{author}{Rudin, C.} (\bibinfo{year}{2019}).
\newblock \bibinfo{title}{Stop explaining black box machine learning models for
  high stakes decisions and use interpretable models instead}.
\newblock {\it \bibinfo{journal}{Nature Machine Intelligence}\/},  {\it
  \bibinfo{volume}{1}\/}, \bibinfo{pages}{206--215}.
\bibitem[{Salaken et~al.(2019)Salaken, Khosravi, Nguyen \&
  Nahavandi}]{salaken2019seeded}
\bibinfo{author}{Salaken, S.~M.}, \bibinfo{author}{Khosravi, A.},
  \bibinfo{author}{Nguyen, T.}, \& \bibinfo{author}{Nahavandi, S.}
  (\bibinfo{year}{2019}).
\newblock \bibinfo{title}{Seeded transfer learning for regression problems with
  deep learning}.
\newblock {\it \bibinfo{journal}{Expert Systems with Applications}\/},  {\it
  \bibinfo{volume}{115}\/}, \bibinfo{pages}{565--577}.
\bibitem[{Sampaio et~al.(2022)Sampaio, Rocha \& Dias}]{sampaio2022exploring}
\bibinfo{author}{Sampaio, M.}, \bibinfo{author}{Rocha, M.}, \&
  \bibinfo{author}{Dias, O.} (\bibinfo{year}{2022}).
\newblock \bibinfo{title}{Exploring synergies between plant metabolic modelling
  and machine learning}.
\newblock {\it \bibinfo{journal}{Computational and Structural Biotechnology
  Journal}\/},  {\it \bibinfo{volume}{20}\/}, \bibinfo{pages}{1885--1900}.
\bibitem[{Schat et~al.(2020)Schat, van~de Schoot, Kouw, Veen \& Mendrik}]{drc1}
\bibinfo{author}{Schat, E.}, \bibinfo{author}{van~de Schoot, R.},
  \bibinfo{author}{Kouw, W.~M.}, \bibinfo{author}{Veen, D.}, \&
  \bibinfo{author}{Mendrik, A.~M.} (\bibinfo{year}{2020}).
\newblock \bibinfo{title}{The data representativeness criterion: Predicting the
  performance of supervised classification based on data set similarity}.
\newblock {\it \bibinfo{journal}{Plos one}\/},  {\it \bibinfo{volume}{15}\/},
  \bibinfo{pages}{e0237009}.
\bibitem[{Sen et~al.(2022)Sen, Agarwal, Chakraborty \&
  Singh}]{sen2022astronomical}
\bibinfo{author}{Sen, S.}, \bibinfo{author}{Agarwal, S.},
  \bibinfo{author}{Chakraborty, P.}, \& \bibinfo{author}{Singh, K.~P.}
  (\bibinfo{year}{2022}).
\newblock \bibinfo{title}{Astronomical big data processing using machine
  learning: A comprehensive review}.
\newblock {\it \bibinfo{journal}{Experimental Astronomy}\/},  (pp.
  \bibinfo{pages}{1--43}).
\bibitem[{Settles(2011)}]{settles2011theories}
\bibinfo{author}{Settles, B.} (\bibinfo{year}{2011}).
\newblock \bibinfo{title}{From theories to queries: Active learning in
  practice}.
\newblock In {\it \bibinfo{booktitle}{Active Learning and Experimental Design
  workshop In conjunction with AISTATS 2010}\/} (pp. \bibinfo{pages}{1--18}).
\bibitem[{Sill(1998)}]{sill1998monotonic}
\bibinfo{author}{Sill, J.} (\bibinfo{year}{1998}).
\newblock \bibinfo{title}{Monotonic networks}.
\newblock In {\it \bibinfo{booktitle}{Advances in Neural Information Processing
  Systems}\/} (pp. \bibinfo{pages}{661--667}).
\bibitem[{Singh et~al.(2014)Singh, Riedel, Hewitt \&
  Rockt{\"a}schel}]{singh2014designing}
\bibinfo{author}{Singh, S.}, \bibinfo{author}{Riedel, S.},
  \bibinfo{author}{Hewitt, L.}, \& \bibinfo{author}{Rockt{\"a}schel, T.}
  (\bibinfo{year}{2014}).
\newblock \bibinfo{title}{Designing an {IDE} for probabilistic programming:
  Challenges and a prototype}.
\newblock In {\it \bibinfo{booktitle}{Advances in Neural Information Processing
  Systems Workshop on Probabilistic Programming}\/}.
\bibitem[{Sparks et~al.(2015)Sparks, Talwalkar, Haas, Franklin, Jordan \&
  Kraska}]{sparks2015automating}
\bibinfo{author}{Sparks, E.~R.}, \bibinfo{author}{Talwalkar, A.},
  \bibinfo{author}{Haas, D.}, \bibinfo{author}{Franklin, M.~J.},
  \bibinfo{author}{Jordan, M.~I.}, \& \bibinfo{author}{Kraska, T.}
  (\bibinfo{year}{2015}).
\newblock \bibinfo{title}{Automating model search for large scale machine
  learning}.
\newblock In {\it \bibinfo{booktitle}{ACM Symposium on Cloud Computing}\/} (pp.
  \bibinfo{pages}{368--380}).
\bibitem[{Stewart \& Ermon(2017)}]{stewart2017label}
\bibinfo{author}{Stewart, R.}, \& \bibinfo{author}{Ermon, S.}
  (\bibinfo{year}{2017}).
\newblock \bibinfo{title}{Label-free supervision of neural networks with
  physics and domain knowledge}.
\newblock In {\it \bibinfo{booktitle}{AAAI Conference on Artificial
  Intelligence}\/}.
\bibitem[{Taha \& Hanbury(2015)}]{taha2015metrics}
\bibinfo{author}{Taha, A.~A.}, \& \bibinfo{author}{Hanbury, A.}
  (\bibinfo{year}{2015}).
\newblock \bibinfo{title}{Metrics for evaluating 3d medical image segmentation:
  analysis, selection, and tool}.
\newblock {\it \bibinfo{journal}{BMC medical imaging}\/},  {\it
  \bibinfo{volume}{15}\/}, \bibinfo{pages}{1--28}.
\bibitem[{Tapia et~al.(2016)Tapia, Elwany \& Sang}]{tapia2016prediction}
\bibinfo{author}{Tapia, G.}, \bibinfo{author}{Elwany, A.}, \&
  \bibinfo{author}{Sang, H.} (\bibinfo{year}{2016}).
\newblock \bibinfo{title}{Prediction of porosity in metal-based additive
  manufacturing using spatial gaussian process models}.
\newblock {\it \bibinfo{journal}{Additive Manufacturing}\/},  {\it
  \bibinfo{volume}{12}\/}, \bibinfo{pages}{282--290}.
\bibitem[{Tipping(2001)}]{tipping2001sparse}
\bibinfo{author}{Tipping, M.~E.} (\bibinfo{year}{2001}).
\newblock \bibinfo{title}{Sparse bayesian learning and the relevance vector
  machine}.
\newblock {\it \bibinfo{journal}{Journal of machine learning research}\/},
  {\it \bibinfo{volume}{1}\/}, \bibinfo{pages}{211--244}.
\bibitem[{Tong \& Koller(2001)}]{tong2001support}
\bibinfo{author}{Tong, S.}, \& \bibinfo{author}{Koller, D.}
  (\bibinfo{year}{2001}).
\newblock \bibinfo{title}{Support vector machine active learning with
  applications to text classification}.
\newblock {\it \bibinfo{journal}{Journal of Machine Learning Research}\/},
  {\it \bibinfo{volume}{2}\/}, \bibinfo{pages}{45--66}.
\bibitem[{Trittenbach et~al.(2020)Trittenbach, Englhardt \&
  B{\"o}hm}]{trittenbach2020overview}
\bibinfo{author}{Trittenbach, H.}, \bibinfo{author}{Englhardt, A.}, \&
  \bibinfo{author}{B{\"o}hm, K.} (\bibinfo{year}{2020}).
\newblock \bibinfo{title}{An overview and a benchmark of active learning for
  outlier detection with one-class classifiers}.
\newblock {\it \bibinfo{journal}{Expert Systems with Applications}\/},  (p.
  \bibinfo{pages}{114372}).
\bibitem[{Van~Buuren et~al.(1999)Van~Buuren, Boshuizen \&
  Knook}]{van1999multiple}
\bibinfo{author}{Van~Buuren, S.}, \bibinfo{author}{Boshuizen, H.~C.}, \&
  \bibinfo{author}{Knook, D.~L.} (\bibinfo{year}{1999}).
\newblock \bibinfo{title}{Multiple imputation of missing blood pressure
  covariates in survival analysis}.
\newblock {\it \bibinfo{journal}{Statistics in medicine}\/},  {\it
  \bibinfo{volume}{18}\/}, \bibinfo{pages}{681--694}.
\bibitem[{Varberg \& Roberts(1973)}]{varberg1973convex}
\bibinfo{author}{Varberg, D.~E.}, \& \bibinfo{author}{Roberts, A.~W.}
  (\bibinfo{year}{1973}).
\newblock {\it \bibinfo{title}{Convex functions}\/}.
\newblock \bibinfo{publisher}{Academic Press}.
\bibitem[{Vartak et~al.(2015)Vartak, Ortiz, Siegel, Subramanyam, Madden \&
  Zaharia}]{vartak2015supporting}
\bibinfo{author}{Vartak, M.}, \bibinfo{author}{Ortiz, P.},
  \bibinfo{author}{Siegel, K.}, \bibinfo{author}{Subramanyam, H.},
  \bibinfo{author}{Madden, S.}, \& \bibinfo{author}{Zaharia, M.}
  (\bibinfo{year}{2015}).
\newblock \bibinfo{title}{Supporting fast iteration in model building}.
\newblock In {\it \bibinfo{booktitle}{Advances in Neural Information Processing
  Systems Workshop on Machine Learning Systems}\/}.
\bibitem[{Vartak et~al.(2016)Vartak, Subramanyam, Lee, Viswanathan, Husnoo,
  Madden \& Zaharia}]{modeldb}
\bibinfo{author}{Vartak, M.}, \bibinfo{author}{Subramanyam, H.},
  \bibinfo{author}{Lee, W.-E.}, \bibinfo{author}{Viswanathan, S.},
  \bibinfo{author}{Husnoo, S.}, \bibinfo{author}{Madden, S.}, \&
  \bibinfo{author}{Zaharia, M.} (\bibinfo{year}{2016}).
\newblock \bibinfo{title}{{ModelDB}: a system for machine learning model
  management}.
\newblock In {\it \bibinfo{booktitle}{Workshop on Human-In-the-Loop Data
  Analytics}\/} (pp. \bibinfo{pages}{1--3}).
\bibitem[{Wang et~al.(2017)Wang, Min, Zhang \& Wu}]{wang2017active}
\bibinfo{author}{Wang, M.}, \bibinfo{author}{Min, F.}, \bibinfo{author}{Zhang,
  Z.-H.}, \& \bibinfo{author}{Wu, Y.-X.} (\bibinfo{year}{2017}).
\newblock \bibinfo{title}{Active learning through density clustering}.
\newblock {\it \bibinfo{journal}{Expert Systems with Applications}\/},  {\it
  \bibinfo{volume}{85}\/}, \bibinfo{pages}{305--317}.
\bibitem[{van~der Weide et~al.(2017)van~der Weide, Papadopoulos, Smirnov,
  Zielinski \& van Kasteren}]{van2017versioning}
\bibinfo{author}{van~der Weide, T.}, \bibinfo{author}{Papadopoulos, D.},
  \bibinfo{author}{Smirnov, O.}, \bibinfo{author}{Zielinski, M.}, \&
  \bibinfo{author}{van Kasteren, T.} (\bibinfo{year}{2017}).
\newblock \bibinfo{title}{Versioning for end-to-end machine learning
  pipelines}.
\newblock In {\it \bibinfo{booktitle}{Workshop on Data Management for
  End-to-End Machine Learning}\/} (pp. \bibinfo{pages}{1--9}).
\bibitem[{Weiss et~al.(2016)Weiss, Khoshgoftaar \& Wang}]{weiss2016survey}
\bibinfo{author}{Weiss, K.}, \bibinfo{author}{Khoshgoftaar, T.~M.}, \&
  \bibinfo{author}{Wang, D.} (\bibinfo{year}{2016}).
\newblock \bibinfo{title}{A survey of transfer learning}.
\newblock {\it \bibinfo{journal}{Journal of Big data}\/},  {\it
  \bibinfo{volume}{3}\/}, \bibinfo{pages}{9}.
\bibitem[{Werner et~al.(2017)Werner, Bedford, Cooke, Hanea \&
  Morales-Napoles}]{werner2017expert}
\bibinfo{author}{Werner, C.}, \bibinfo{author}{Bedford, T.},
  \bibinfo{author}{Cooke, R.~M.}, \bibinfo{author}{Hanea, A.~M.}, \&
  \bibinfo{author}{Morales-Napoles, O.} (\bibinfo{year}{2017}).
\newblock \bibinfo{title}{Expert judgement for dependence in probabilistic
  modelling: A systematic literature review and future research directions}.
\newblock {\it \bibinfo{journal}{European Journal of Operational Research}\/},
  {\it \bibinfo{volume}{258}\/}, \bibinfo{pages}{801--819}.
\bibitem[{Yoon et~al.(2020)Yoon, Arik \& Pfister}]{yoon2020data}
\bibinfo{author}{Yoon, J.}, \bibinfo{author}{Arik, S.}, \&
  \bibinfo{author}{Pfister, T.} (\bibinfo{year}{2020}).
\newblock \bibinfo{title}{Data valuation using reinforcement learning}.
\newblock In {\it \bibinfo{booktitle}{International Conference on Machine
  Learning}\/} (pp. \bibinfo{pages}{10842--10851}).
\bibitem[{Yu et~al.(2019)Yu, Berkovsky, Taib, Zhou \& Chen}]{yu2019trust}
\bibinfo{author}{Yu, K.}, \bibinfo{author}{Berkovsky, S.},
  \bibinfo{author}{Taib, R.}, \bibinfo{author}{Zhou, J.}, \&
  \bibinfo{author}{Chen, F.} (\bibinfo{year}{2019}).
\newblock \bibinfo{title}{Do i trust my machine teammate? an investigation from
  perception to decision}.
\newblock In {\it \bibinfo{booktitle}{ACM International Conference on
  Intelligent User Interfaces}\/} (pp. \bibinfo{pages}{460--468}).
\bibitem[{Zhang et~al.(2016)Zhang, Kumar \& R{\'e}}]{zhang2016materialization}
\bibinfo{author}{Zhang, C.}, \bibinfo{author}{Kumar, A.}, \&
  \bibinfo{author}{R{\'e}, C.} (\bibinfo{year}{2016}).
\newblock \bibinfo{title}{Materialization optimizations for feature selection
  workloads}.
\newblock {\it \bibinfo{journal}{ACM Transactions on Database Systems}\/},
  {\it \bibinfo{volume}{41}\/}, \bibinfo{pages}{1--32}.
\bibitem[{Zhi et~al.(2013)Zhi, Wang, Qian, Butler, Ramakrishnan \&
  Davidson}]{zhi2013clustering}
\bibinfo{author}{Zhi, W.}, \bibinfo{author}{Wang, X.}, \bibinfo{author}{Qian,
  B.}, \bibinfo{author}{Butler, P.}, \bibinfo{author}{Ramakrishnan, N.}, \&
  \bibinfo{author}{Davidson, I.} (\bibinfo{year}{2013}).
\newblock \bibinfo{title}{Clustering with complex constraints-algorithms and
  applications.}
\newblock In {\it \bibinfo{booktitle}{AAAI Conference on Artificial
  Intelligence}\/}.
\bibitem[{Zhou(2012)}]{zhou2012ensemble}
\bibinfo{author}{Zhou, Z.-H.} (\bibinfo{year}{2012}).
\newblock {\it \bibinfo{title}{Ensemble methods: foundations and
  algorithms}\/}.
\newblock \bibinfo{publisher}{CRC press}.
\bibitem[{Zhou \& Xu(2007)}]{zhou2007relation}
\bibinfo{author}{Zhou, Z.-H.}, \& \bibinfo{author}{Xu, J.-M.}
  (\bibinfo{year}{2007}).
\newblock \bibinfo{title}{On the relation between multi-instance learning and
  semi-supervised learning}.
\newblock In {\it \bibinfo{booktitle}{International Conference on Machine
  Learning}\/} (pp. \bibinfo{pages}{1167--1174}).
\bibitem[{Zhuang et~al.(2016)Zhuang, Lin, Shen \& Reid}]{zhuang2016fast}
\bibinfo{author}{Zhuang, B.}, \bibinfo{author}{Lin, G.}, \bibinfo{author}{Shen,
  C.}, \& \bibinfo{author}{Reid, I.} (\bibinfo{year}{2016}).
\newblock \bibinfo{title}{Fast training of triplet-based deep binary embedding
  networks}.
\newblock In {\it \bibinfo{booktitle}{IEEE Conference on Computer Vision and
  Pattern Recognition}\/} (pp. \bibinfo{pages}{5955--5964}).

\end{thebibliography}






\end{document}